\documentclass[lettersize,journal]{IEEEtran}
\usepackage{amsmath,amsfonts}
\usepackage{algorithmic}
\usepackage{array}
\usepackage[caption=false,font=normalsize,labelfont=sf,textfont=sf]{subfig}
\usepackage{textcomp}
\usepackage{stfloats}
\usepackage{url}
\usepackage{verbatim}
\usepackage{graphicx}

\usepackage{amssymb}
\usepackage[colorlinks,linkcolor=red,anchorcolor=blue,citecolor=green]{hyperref}
\usepackage{makecell}
\usepackage{color}
\usepackage{wrapfig}
\usepackage{booktabs}
\usepackage{multirow}
\usepackage{hyperref}
\usepackage{tabularx}
\usepackage{mathtools}
\usepackage{float}
\usepackage{cite}
\usepackage{xcolor}
\usepackage[normalem]{ulem}
\usepackage{comment}

\def\BibTeX{{\rm B\kern-.05em{\sc i\kern-.025em b}\kern-.08em
    T\kern-.1667em\lower.7ex\hbox{E}\kern-.125emX}}
\usepackage{balance}

\newcommand{\hwj}[1]{\textcolor{black}{#1}}
\newcommand{\hwjre}[1]{\textcolor{black}{#1}}

\newcommand{\taore}[1]{\textcolor{black}{#1}}
\newcommand{\wang}[1]{\textcolor{black}{#1}}
\newcommand{\Best}[1]{\textcolor{red}{\textbf{#1}}}
\newcommand{\SecondBest}[1]{\textcolor{cyan}{\textbf{#1}}}
\newcommand{\wangre}[1]{\textcolor{black}{#1}}

\def\etal{{\it{et al.}}}

\begin{document}
\title{MDeRainNet: An Efficient Macro-pixel Image Rain Removal Network}
\author{Tao Yan, Weijiang He, Chenglong Wang, Cihang Wei, Xiangjie Zhu, Yinghui Wang and Rynson W.H. Lau
\thanks{This work was supported by the National Natural Science Foundation of China (Grant No. 61902151 and 62172190) and the Natural Science Foundation of Jiangsu Province, China (Grant No. BK20170197).}
\thanks{Tao Yan,  Weijiang He,  Chenglong Wang, Cihang Wei, Xiangjie Zhu and Yinghui Wang are with the School of Artificial Intelligence and Computer Science, Jiangnan University, 214122, Wuxi, Jiangsu, China.}
\thanks{Rynson W.H. Lau is with the Department of Computer Science, City University of Hong Kong, Hong Kong SAR, China.}
\thanks{Corresponding author is Tao Yan, E-mail: yantao.ustc@gmail.com.}
}

\markboth{Journal of \LaTeX\ Class Files,~Vol.~18, No.~9, September~2020}%
{How to Use the IEEEtran \LaTeX \ Templates}


\maketitle

\begin{abstract}
Since raining weather always degrades image quality and poses significant challenges to most computer vision-based intelligent systems, image de-raining has been a hot research topic in computer vision community. 
Fortunately, in a rainy light field (LF) image, background obscured by rain streaks in one sub-view may be visible in the other sub-views, and implicit depth information and recorded 4D structural information may benefit rain streak detection and removal.
However, existing LF image rain removal methods either do not fully exploit the global correlations of 4D LF data, or only utilize partial sub-views (i.e., under-utilization of the rich angular information), resulting in sub-optimal rain removal performance and no-equally good quality for all de-rained sub-views.
In this paper, we propose an efficient neural network, called \textit{MDeRainNet}, for rain streak removal from LF images. 
The proposed network adopts a multi-scale encoder-decoder architecture, which directly works on Macro-pixel images (MPIs) for improving the rain removal performance. To fully model the global correlation between the spatial information and the angular information, we propose an \textit{Extended Spatial-Angular Interaction (ESAI)} module to merge the two types of information, in which a simple and effective Transformer-based \textit{Spatial-Angular Interaction Attention (SAIA)} block is also proposed for modeling long-range geometric correlations and making full use of the angular information. 
\taore{Furthermore, to improve the generalization performance of our network on real-world rainy scenes, we propose a novel semi-supervised learning framework four our \textit{MDeRainNet}, which utilizes multi-level KL loss to bridge the domain gap between features of synthetic and that of real-world rain streaks and introduces colored-residue image guided contrastive regularization to reconstruct rain-free images.}
Extensive experiments conducted on both synthetic and real-world LFIs demonstrate that our method outperforms the state-of-the-art methods both quantitatively and qualitatively. 
\end{abstract}

\begin{IEEEkeywords}
Rain removal, light field images, macro-pixel image (MPI), deep learning, \hwj{semi-supervised learning}.
\end{IEEEkeywords}

\newcommand{\subwidth}{0.240}
\newcommand{\ssubwidth}{0.120}
\begin{figure}[t]
	\renewcommand{\tabcolsep}{0.6pt}
	\renewcommand\arraystretch{0.6}
	\begin{center}
		\begin{tabular}{cccccccc}
            \multicolumn{2}{c}{\includegraphics[width=\subwidth\linewidth]{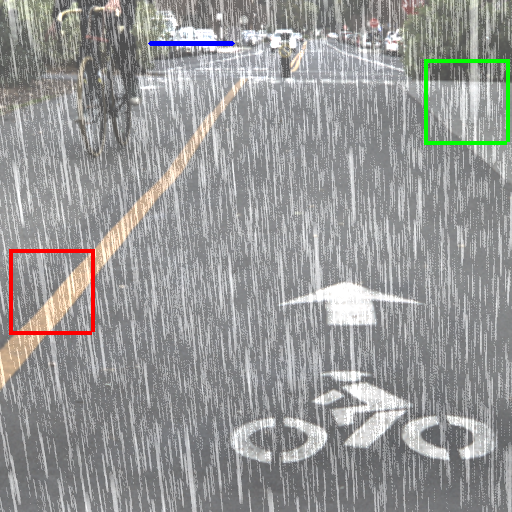}} &
            \multicolumn{2}{c}{\includegraphics[width=\subwidth\linewidth]{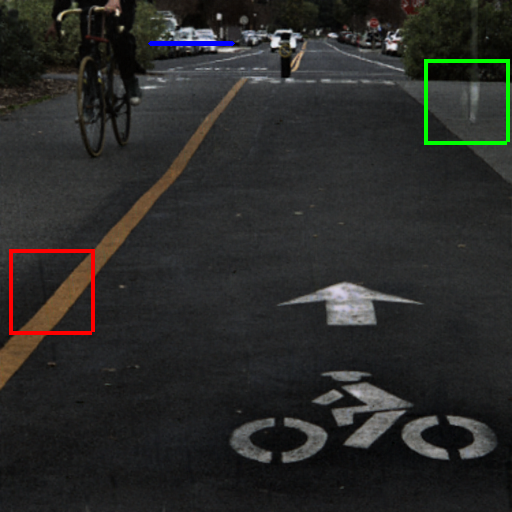}} &
            \multicolumn{2}{c}{\includegraphics[width=\subwidth\linewidth]{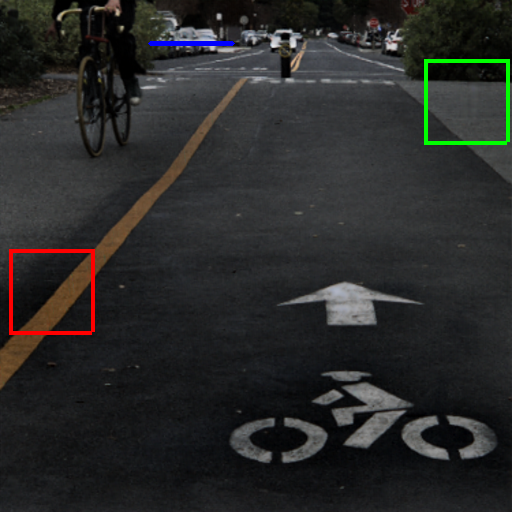}} &
            \multicolumn{2}{c}{\includegraphics[width=\subwidth\linewidth]{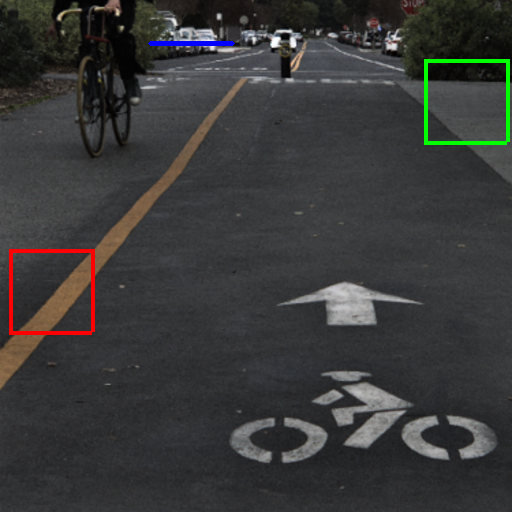}} \\

            \includegraphics[width=\ssubwidth\linewidth]{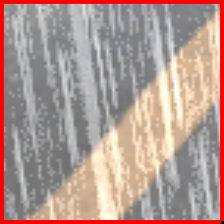} &
            \includegraphics[width=\ssubwidth\linewidth]{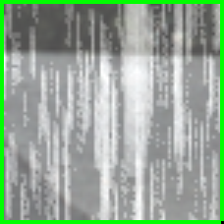} &
            \includegraphics[width=\ssubwidth\linewidth]{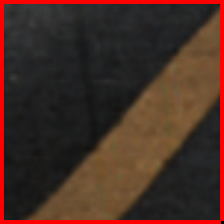} &
            \includegraphics[width=\ssubwidth\linewidth]{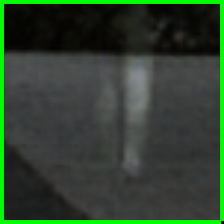} &
            \includegraphics[width=\ssubwidth\linewidth]{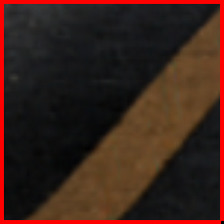} &
            \includegraphics[width=\ssubwidth\linewidth]{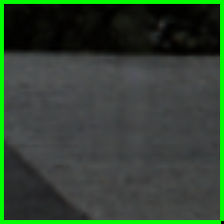} &
            \includegraphics[width=\ssubwidth\linewidth]{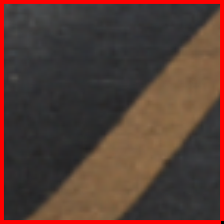} &
            \includegraphics[width=\ssubwidth\linewidth]{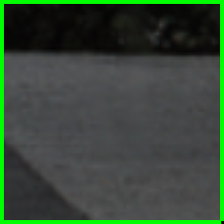}  \\

            \multicolumn{2}{c}{\includegraphics[width=0.117\textwidth,height=0.017\textwidth]{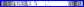}} &
            \multicolumn{2}{c}{\includegraphics[width=0.117\textwidth,height=0.017\textwidth]{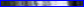}} &
            \multicolumn{2}{c}{\includegraphics[width=0.117\textwidth,height=0.017\textwidth]{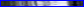}}&
            \multicolumn{2}{c}{\includegraphics[width=0.117\textwidth,height=0.017\textwidth]{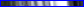}}  \\

            \multicolumn{2}{c}{\includegraphics[width=\subwidth\linewidth]{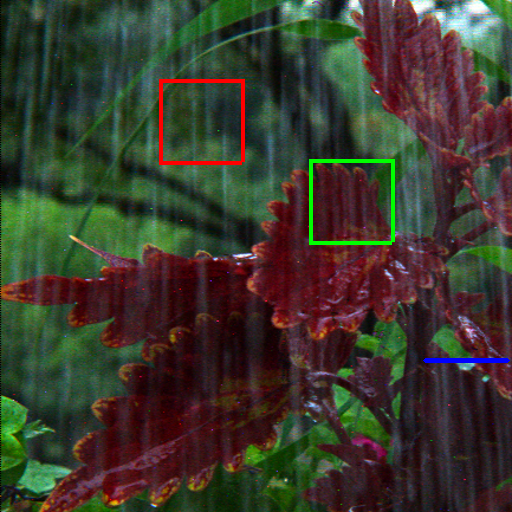}} &
            \multicolumn{2}{c}{\includegraphics[width=\subwidth\linewidth]{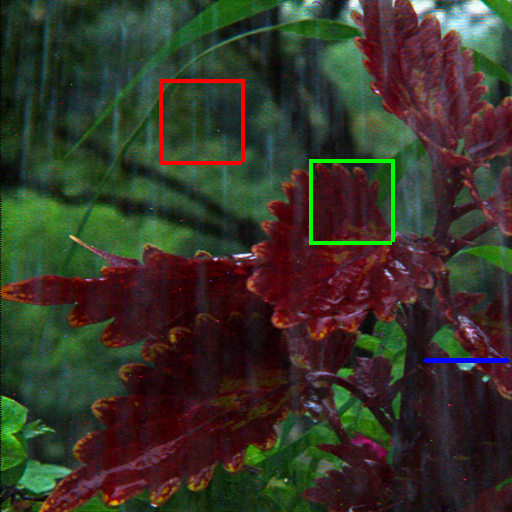}} &
            \multicolumn{2}{c}{\includegraphics[width=\subwidth\linewidth]{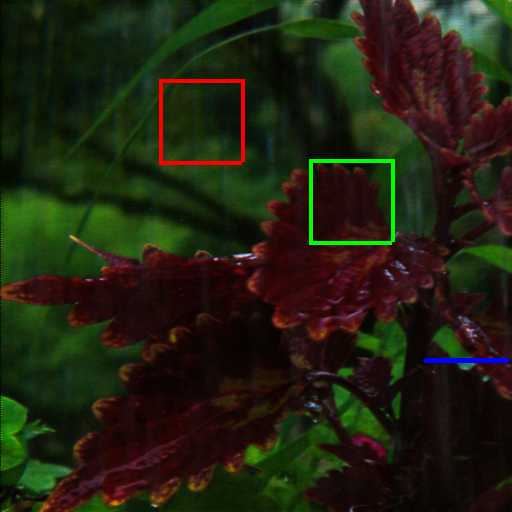}} &
            \multicolumn{2}{c}{\includegraphics[width=\subwidth\linewidth]{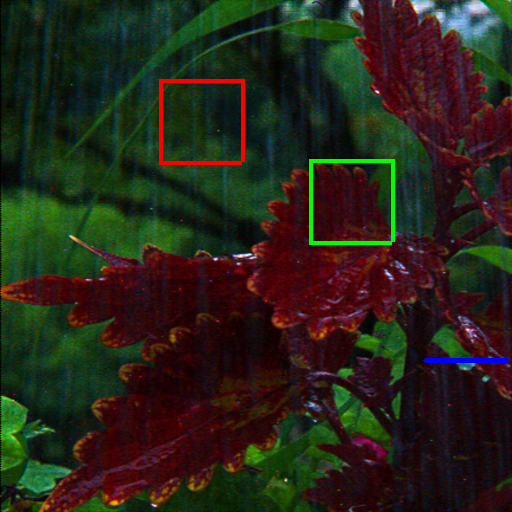}} \\

            \includegraphics[width=\ssubwidth\linewidth]{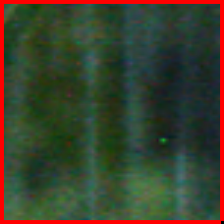} &
            \includegraphics[width=\ssubwidth\linewidth]{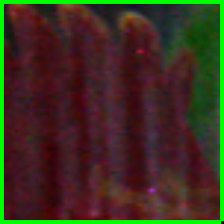} &
            \includegraphics[width=\ssubwidth\linewidth]{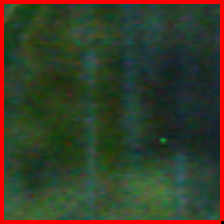} &
            \includegraphics[width=\ssubwidth\linewidth]{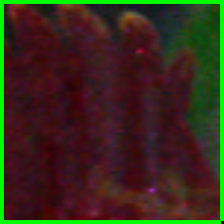} &
            \includegraphics[width=\ssubwidth\linewidth]{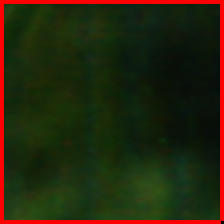} &
            \includegraphics[width=\ssubwidth\linewidth]{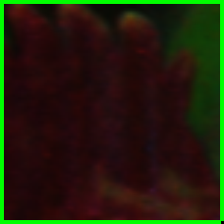} &
            \includegraphics[width=\ssubwidth\linewidth]{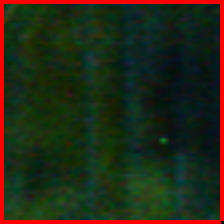} &
            \includegraphics[width=\ssubwidth\linewidth]{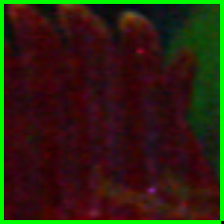}  \\

            \multicolumn{2}{c}{\includegraphics[width=0.117\textwidth,height=0.017\textwidth]{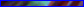}} &
            \multicolumn{2}{c}{\includegraphics[width=0.117\textwidth,height=0.017\textwidth]{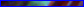}} &
            \multicolumn{2}{c}{\includegraphics[width=0.117\textwidth,height=0.017\textwidth]{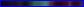}}&
            \multicolumn{2}{c}{\includegraphics[width=0.117\textwidth,height=0.017\textwidth]{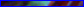}}  \\

            \multicolumn{2}{c}{\footnotesize{(a) Input LFI}}&
            \multicolumn{2}{c}{\footnotesize{(b) Restormer~\cite{zamir2022restormer}}}&
            \multicolumn{2}{c}{\footnotesize{(c) Yan~\etal~\cite{yan2023rain}}}&
            \multicolumn{2}{c}{\footnotesize{(d) Ours}}\\
		\end{tabular}
	\end{center}
	\vspace{-0.016\textwidth}
	\caption{Comparison of our method with two state-of-the-art de-raining methods~\cite{zamir2022restormer,yan2023rain}, \hwj{tested on a synthetic rainy LFI (first row) and a real-world rainy LFI (second row) from RLMB~\cite{yan2023rain}.} 
 The horizontal EPI (epipolar plane image) extracted along the blue line across the sub-views in the same row of the rainy LF image is also visualized to show the angular consistency (i.e., image content consistency across different de-rained sub-views).}
	\label{firstfigure}
        \vspace{-4.5mm}
\end{figure}

\section{Introduction}
Unlike regular cameras, plenoptic cameras could record target scenes with 4D light field data, i.e., spatial and angular information. Comparing with regular images, LFIs benefit many attractive applications, such as depth estimation~\cite{wang2022occlusion}, foreground de-occlusion~\cite{wang2022effective} and saliency detection~\cite{jing2021occlusion}. Raining as a common weather condition, may greatly affect the performance of downstream computer vision systems, such as autonomous driving and video surveillance. Therefore, removing rainy artifacts (i.e., rain streaks and raindrops) from rainy images is an important pre-procedure for most outdoor vision systems and has received extensive research attention in recent years.

\footnotetext[1]{Our code will be available at: \url{https://github.com/YT3DVision/MDeRainNet}.}

To remove rainy artifacts from images, early research mainly focused on single image rain removal (SIRR)~\cite{ren2019progressive,jiang2020multi}, and a few works focus on video rain removal~\cite{yang2020self,yan2021self}. 
A simple solution for LFI rain streak removal (LFRSR) is to apply SIRR methods to each sub-view of the input LF image independently. However, despite the great progress of SIRR methods~\cite{quan2021removing,wang2021attentive,zamir2021multi,zamir2022learning,liang2022drt,valanarasu2022transweather,zamir2022restormer,xiao2022image,chen2023learning} based on CNNs \taore{and/or} Transformers, their solutions are always sub-optimal due to one inherent limitation that these rain streak detection and removal are highly dependent on the appearance (e.g., textures and shapes) of rain streak observed in a single image. Therefore, directly utilizing these SIRR methods to remove rain streaks from LF images would ignore the intrinsic relationship between different sub-views of an LF image, which may lead to angular inconsistency in the de-rained LF image. To address this problem, some recently proposed methods~\cite{ding2021rain,yan2023rain,jing2022light} have made great efforts to fully explore and exploit information from LF images for rain removal. An example of the performances of the single image de-raining method~\cite{zamir2022restormer} and the LF image de-raining method~\cite{yan2023rain} working for LF image de-raining are shown in Figure~\ref{firstfigure}.


Despite these LF image rain removal methods have achieved great progress, there are still two limitations.
First, the structure (e.g. high dimension) of 4D LF data leads to difficulty in feature learning. Thus, it is necessary to re-organize the input LFI reasonably to facilitate networks to learn the feature representation for removing rain streaks from all sub-views of an LF image.
Second, the receptive fields of such CNN-based de-raining networks are always limited and their representation learning efficiencies are low. 2D or high-dimensional (3D or 4D) convolutions utilized in these methods may be inefficient for feature learning and lead to high computation costs. Recently, Transformer has been proven to be effective for modeling long-range dependency, and has been successfully applied to image restoration~\cite{wang2022uformer,zamir2022restormer,xiao2022image,liang2022light,wang2022effective}.

\begin{figure}[h]
        \renewcommand{\subwidth}{0.9}
	\renewcommand{\tabcolsep}{0.02pt}
	\renewcommand\arraystretch{0.8}
	\begin{center}
	\includegraphics[width = \subwidth\linewidth]{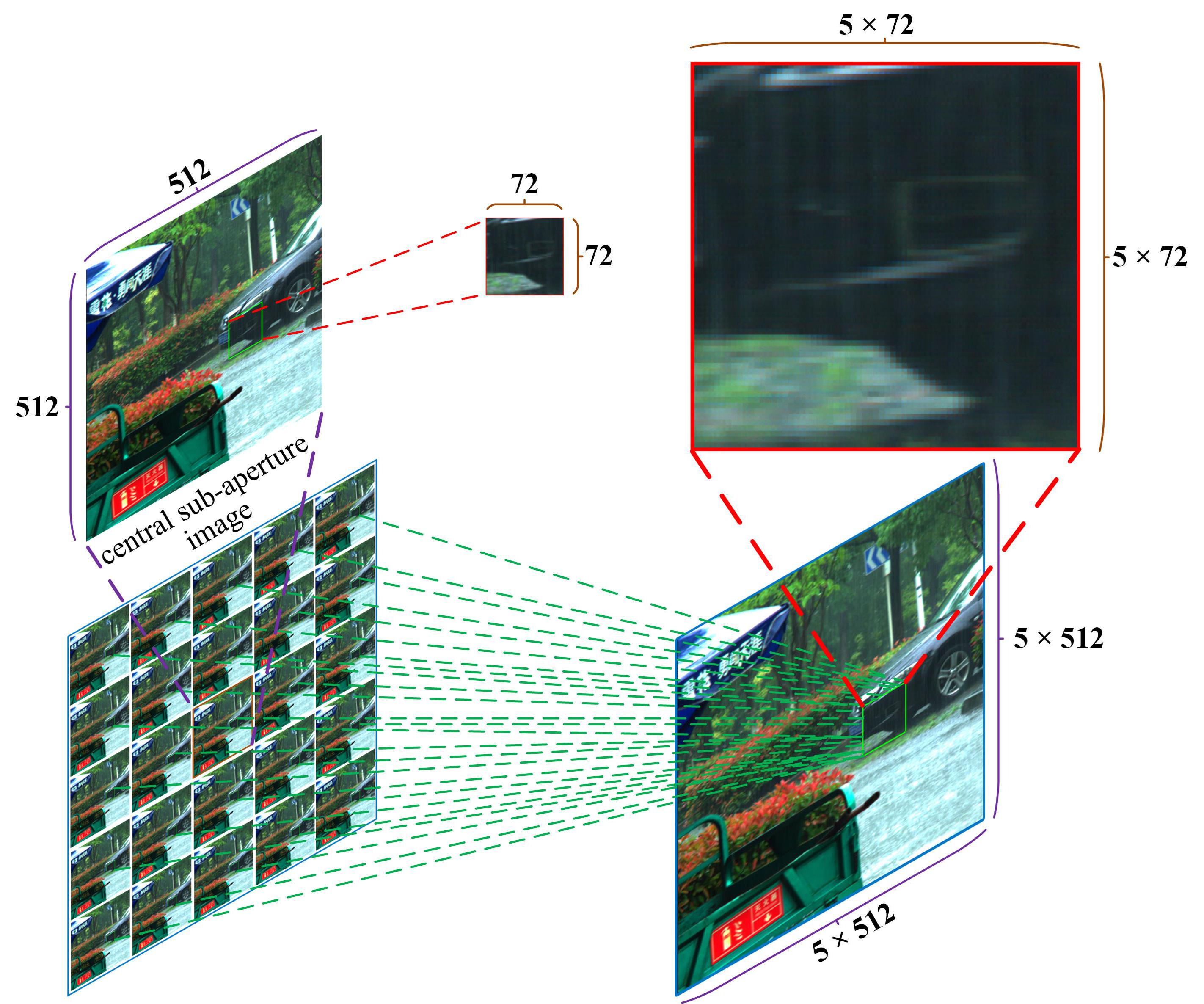} 
	\end{center}
        \vspace{-0.016\textwidth}
\caption{An illustration of the relationship between the sub-views and the MPI (Macro-pixel image) representations of an LF image. The MPI consists of a large number of macro-pixels, and the number of pixels within a single macro-pixel is equal to the number of sub-views of the LF image. Each macro-pixel corresponds to a position sample of LF, and each pixel within a macro-pixel corresponds to a direction sample of LF at that position.
 It is worth noting that the rain streaks on MPI are \taore{always} larger and sharper than the rain streaks in a single sub-view, as shown in the zoomed-in patch.}
 \label{fig:mpi}
 \vspace{-0.016\textwidth}
\end{figure}

LF images can record depth information and multiply sub-views of target scenes, but due to limited spatial resolutions of LF images, rain streaks appearing on real-world LF images may be always tiny and hard to accurately detect.
\taore{Fortunately,} we observed that rain streaks appearing in the MPI of an LF image are much larger and sharper than those in single sub-views of the LF image, as shown in Figure~\ref{fig:mpi}. The motivation of our method is that to detect rain streaks on the MPI of a rainy LFI would be more effective and efficient than that on the sub-views or the EPIs of a rainy LF Image. 

In this paper, we propose an effective LFI rain streak removal network, called \textit{MDeRainNet}, for rain streak removal from MPIs of LFIs. Specifically, our network takes as input the MPI of a rainy LF image, and adopts an encoder-decoder architecture to extract multi-scale features. \taore{Our \textit{MDeRainNet} can fully explore and exploit the angular and spatial information of an LF image. Moreover, it can better preserve the content coherence of subviews of an LF image than existing methods.}
Since the low-level stages can always extract rich detailed features and the size of the feature maps is large, an \textit{Spatial Feature Extractor} (SFE) and a convolution-based \textit{Modified Disentangling Block} (MDB) are used to extract low-level features. 
We also propose an \textit{Extended Spatial-Angular Interaction} (ESAI) module for extracting high-level features in the later stage of our network, 
\taore{which could model the correlation between spatial information and angular information over long range to capture global patterns of rain streaks.} 

\hwj{\taore{Though} the semi-supervised de-raining method recently proposed by Yan~\etal~\cite{yan2023rain} achieves good performance on real-world rainy LFIs, its de-rained results \taore{may be} over-smoothed and \taore{cannot} maintain angular consistency well.
\taore{Given that} the high-level (semantic) features of rain streaks extracted from synthetic and real-world rainy images \taore{should be} similar, \taore{we enforce the consistency of the distributions of high-level features of rain streaks from synthetic rainy images and that from real-world images} by minimizing Kullback–Leibler Divergence loss (KL-Loss) at multi-level \taore{scales}~\cite{wei2019semi,cui2022semi}.} 
\hwj{Furthermore, we design a contrastive regularization guided by colored rain-free residue images~\cite{li2018robust}. 
\taore{There are two de-raining methods\cite{li2019heavy,yi2021structure} that utilize rain-free residue images to decompose and maintain high-frequency background details.}
We propose to utilize rain-free residue images as positive samples in contrastive regularization for semi-supervised learning-based rain removal in our method. A real-world de-rained image should be closer to its corresponding rain-free residue image but away from the original rainy image in feature space.}  Therefore, by combining the proposed multi-level KL divergence and colored-residue image guided contrastive regularization, both the rain streaks domain and clean background domain can be fully utilized \taore{for supervision in our method}. Thus, \taore{our \textit{MDeRainNet} is able to well} solve the domain-shift problem and \taore{improve} generalization on real-world rainy LF images.

The main contributions of our work can be summarized as follows:
\begin{itemize}
	\item We propose a novel and efficient rain removal network, called \textit{MDeRainNet}, for rain streak removal from LFIs, which takes the MPI of an LFI as input. 
	\item  We propose an Extended Spatial-Angular Interaction (ESAI) module for fusing spatial and angular features, which can interact with spatial and angular features over long-range.
 
        \item \taore{We propose a novel semi-supervised learning framework for improving the generalization of our network on real-world rainy images, which utilizes multi-level KL loss to bridge the domain gap between features of synthetic and that of real-world rain streaks and introduces colored-residue image guided contrastive regularization to reconstruct the rain-free image.}
        
        \item Extensive experiments conducted on synthetic and real-world rainy datasets demonstrate that our network outperforms other state-of-the-art methods both quantitatively and qualitatively, with low computation cost and less inference time.
\end{itemize}

\section{Related Works} 
In this section, we briefly review the existing LF image rain removal methods, single image rain removal methods and video rain removal methods. In addition, some recently proposed multiple weather removal methods are also surveyed.

\subsection{LFI Rain Removal}  
The research on LFI rain removal has not yet flourished, but several landmark works have emerged recently. Ding~\etal~\cite{ding2021rain} proposed a semi-supervised rain streak removal method based on GAN and Gaussian Mixture Model. The disadvantage is that it can only remove rain streaks by using sub-views in the same row/column of the rainy LFI at a time. Most recently, Yan~\etal~\cite{yan2023rain} proposed a semi-supervised rain streak removal network, which uses 4D convolutional layers to make full use of all sub-views of the input LF image for rain streak detection and removal. First, a dense sub-network based on multi-scale Gaussian processes is proposed to extract rain streaks from all input LF sub-views, and then the depth map and the fog map are estimated from the rough de-rained LF image. Finally, an adversarial and recurrent neural sub-network is constructed to progressively recover the rain-free LFI. 
Jing~\etal~\cite{jing2022light} proposed a 4D re-sampling network for LFI raindrop removal, which exploits the complementary pixel information of the raindrop-free regions of the input LFI to produce the raindrop-free LFI. 
All these de-raining methods rely on 3D/4D convolution to exploit sub-views of an LFI, which may not fully utilize the abundant information embedding in rainy LFIs and are non-efficient due to high computation cost. 

Wang~\etal~\cite{wang2022effective} proposed a novel LFI de-occlusion network, which combines Swin Transformer with CNN to extract both global and local features for restoring occlusion-free images. The network first uses CNNs to extract local features at shallow layers, and then utilizes Transformers to capture global features of large size occlusions at deep layers. Wang~\etal~\cite{wang2022disentangling} proposed a general disentangling mechanism to extract the spatial feature, the angular feature and the EPI feature from the input MPI, for LFI super-resolution and depth estimation.

\subsection{Single-Image Rain Removal}
Unlike LF-based methods that utilize complementary information among different sub-view images, single-image de-raining methods remove rain from the background by analyzing the visual information of a single image. Traditional de-raining methods~\cite{kang2011automatic,luo2015removing,li2016rain,zhu2017joint,deng2018directional} usually separate rain layer from the background layers by exploring the physical properties or prior knowledge of rain streaks (raindrops).
In recent years, deep learning-based methods have dominated the single-image de-raining research topic. Fu~\etal~\cite{fu2017removing,fu2017clearing} proposed the earliest deep learning-based methods, which adopt CNN to remove rain streaks from the high-frequency detail layer of a rainy image. Later, a lot of deep learning-based networks have been proposed to improve the single image de-raining performance, including multi-scale deep networks~\cite{jiang2020multi,yasarla2019uncertainty}, multi-stage neural networks~\cite{ren2019progressive,zamir2021multi}, conditional generative adversarial networks~\cite{zhang2019image} and attention-based encoder-decoder networks~\cite{wang2020cascaded,wang2021attentive,guo2023sky}. \wang{For the effectiveness and efficiency of image deraining, Jiang~\etal~\cite{jiang2021rain} firstly devised a novel coupled representation module named CRM to investigate the blending correlations between rain streaks and rain-free details. Later, Jiang~\etal~\cite{jiang2022magic} propose a network that combines the advantages of Transformer and CNN for rain perturbation removal and background recovery.}
Most recently, several semi-supervised and unsupervised de-raining methods~\cite{yasarla2020syn2real,huang2021memory,yu2021unsupervised,chen2022unpaired} have also been proposed for de-raining real-world rainy images. \wang{Most recently, Fu~\etal~\cite{fu2023continual} designed a patchwise hypergraph convolutional module and a biological brain-inspired continual learning algorithm to achieve better generalizability and adaptability in real-world scenes. }

\wang{There are two methods that improve the de-raining performance by utilizing rain-free residue images~\cite{li2018robust}. Li et al.\cite{li2019heavy} proposed a filtering framework guided by a rain-free residue image, which uses the residue image as reference to guide the decomposition of low-frequency and high-frequency components. Later, Yi et al.\cite{yi2021structure} proposed a structure-preserving de-raining network with residue image guidance.}

These single image de-raining methods could not be directly utilized for LFI de-raining due to two main reasons. First, they can not make full use of abundant information embedding in LFIs for high-performance de-raining. Second, they could not ensure image content consistency between all sub-views of a de-rained LFI.

\subsection{Multiple Weather Removal}


Recently, some unified image restoration networks, such as IPT~\cite{chen2021pre}, Uformer~\cite{wang2022uformer}, TransWeather~\cite{valanarasu2022transweather} and Restormer~\cite{zamir2022restormer} have been proposed.
IPT~\cite{chen2021pre} is the first network, which uses standard Transformer blocks to form a multi-task learning framework for image super-resolution, de-raining and de-noising. 
Uformer~\cite{wang2022uformer} build a U-shaped Transformer network based on Swin Transformer for multi-task image restoration, including de-noising, de-blurring and de-raining. 
Though these methods achieve impressive performance for multi-task single image restoration, they can not be directly utilized for LFI de-raining.

\begin{figure*}[t]
	\centering
	\includegraphics[width=0.9\textwidth]{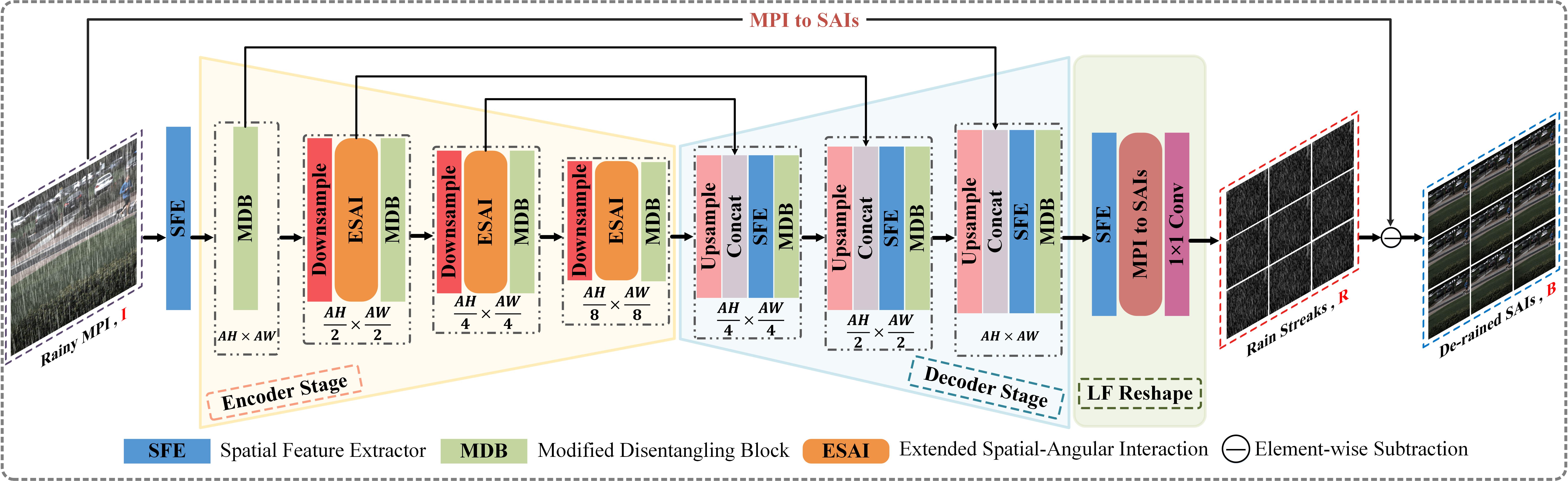} 
	\caption{Overall architecture of our \hwj{\textit{MDeRainNet}}. The input of our network is rainy MPI, and the network learns LF rain streak residuals as a whole. 
    We feed the MPI into an encoder-decoder network to extract hierarchical features. Firstly, the combination of SFE and MDB is used to extract low-level features from the input MPI in the encoder. Subsequently, we combine efficient ESAI and MDB to complement the receptive fields of these two modules on multiple scales of the encoder, so as to obtain locality while capturing long-range dependencies. Finally, at the end of the encoder, the rain residuals estimated on MPI are reshaped into a sub-aperture images array to get the final de-rained SAIs.}
    \label{fig:network}\
    \vspace{-0.016\textwidth}
\end{figure*} 

\begin{figure}[h]
    \centering
    \includegraphics[width=0.36\textwidth]{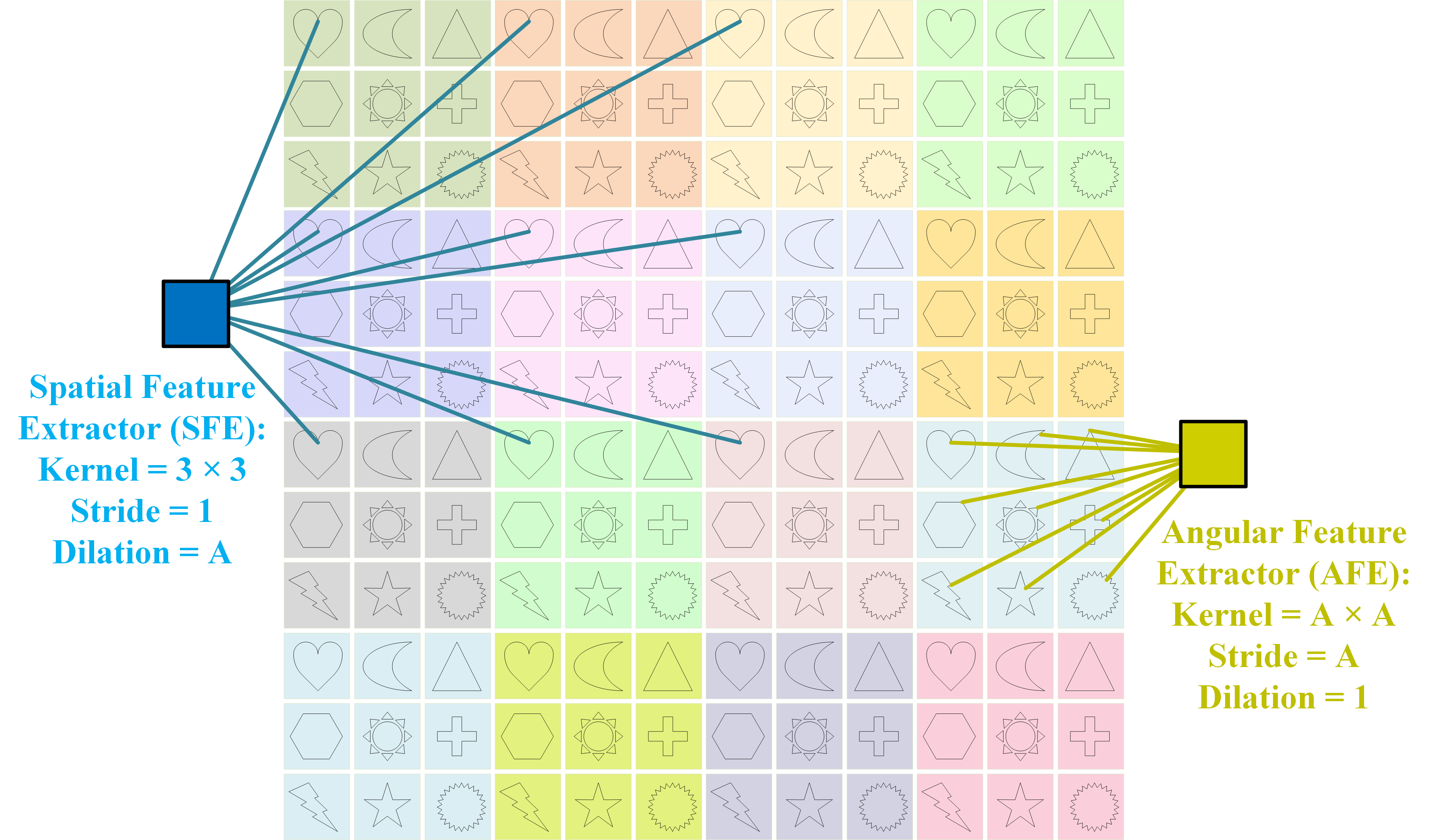} 
    \caption{The working principle of spatial and angular feature extractors. We show an example of an LF with an angular resolution of $3\times 3$ (i.e., $A = 3$) and a spatial resolution of $4\times 4$ (i.e., $H = 4$, $W = 4$). For better visualization, different macro-pixels are filled with different background colors, and pixels from different sub-views are represented by different labels (e.g., moon or pentagram). SFE extracts spatial features within the same view, while AFE extracts angular features within the same macro-pixel, thereby decoupling spatial and angular information.}
    \label{fig:SFE_AFE}
    \vspace{-0.016\textwidth}
\end{figure}

\section{Our Proposed Network}
In this section, \hwj{we describe our \textit{MDeRainNet} and its semi-supervised learning framework in detail.}

\subsection{Background and Motivation}
A 4D LFI can be expressed as $L\in\mathbb{R}^{U\times V\times H\times W\times C}$, where $(H,W)$, $(U,V)$ and $C$ denotes the spatial resolution, the angular resolution and the RGB channels, respectively. 
As shown in Figure~\ref{fig:mpi} (a), an array of sub-views with different angular coordinates $(u,v)$ is shown, each of which is called a sub-aperture image (SAI) or sub-view $L(u,v)\in\mathbb{R}^{H\times W\times C}$. Recently, most LFI processing methods take as input a subset of sub-views of an input LFI. However, such form of input is not convenient for extracting spatial and angular information, since angular information is implicitly embedded in different sub-views. On the other hand, some LFI processing methods take EPIs (Epipolar Plane Images) as input. Since an EPI is only a 2D horizontal or vertical slice of a 4D LF image, it is difficult to exploit both spatial and angular information from EPIs. Therefore, in order to make full use of both the spatial information and angular information for rain streak removal, following previous super-resolution methods\cite{wang2020spatial,wang2022disentangling}, our network directly takes the MPI of the input LFI as input. As far as we know, our \textit{MDeRainNet} is the first method, which directly takes as input MPIs for LF image rain streak removal.

In Figure~\ref{fig:SFE_AFE}, we provide an illustration of the angular and spatial feature extractors. By taking the MPI of an LFI as input, the spatial information and angular information embedding in the input LFI can be well extracted by leveraging domain-specific convolutions, i.e, spatial feature extractor (SFE) and angular feature extractor (AFE)~\cite{wang2020spatial,wang2022disentangling}. 

Motivated by our observations that rain streaks appearing in MPI not only have larger sizes than that of rain streaks appearing in each single sub-view of the input LFI, but also have sharper edges (as shown in the zoomed-in area of Figure~\ref{fig:mpi}), it may be much effective to remove rain streaks from MPIs. 
In another word, although both the MPI and sub-views of an LFI are 2D images, the rain streaks in MPIs contain richer structure information than that in single sub-views, which may help our method to extract rain streak features from MPIs. Furthermore, since the MPI contains both spatial and angular information of an LFI in a 2D image format, extracting rain streaks from it will be more convenient than that from sub-views of the input rainy LFI. 

\subsection{Overview of Our MDeRainNet}

Our \textit{MDeRainNet} adopts an encoder-decoder structure, as shown in Figure~\ref{fig:network}. \taore{The complete architecture of our \textit{MDeRainNet} with the semi-supervised learning framework is shown in Figure~\ref{fig:framework}}. To effectively detect rain streaks from MPIs of LFIs, we propose an \textit{Extended Spatial-Angular Interaction} (ESAI) module, which is able to model the correlation between spatial information and angular information over long-range to capture global patterns of rain streaks. We also introduced a Modified Disentangling Block (MDB) for extracting and integrating spatial and angular features. The ESAI and MDB are used to form the encoder of our \textit{MDeRainNet}, which can combine the advantages of Transformer and CNN to ensure that both g lobal features and local features can be retained. 
Then, at each scale of the decoder of our \textit{MDeRainNet}, the SFE and the MDB are adopted to progressively estimate rain streaks from the input MPI. Skip-connections are also adopted to concatenate the encoded feature extracted at the same scale with the decoded feature to avoid losing meaningful image details and make our network converge fast. Pixel-unshuffling and pixel-shuffling operations~\cite{wang2022disentangling,shi2016real} are applied for feature downsampling and upsampling, respectively. Finally, at the LF Reshape stage, an SFE is used to further refine the input feature, and then the enhanced MPI feature is rearranged into the format of an array of sub-views. A $1\times 1$ convolution is also applied to squeeze the number of feature channels to $3$ in order to produce the rain streak (RGB) sub-views, termed as $R$. \hwj{The de-rained sub-views, \taore{termed} $B$, can be obtained by converting the input rainy MPI $I$ into \taore{an array of sub-views and followed by} subtracting $R$ from it.}

\hwj{To improve the generalization performance of our \textit{MDeRainNet} for real-world \taore{scenes}, a semi-supervised learning framework described in detail in Section~\ref{sec:Semi} is proposed, where multi-level KL divergence \taore{is utilized} to \taore{enforces} the feature distribution consistency between \taore{features of rain streaks of synthetic images and that of real-world rainy images}, and colored rain-free residue image guided contrastive regularization \taore{is introduced} to reconstruct the rain-free image.}

\subsection{Modified Disentangling Block (MDB)}

\begin{figure}[h]
\centering
\includegraphics[width=0.36\textwidth]{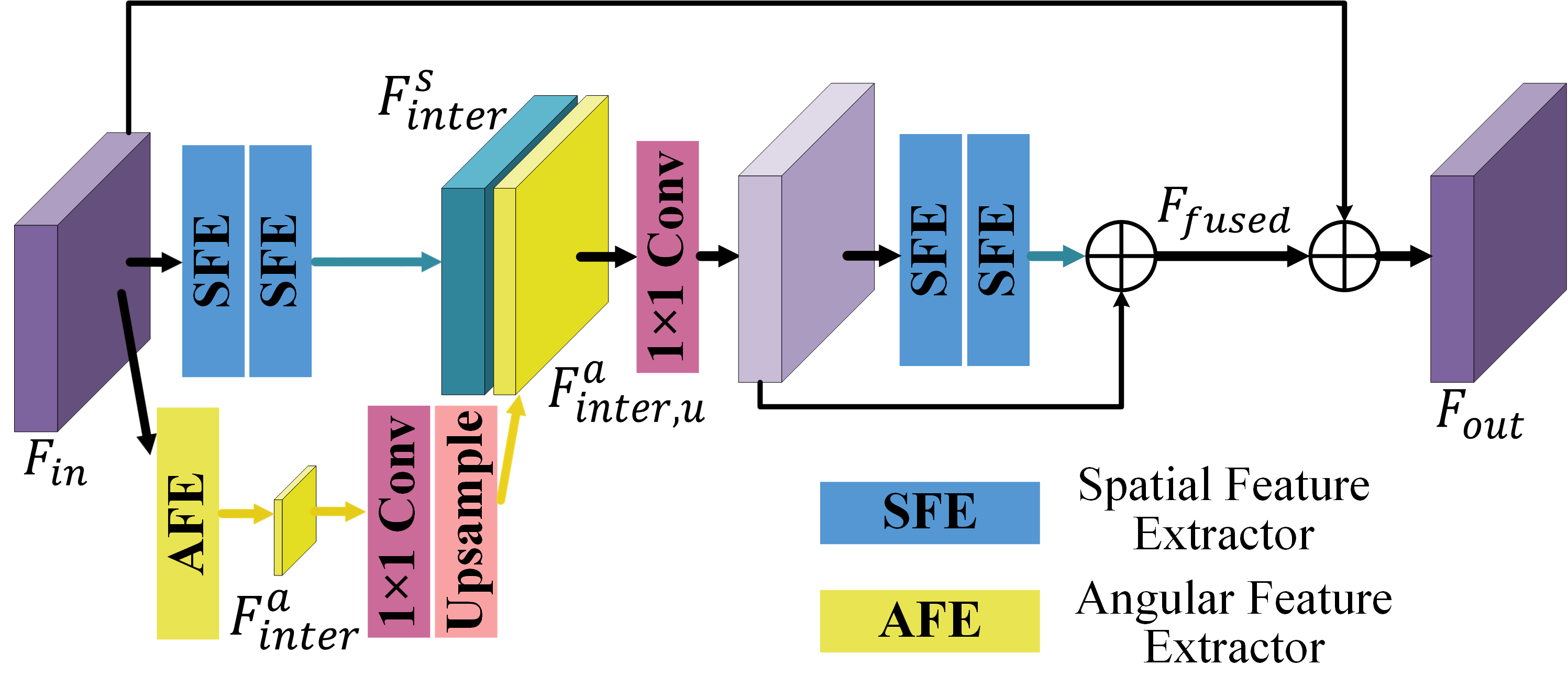} 
\caption{The structure of our Modified Disentangling Block (MDB).}
\label{fig:MDB}
\end{figure}

We introduced a modified disentangling block (MDB), as shown in Figure~\ref{fig:MDB}, for disentangling features of the input MPI.
The typical disentangling block~\cite{wang2022disentangling} proposed for disentangling features of an LF consists of an SFE, an AFE and an epipolar feature extractor (EFE) for extracting the spatial feature, the angular feature and the EPI feature from the MPI of an LFI, respectively.
Since AFE can extract angular information from all sub-views, while EFE can only extract angular information from EPIs, we believe that the angular information extracted by AFE would be more abundant than that extracted by EFE. 
Thus, our MDB discards the EFE, but extends the SFE and the AFE.

Specifically, as shown in Figure~\ref{fig:MDB}, given the MPI feature $F_{in} \in \mathbb{R}^{AH\times AW\times C'}$, a spatial feature extraction branch and an angular feature extraction branch of the MDB are used to disentangle the spatial and angular features, respectively. In the angular feature extraction branch, an AFE is first applied to generate an intermediate angular feature $F_{inter}^{a} \in \mathbb{R}^{H\times W\times \frac{C'}{2}}$. Then, the output feature $F_{inter,u}^{a} \in \mathbb{R}^{AH\times AW\times \frac{C'}{2}}$ can be obtained by upsampling this intermediate feature $F_{inter}^{a}$ by a factor of $A$, with a $1\times 1$ convolution layer and an upsample layer. 
Note that compared to the typical disentangling block in~\cite{wang2022disentangling}, the channel number of the angular feature $F_{inter,u}^{a}$ is doubled in our MDB.
Next, the feature $F_{inter,u}^{a}$ output by the angular feature extraction branch is concatenated with the intermediate spatial feature $F_{inter}^{s} \in \mathbb{R}^{AH\times AW\times C'}$ extracted by two SFEs of the spatial branch, and further fed into a $1\times 1$ convolution to reduce the channels.
A spatial residual block consisting of two SFEs and a skip connection is introduced to fuse $F_{inter,u}^{a}$ and $F_{inter}^{s}$ adaptively. 
Finally, the fusion feature, $F_{fused}$, is added to the input feature $F_{in}$ to obtain the enhanced feature $F_{out}$.

In our network, MDB is first used to extract shallow features at the beginning of the encoder. MDB is also placed behind each ESAI module to enhance features extracted by ESAIs.

\subsection{Extended Spatial-Angular Interaction (ESAI) Module}
\begin{figure}[h]
\centering
\includegraphics[width=0.4\textwidth]{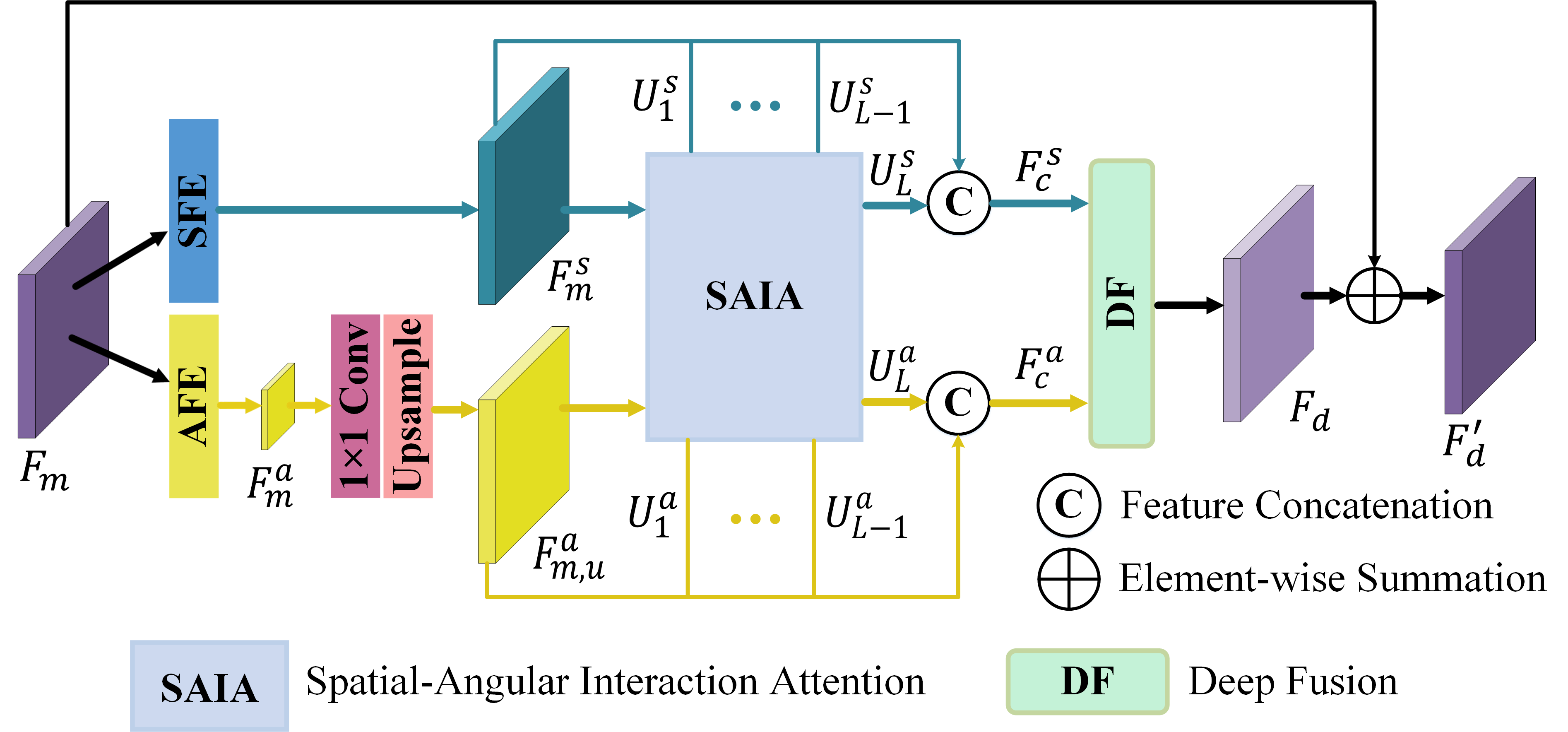} 
\caption{The structure of our proposed ESAI module.}
\label{fig:ESAI}
\end{figure}

We propose an ESAI module for interacting and fusing spatial feature and angular feature of LFIs, as shown in Figure~\ref{fig:ESAI}. Our ESAI contains two main blocks: a \textit{Spatial-Angular Interaction Attention} (SAIA) block and a \textit{Deep Fusion} (DF) block.

\begin{figure}[h]
    \centering
    \includegraphics[width=0.5\textwidth]{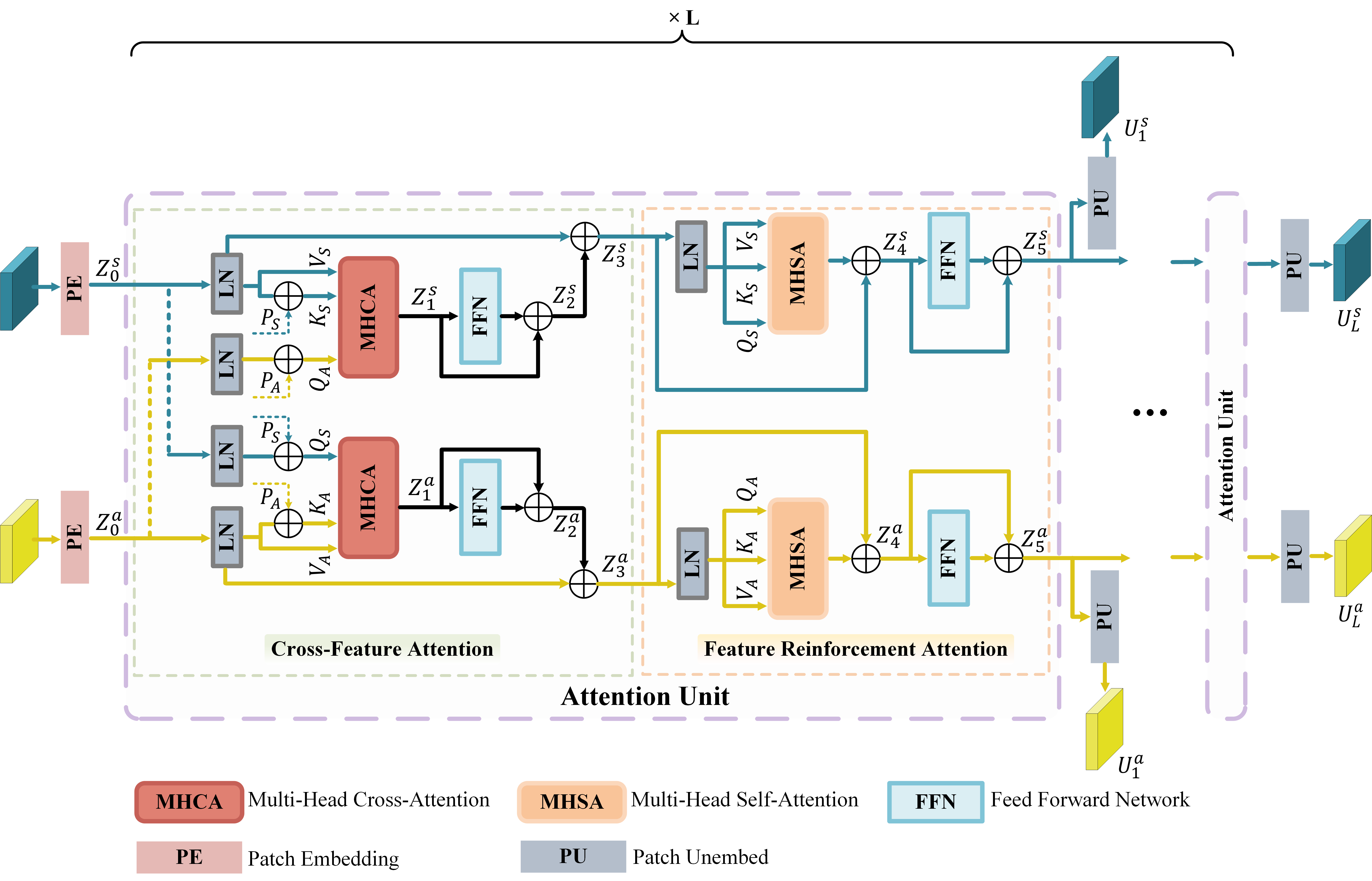} 
    \vspace{-4.5mm}
    \caption{The structure of SAIA (Spatial-Angular Interaction Attention) block.}
    \label{fig:SAIA}
\end{figure}

The input feature $F_m$ is first separated into the spatial feature $F_m^s$ and the angular feature $F_m^a$ via an SFE and an AFE. The angular feature $F_m^a$ is upsampled with a factor of $A$ to obtain the upsampled angular feature $F_{m,u}^a$. Then, $F_m^s$ and $F_{m,u}^a$ are fed into the SAIA block, passing through each attention unit in SAIA to make cross-feature attention process and feature reinforcement attention process. The spatial and angular features generated by the $i$-th attention unit of SAIA (as shown in Figure~\ref{fig:SAIA}), termed $U_i^s$ and $U_i^a$, are then concatenated with $F_m^s$ and $F_{m,u}^a$, respectively. Finally, the concatenated spatial feature $F_{c}^s$ and angular feature $F_{c}^a$ are fused by a DF block to obtain the fused feature $F_{d}$, which is later added to the input feature $F_m$ by a skip connection to obtain the final fused feature $F_{d}'$.

\textbf{SAIA Block.}
The SAIA block, as shown in Figure~\ref{fig:SAIA}, contains $L$ attention units, each consisting of two main components. First, the cross-feature attention based on the cross-attention mechanism~\cite{carion2020end,liu2021visual,cheng2021multimodal,lu2019vilbert,li2022transiam,chen2021key,he2021unimodal} is used to make interaction between the spatial feature and the angular feature. Second, the feature reinforcement attention based on the self-attention mechanism~\cite{vaswani2017attention,dosovitskiy2020image,chen2021pre} is executed to integrate the correlations between different features established by the cross-feature attention.

Following~\cite{wang2021pyramid}, the Patch Embedding (PE) of our SAIA block first splits the input spatial and angular features, $F_m^s \in \mathbb{R}^{H' \times W' \times C'}$ and $F_{m,u}^a \in \mathbb{R}^{H' \times W' \times C'}$, into patches with the spatial size of $P\times P$, as shown in Figures~\ref{fig:SAIA}.
Then, it reshapes each patch into a flattened 1-D feature, and a total of $N= \frac{H'W'}{P^2}$ flattened features can be obtained. 
Finally, the array of flattened features, termed $F_P$, is mapped to the latent $D$-dimensional embedding space via a trainable linear projection, which plays a role in feature transformation and avoids the flattened features being too large, as follows:
\begin{equation}
    Z_{0}^{s} = [ F^s_{p,1}E, F^s_{p,2}E,\dots ,F^s_{p,N}E], 
\end{equation}
\begin{equation}
    Z_{0}^{a} = [ F^a_{p,1}E, F^a_{p,2}E,\dots ,F^a_{p,N}E],
\end{equation}
where $E\in \mathbb{R}^{(P^2C') \times D}$ denotes the linear projection used for patch embedding. After completing the patch embedding, the features $Z_{0}^{s}$ and $Z_{0}^{a}$ are then fed into the cross-feature attention.

The cross-feature attention, as shown in Figure~\ref{fig:SAIA}, contains two branches, i.e., one spatial branch and one angular branch, each of which consists of a multi-head cross-attention (MHCA) and a feed-forward network (FFN). In the spatial branch, the features $Z_0^a$ and $Z_0^s$ pass through a layer normalization (LN)~\cite{ba2016layer} and are then directly added with the learnable angular position encoding $P_A$ and spatial position encoding $P_S$, respectively, to generate query $Q_A$ and key $K_S$. 
Here, the angular and spatial position encodings can make the global correlation of features calculated cross-feature attention stronger. 
$V_S$ is directly assigned as $Z_0^s$ processed by a layer normalization without adding the spatial position encoding $P_S$. Afterwards, the produced $Q_A$, $K_S$, and $V_S$ are fed into MHCA to learn the relationship between different spatial and angular tokens, as follows:
\begin{equation}
    Q_A= \mathrm{LN}(Z_0^a)+P_A,
\end{equation}
\begin{equation}
    K_S= \mathrm{LN}(Z_0^s)+P_S,
\end{equation}
\begin{equation}
    V_S= \mathrm{LN}(Z_0^s),
\end{equation}
\begin{equation}
    \mathrm{MHCA}(Q_A,K_S,V_S)= \mathrm{Concat}( h_{1}\cdots h_n)W_O, 
\end{equation}
where $n$ is the number of heads, and the embedding dimension of $K_S$, $V_S$ and $Q_A$ is split into $n$ groups.  
$h_{i}$ refers to the attention matrix calculated by the $i$-th head, and similar to~\cite{liang2022light,wang2022uformer}, it can be defined as:
\begin{equation}
     h_{i}=\mathrm{Softmax}(\frac{Q_{A,i}W_{Q,i}(K_{S,i}W_{K,i})^{T}}{\sqrt{d} })V_{S,i}W_{V,i},
\end{equation}
 where $W_{Q,i}$, $W_{K,i}$ and $W_{V,i} \in \mathbb{R}^{d \times d}$ denote the linear projection matrices of the queries, keys and values for the $i$-th head of MHCA, respectively.
$d=D/n$ is the feature dimension for each head.
$W_O \in \mathbb{R}^{D \times D}$ denotes the output projection matrix.

As shown in Figure~\ref{fig:SAIA}, the features passing through MHCA are further fed to an FFN for further feature transformation, which consists of an LN layer and a multi-layer perception (MLP) layer. In summary, the calculation formula of the spatial branch in the cross-feature attention can be written as:
\begin{equation}
    Z_1^s=\mathrm{MHCA}(Q_A,K_S,V_S),
\end{equation}
\begin{equation}
    Z_2^s =\mathrm{MLP}(\mathrm{LN}( Z_1^s))+Z_1^s,
\end{equation}
\begin{equation}\label{eq9}
    Z_3^s = Z_2^s + \mathrm{LN}(Z_0^s).
\end{equation}


The calculation of the angular branch is similar to that of the spatial branch, except that in the calculation of multi-head cross-attention, the key ($K_{A}$) and value ($V_{A}$) are calculated from the angular features, while the query ($Q_{S}$) is calculated from the spatial features. 
Through the above calculation process in the spatial and angular branches, the long-range dependencies between spatial and angular features can be modeled to obtain the cross-feature attention.



The calculation of feature reinforcement attention in the spatial branch is the same as that in the angular branch. We take the spatial branch as an example to describe it. The structure of our feature reinforcement attention is the same as a single Transformer layer, which consists of a multi-head self-attention (MHSA) and an FFN. First, the output of cross-feature attention is utilized to generate query $Q_{S}$, key $K_{S}$ and value $V_{S}$ via a LN, i.e., $Q_{S}=K_{S}=V_{S}=\mathrm{LN}(Z_{3}^{s})$. The process of the Feature Reinforcement Attention with an MHSA and an FFN can be described as:
\begin{equation}
    Z_4^s=\mathrm{MHSA}( Q_S,K_S,V_S)+Z_3^s,
\end{equation}
\begin{equation}
    Z_5^s =\mathrm{MLP} ( \mathrm{LN}( Z_4^s))+Z_4^s. 
\end{equation}

Similarly, the enhanced feature $Z_5^a$ can be obtained by the angular branch.

\begin{figure}[h]
\centering
\vspace{-3mm}
\includegraphics[width=0.3\textwidth]{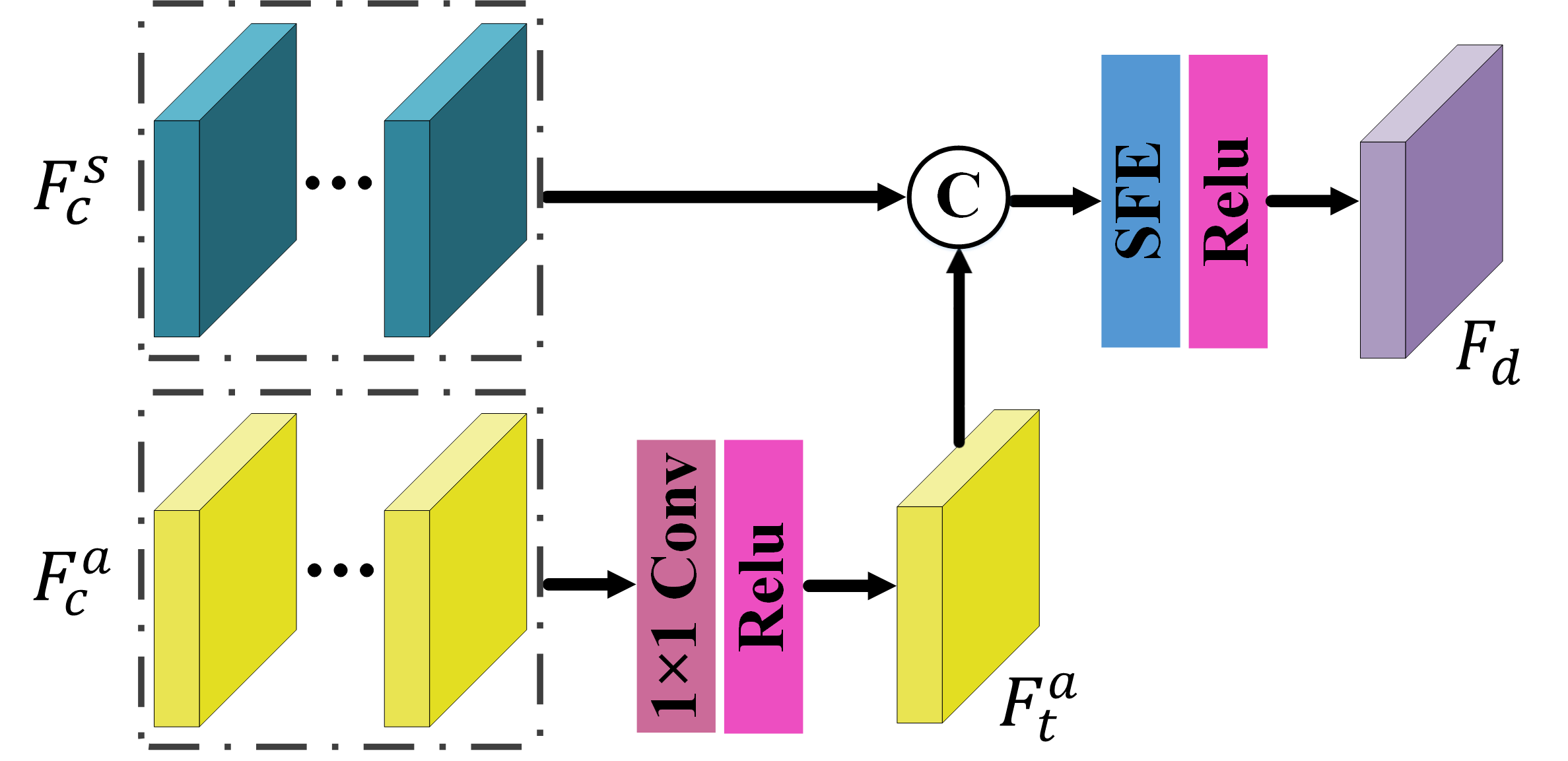}
\vspace{-3mm}
\caption{The structure of our Deep Fusion (DF) block. As the bottleneck of ESAI, DF is used to fully merge the spatial and angular features in ESAI.}
\label{fig:Deep Fusion}
\end{figure}

\textbf{Deep Fusion.} The output of the SAIA block is fed into the final deep fusion block to fully integrate the interacted features, as shown in Figure~\ref{fig:Deep Fusion}. Specifically, we first perform Patch Unembed (PU) on the features output by the $i$-th attention unit of SAIA  (such as $ Z_5^s$ and $Z_5^a$ output by the first attention unit) to obtain the spatial feature $U_i^s$ and the angular feature $U_i^a$. Then, $U_i^s$ and $U_i^a$ are concatenated with the input spatial and angular features, $F_m^s$ and $F_{m,u}^a$, respectively, to form the inputs of DF as follows:
\begin{equation}
    F_c^s = [F_{m}^{s},U_1^s,\cdots,U_L^s]\in\mathbb{R}^{H' \times W'\times (L+1)C'},
\end{equation}
\begin{equation}
    F_c^a = [F_{m,u}^a,U_1^a,\cdots,U_L^a]\in\mathbb{R}^{H' \times W'\times (L+1)C'}.
\end{equation}

Later, the concatenated angular feature, $F_c^a$, is fed into a $1\times 1$ convolution followed by a ReLU layer to reduce the channels and generate the intermediate feature $F_t^a \in\mathbb{R}^{H' \times W'\times C'}$. 
Finally, $F_t^a$ is concatenated with the spatial features $F_c^s$, and then the combined feature is further processed by an SFE and a ReLU layer to obtain the final fused feature, $F_d$.

\begin{figure*}[t]
	\centering
	\includegraphics[width=0.7\textwidth]{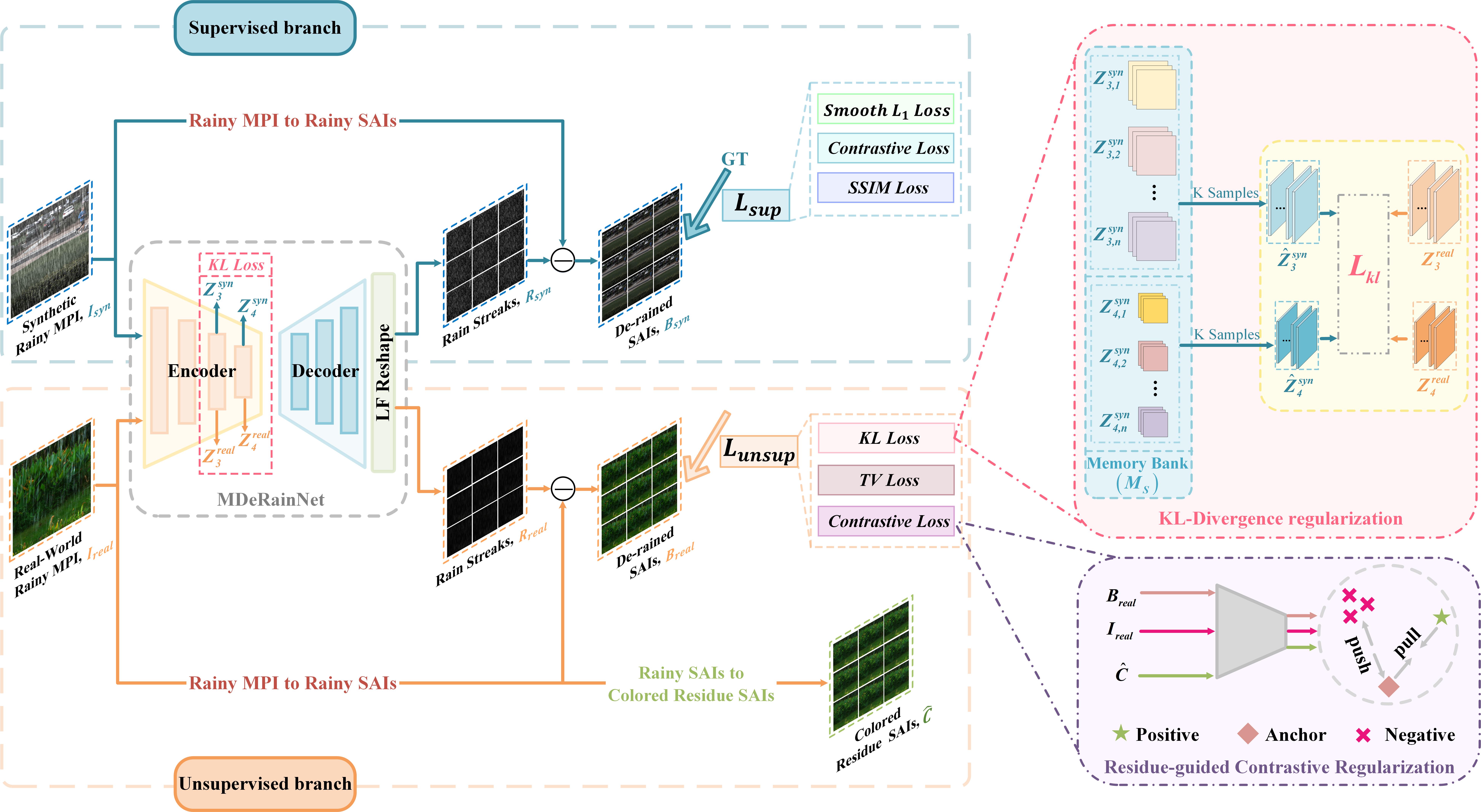} 
	\caption{An overview of \taore{our semi-supervised learning framework for our \textit{MDeRainNet} for improving its generalization on real-world rainy LF images.}}
	\label{fig:framework}\
 	\vspace{-4mm}
\end{figure*} 

\subsection{Semi-supervised Learning} \label{sec:Semi}
\hwj{We propose a novel semi-supervised learning framework \taore{to our network for improving its generalizatio on} real-world rainy images without ground-truths. As shown in Figure~\ref{fig:framework}, the semi-supervised framework consists of a supervised branch and an unsupervised branch, both of which share the \taore{same} rain removal network shown in Figure~\ref{fig:network}. Specifically, our network is trained with two stages. In the first
stage, supervised \taore{learning} is conducted on paired synthetic rainy LF images to obtain the best de-raining model on the synthetic datasets. In the second stage, the de-raining model obtained in the previous stage is further trained by utilizing both synthetic and real-world rainy LF images simultaneously. For the second phase of training, we \taore{utilize} a multi-level KL divergence regularization to enhance the feature distribution consistency between synthetic and real-world rain streaks, and \taore{introduce} a contrastive regularization guided by colored-residue images to achieve high-quality background restoration.} 

\textbf{Multi-level KL:} \hwj{In our task, the synthetic and real-world rain streak domains can be regarded as two different probability distributions. By modeling the closing process on the distribution probability of \taore{deep features extracted from synthetic LF images and that from real-world images}, the interdomain distance can be reduced. To achieve this goal, we use \taore{a multi-level KL-divergence regularization to align the deep features of rain streaks extracted from syntheci rainy LF images with that of real-world rainy LF images by the shared encoder at multiple scales to achieve distribution consistency.}} 


\hwj{Specifically, the shared encoder first extracts multi-scale rain streak features from synthetic and real-world rainy LFIs, and the extracted synthetic rain streak features $Z_{3}^{syn}$ and $Z_{4}^{syn}$ are stored in $M_{S}$. Here, $M_{S}$ is a memory bank that stores the latent space vectors at each level of the synthetic rain streaks to generate pseudo-GT for the real-world rain streaks, which is then used for KL divergence regularization. The creation of the pseudo-GT is based on the assumption that \taore{while projecting high-level features of real-world rain streaks to the latent space, they can be approximately represented as a weighted combination of high-level feature of synthetic rain streaks~\cite{yasarla2020syn2real}}. In $M_{S}$, there are two scales of rain streak features, termed $Z_{3,i}^{syn} \in \mathbb{R}^{\frac{AH}{4}\times \frac{AW}{4}\times C'}$ and $Z_{4,i}^{syn} \in \mathbb{R}^{\frac{AH}{8}\times \frac{AW}{8}\times C'}$, where $i$ represents $i$-th batch. Before each epoch of training in the second stage, the encoder of our network will extract latent features of rain streaks from the synthetic rainy LFIs to \taore{totally replace the features of synthetic rain streaks stored in} the memory bank.}

\hwj{Following~\cite{yasarla2020syn2real}, we use the top $k$ (set to $16$ by default) nearest synthetic rain streak \taore{features for} the real-world rain streak \taore{feature termed $Z^{real}$ to obtain the weighted combination $\hat{Z}^{syn}$}. Finally, $\hat{Z}^{syn}$ and $Z^{real}$ are \taore{used} to calculate the KL loss. The above process can be described as:}
\begin{eqnarray}
    \hat{Z}_{j}^{syn} =\sum_{i=1}^{k} \alpha_{i} Z_{j,i}^{syn}, Z_{j,i}^{syn}\in \mathrm{nearest}(Z_{j}^{real},M_{S},k), \\
    \mathcal{L}_{kl} =\sum_{j=3}^{4} \mathrm{KL}(\mathrm{Softmax}(\hat{Z}_{j}^{syn}),\mathrm{Softmax}(Z_{j}^{real})),
\end{eqnarray}
where $i$ \taore{is used to index} the $i$-th \taore{nearest feature for $Z^{real}$}. $j$ refers to the $j$-th scale of the encoder. $\alpha_{i}$ is the weighting coefficients. $\mathrm{nearest}(P, Q, N)$ is a function \taore{for finding} the top $N$ nearest neighbors of $P$ in $Q$. 

\textbf{Residue Image:}
\hwj{After narrowing the gap of the rain streak domain in the latent space, a contrastive regularization guided by rain-free residue image is introduced into the unsupervised branch to further bridge the gap of the clean background domain between synthetic and the real-world rainy images. The residue image was first proposed by Li~\etal~\cite{li2018robust}. It is a single-channel image, which is the residual result of the maximum channel value and the minimum channel value of the \taore{input} rainy image. Given a rainy LF image $I$ in the form of a SAI array, the residue image of $I$ can be defined as:
\begin{equation} \label{eq18}
   I_{k}^{r}(x)=\max_{i\in \left \{ R,G,B \right \} } I_{k}^{i}(x)- \min_{j\in \left \{ R,G,B \right \} } I_{k}^{j}(x),
\end{equation}
where $I_{k}^{i}$, $I_{k}^{j}$ denote the color channels of the $k$-th sub-view of $I$, $k\in \left \{ 1,2,..., A^{2} \right \} $. $x$ denotes the pixels of the sub-view. $A$ \taore{refers to} the angular dimension, and} \taore{$I^{r}$ represents the residue image.} The appearance of rain streaks in \taore{residue images} is colorless, i.e., $I^{r}$ is free of rain streaks and contains only a transformed version of the background details. In practice, we use colored-residue images~\cite{li2018robust} to achieve better de-raining performance in real-world rainy scenarios. Here, we take the $k$-th rainy sub-view as an example to introduce the acquisition process of its colored-residue image.

\hwj{For the $k$-th original \taore{rainy} sub-view $I_{k}$, \taore{we} first convert it to the YCbCr domain. Considering the fact that the impact of rain streaks on the image is mainly concentrated in the luminance channel (Y channel) while the chrominance channel (CbCr channels) is not affected much, we replace the Y channel with the single-channel residue image \taore{$I_{k}^{r}$} without rain streaks. Having obtained the values of all image intensities in the $I^{r}$CbCr domain, we convert them back to the RGB domain, in order to obtain the colored-residue image \taore{$C_{k}^{r}$}.
Though most of the background details \taore{are well preserved in} $C_{k}^{r}$, there is \taore{still} some color deviation compared to the original rainy sub-view. \taore{The reason can be explained that} when \taore{some background areas are} achromatic (i.e., white, gray or black), \taore{they will be} cancelled out along with the rain streaks \taore{as defined in} Equation~\ref{eq18}, resulting in some background areas being darkened. To alleviate this problem, we utilize a rain-free residue image-guided filter~\cite{li2019heavy} working on the original rainy sub-view to obtain another colored rain-free image $F_{k}^{r}$, which can be defined as:
\begin{equation}
   F_{k}^{r} = \mathrm{GF}(I_{k}, I_{k}^{r}),
\end{equation}
where $\mathrm{GF}(\cdot)$ is the guided filtering \taore{function}~\cite{li2019heavy}. 
The loss of details in $F_{k}^{r}$ is severe, but \taore{its} color fidelity is good.
\hwjre{Compared to $C_{k}^{r}$, $F_{k}^{r}$ can maintain better color consistency \taore{between the de-rained sub-views and} the \taore{input} rainy sub-views, but the loss of background details is more severe due to the filtering operation.} In order to preserve both color information and background \taore{textural} details, we perform a weighted summation of $C_{k}^{r}$ and $F_{k}^{r}$ to obtain our final colored-residue image. Thus, the colored-residue image of the $k$-th rainy sub-view can be expressed as:
\begin{equation} \label{eq19}
   \hat{C_{k}} = \mu \cdot C_{k}^{r}+ (1-\mu)\cdot F_{k}^{r},
\end{equation}
\hwjre{where $\mu$ is a hyper-parameter used to balance $C^r$ and $F^r$ for the formation of the final colored-residue image $\hat{C}$.} For \taore{more} details on the formation and visualization of colored-residue images used in our method, please refer to the Supplementary Material.}


\subsection{Loss Function}
\hwj{Based on our proposed semi-supervised learning framework, we use supervised and unsupervised losses to \wang{separately} train the corresponding branches. Next, we will introduce these two types of loss functions respectively.}

\textbf{Supervised Loss.} \hwj{In the supervised branch, we combine smooth $L_{1}$ loss~\cite{girshick2015fast}, SSIM loss, and contrastive regularization~\cite{wu2021contrastive} to constrain our network, and each one is designed for a specific purpose.} We first use a smooth $L_{1}$ loss between the de-rained image ($B$) and the ground truth ($G$), which is a good way to avoid the shortcomings of MSE and MAE loss, such as generating over-smoothed results and failing to distinguish the rain streaks and the background objects. Smooth $L_{1}$ loss is defined as follows: 
\begin{equation}
    \mathcal{L}_{smoothL_1} =
    \left\{
    \begin{array}{lcr}
    \begin{aligned}
    0.5(B- G) ^{2}    &&\mathrm{if} \ |B- G| <1 \\
    |B - G|-0.5 &&\mathrm{otherwise}.
    \end{aligned}
    \end{array}
    \right. 
\end{equation}
\hwj{Second, to make the network generate \taore{high-quality results}, we also adopt SSIM loss to constrain the structural similarities. The SSIM loss can be expressed as:
\begin{equation}
    \mathcal{L}_{ssim} = 1 - \mathrm{SSIM}(B, G).
\end{equation}
Third}, we employ contrastive regularization (CR) ~\cite{wu2021contrastive} to \hwj{further} improve the de-raining performance of our method. By adding a contrastive regularization term, our method is able to generate more natural de-rained images, and improve the visual quality of the restored images. 
CR is based on contrastive learning, which ensures that the predicted de-rained result is pulled closer to the clear image and pushed farther away from the rainy image in the representation space. Therefore, CR needs to construct `positive' and `negative' pairs and find the potential feature space of these pairs for contrast. In the LF image rain removal task, the positive pair is generated by the group of clean LFI ($G$) and de-rained LFI ($B$), and the negative pair is generated by the group of $B$ and a rainy LFI ($I$). The same fixed pre-trained model $\mathrm{M}(\cdot)$ (e.g., VGG-19~\cite{simonyan2014very}) is used as a latent feature space for common intermediate feature extraction. Therefore, CR can be formulated as:
\begin{equation}
    \mathcal{L}_{cr} = \sum_{k=1}^{n}w_{k}\cdot \frac{\mathrm{D}(\mathrm{M}_{k}(G), \mathrm{M}_{k}(B))}{\mathrm{D}(\mathrm{M}_{k}(I), \mathrm{M}_{k}(B))} ,
\end{equation}
where $\mathrm{M}_{k}$, $k=1,2,\dots ,n$ denotes extracting the $k$-th hidden features from the pre-trained model, $w_{k}$ is the weight coefficient, and $\mathrm{D}(\cdot)$ denotes the L1 distance.

\hwj{Finally, the supervised loss is obtained by adding the above three functions:
\begin{equation}\label{eq23}
   \mathcal{L}_{sup} =\mathcal{L}_{smoothL_{1}}+ \lambda _{ssim}\mathcal{L}_{ssim}+ \lambda _{cr}\mathcal{L}_{cr},
\end{equation}
where $\lambda _{ssim}$ and $\lambda _{cr}$ are two hyper-parameters used to control the \taore{contributions (weights) of} SSIM loss and CR.}

\textbf{Unsupervised Loss.} \hwj{For the unsupervised branch, we introduce multi-level KL loss, colored rain-free residue image-guided contrastive regularization (UCR) and total variation loss (TV loss) to constrain the real-world \taore{de-rained} image generation process.}
\hwj{In our UCR, the colored rain-free residue image ($\hat{C}$) is regarded as a ``positive sample". During the contrastive learning, the \taore{de-rained} image is close to $\hat{C}$ \taore{but} away from the input rain image in the feature spaces. Therefore, UCR can be defined as:
\begin{equation}
    \mathcal{L}_{ucr} = \sum_{k=1}^{n}w_{k}\cdot \frac{\mathrm{D}(\mathrm{M}_{k}(\hat{C}), \mathrm{M}_{k}(B))}{\mathrm{D}(\mathrm{M}_{k}(I), \mathrm{M}_{k}(B))} .
\end{equation}
}

\hwj{To further \taore{remove residual rain streaks} and preserve structures and details \taore{of} real-world rainy images, we \taore{adopt} a TV loss. The \taore{TV loss is defined} as follows:
\begin{equation}
    \mathcal{L}_{tv} = || \nabla_{h}B  ||_{1} + || \nabla_{v}B  ||_{1},
\end{equation}
where $\nabla_{h}$ and $\nabla_{v}$ denote the horizontal and vertical differential operators, respectively.
}

\hwj{Overall, the loss of unsupervised branch shown in Figure~\ref{fig:framework} can be expressed as:
\begin{equation}\label{eq26}
    \mathcal{L}_{unsup} =\alpha \mathcal{L}_{kl}+ \beta \mathcal{L}_{ucr}+ \lambda_{tv}\mathcal{L}_{tv},
\end{equation}
\wang{where $\alpha$, $\beta$, and $\lambda_{tv}$ are three hyper-parameters.}}

\textbf{Total Loss.} \hwj{To sum up, we combine supervised and unsupervised losses as our total loss, which can be described as:
\begin{equation}
   \mathcal{L}_{total} =\mathcal{L}_{sup}+ \mathcal{L}_{unsup}.
\end{equation}
}

\begin{table*}
\caption{De-raining results of different methods on the datasets RLFDB~\cite{ding2021rain} and RLMB~\cite{yan2023rain} were evaluated using PSNR and SSIM metrics. The best and second-best results are marked in \Best{red} and \SecondBest{cyan}, respectively. $\uparrow$ means the higher the better, and our method achieves the best performance on both datasets.}
\label{tab:Quantitative_Comparison}
\centering
\renewcommand{\arraystretch}{1.1}
\setlength{\tabcolsep}{4mm}{
\begin{tabular}{c|c|c|cc|cc}
\Xhline{1.2pt}
                               &      &        & \multicolumn{2}{c|}{RLFDB}  
                                       & \multicolumn{2}{c}{RLMB}                             \\ \cline{4-7} 
\multirow{-2}{*}{Input Image} & \multirow{-2}{*}{Methods} & \multirow{-2}{*}{Training Type} & PSNR$\uparrow$      & SSIM$\uparrow$                            & PSNR$\uparrow$              & SSIM$\uparrow$        \\ \hline
                               & PreNet~\cite{ren2019progressive}     & Supervised           & 30.13         & 0.942         & 29.69         & 0.954                  \\
                               & MSPFN~\cite{jiang2020multi}          & Supervised           & 25.10         & 0.836         & 22.92         & 0.749                  \\
                               & Syn2Real~\cite{yasarla2020syn2real}  & Semi-supervised           & 28.94         & 0.923         & 26.74         & 0.883                   \\
                               & RecDerain~\cite{ren2020single}       & Supervised           & 30.18         & \SecondBest{0.946}    & 30.24   & 0.957                  \\
                               & CCN~\cite{quan2021removing}          & Supervised           & 30.22         & 0.936         & 29.08         & 0.959                    \\
                               & MPRNet~\cite{zamir2021multi}         & Supervised           & 31.12         & 0.910         & 30.30         & 0.944                    \\
                               & DRT~\cite{liang2022drt}              & Supervised           & 25.56         & 0.812         & 29.58         & 0.929                     \\
                               & IDT~\cite{xiao2022image}             & Supervised           & 30.02         & 0.931  & \SecondBest{31.84}   & \SecondBest{0.960}                    \\
\multirow{-9}{*}{Single Image}          & Restormer~\cite{zamir2022restormer} & Supervised     & 30.33         & 0.914         & 30.74         & 0.951               \\ \hline
                               & LF-Internet~\cite{wang2020spatial}    & Supervised          & 26.11         & 0.829         & 28.05         & 0.919                        \\
                               & Ding~\etal~\cite{ding2021rain}        & Semi-supervised          & \SecondBest{31.81}  & 0.936         & 28.46     & 0.940                        \\
                               & DistgSSR~\cite{wang2022disentangling} & Supervised          & 28.18         & 0.897         & 29.85         & 0.948                \\
                               & SASRNet~\cite{liu2024adaptive} & Supervised          & 26.98         & 0.873         & 28.19         & 0.922                \\

                               & DSMNet~\cite{liu2023DSMNet} & Supervised          & 27.28         & 0.884         & 28.43         & 0.931                \\
                               
                               & Yan~\etal~\cite{yan2023rain}          & Semi-supervised          & 30.78         & 0.931         & 29.89         & 0.959                 \\
                               \cline{2-7} 
\multirow{-7}{*}{LF Image}           & Ours  & Semi-supervised   & \Best{32.19} & \Best{0.950} & \Best{32.27} & \Best{0.962} \\ \Xhline{1.2pt}
\end{tabular}}
\end{table*}
 


\section{Experiments and Analysis}
We compare our methodwith the state-of-the-art single image de-raining methods~\cite{ren2019progressive,jiang2020multi,yasarla2020syn2real,ren2020single,quan2021removing,zamir2021multi,liang2022drt,xiao2022image,zamir2022restormer} and LF image de-raining methods~\cite{wang2020spatial,ding2021rain,wang2022disentangling,yan2023rain,liu2024adaptive,liu2023DSMNet}. For comparing with the single image de-raining methods, every sub-view of an LFI is taken as a training sample. Since there are few methods for LFI de-raining, four LFI spatial super-resolution methods, LF-InterNet~\cite{wang2020spatial}, DistgSSR~\cite{wang2022disentangling}, SASRNet~\cite{liu2024adaptive}, and DSMNet~\cite{liu2023DSMNet}, which take the MPI of an LF image as input, are slightly modified by removing the upsampling operations and learned to recover rainy LF images. 

\subsection{Datasets}
To verify the effectiveness of our \textit{MDeRainNet}, we conduct experiment on two recently proposed light field rainy datasets~\cite{ding2021rain,yan2023rain}. Ding~\etal~\cite{ding2021rain} proposed the first LF rainy dataset, termed RLFDB. RLFDB contains $80$ synthetic and real-world rainy LF images, but there is no paired ground-truth for real-world scenes. In their synthetic rainy LF image generation approach, the particle system of Blender is adopted to simulate synthetic rain streaks with varying shapes, directions and densities, which are then superimposed onto real-world LFIs to produce realistic rain streak LFIs.
Later, Yan~\etal~\cite{yan2023rain} further considered the motion blur of rain streaks and global atmospheric light, so they proposed a new and large rainy LF image dataset, called RMLB, including $400$ synthetic rainy LFIs and $200$ real-world LF images captured by a Lytro ILLUM camera without ground-truth LF images.


\subsection{Implementation Details}
Our proposed \taore{network with the semi-supervised learning framework, called \textit{MDeRainNet},} is implemented with Pytorch, and all experiments are conducted on a PC with an NVIDIA GeForce RTX $3090$ GPU card.
Our network is trained using the Adam optimizer~\cite{kingma2014adam} for a total of $200$ epochs with the batch size set to $6$. The initial learning rate is $0.00004$, and the learning rate is reduced by a factor of $0.5$ every $45$ epochs. \hwj{In Equations~\ref{eq19}, ~\ref{eq23} and~\ref{eq26}, the hyper-parameters $\mu$, $\lambda_{ssim}$, $\lambda_{cr}$, $\lambda_{tv}$ are empirically set to $0.5$, $1$, $0.1$, and $0.000001$, respectively.} \wang{Inspired by~\cite{lin2020rain}, we do not set $\alpha$ and $\beta$ in Equation~\ref{eq26} to fixed values, but change them dynamically (please refer to our Supplementary Material to see more details).} In the training phase, we keep the angular resolution at $5\times5$ and crop each sub-view into patches of $64\times64$ with a stride of $32$. Then, an array of patches is reorganized into MPI pattern (the resolution of the MPI is $320\times320$) to form the input MPI of our network.  
In experiments, the peak signal-to-noise ratio (PSNR) and structural similarity (SSIM) metrics are used to evaluate the performance of all de-raining methods. 

\subsection{Comparison to the State-of-the-Arts}
A total of nine regular image de-raining methods~\cite{ren2019progressive,jiang2020multi,yasarla2020syn2real,ren2020single,quan2021removing,zamir2021multi,liang2022drt,xiao2022image,zamir2022restormer} and four LFI de-raing methods~\cite{wang2020spatial,ding2021rain,wang2022disentangling,yan2023rain, liu2024adaptive, liu2023DSMNet} are chosen to compare with our proposed \textit{MDeRainNet} by evaluating their performance on two public LF rainy datasets, RLFDB~\cite{ding2021rain} and RLMB~\cite{yan2023rain}.
The number of rainy LFIs in RLFDB~\cite{ding2021rain} is relative small and rain streaks are usually thin and sparse. In contrast, RLMB~\cite{yan2023rain} contains a large number of rainy LFIs, and the rain streaks are always large and dense.

\subsubsection{Quantitative Evaluation}
Table~\ref{tab:Quantitative_Comparison} reports the PSNR and SSIM values of different methods evaluated on two datasets. It can be seen that our method obviously outperforms all the other competing methods. The performances of the competing single image de-raining methods~\cite{ren2019progressive,jiang2020multi,yasarla2020syn2real,ren2020single,quan2021removing,zamir2021multi,liang2022drt,xiao2022image,zamir2022restormer} do not perform well. 
The shortcoming of such single image de-raining methods is exposed, while learning on a small sub-set. In another word, these single image de-raining methods rely more on the visual information of a single input image to recover the regions contaminated by rain streaks, which may lead to unnatural or unreliable de-rained results. In contrast, better rain removal results could be obtained by the LFI rain removal methods, since the LFI rain removal methods can fully exploit the complementary information of the input LFI rather than relying solely on the visual information of single images.
While testing on the RLMB dataset and measuring the results by PSNR metric, our method outperforms the single image rain removal methods IDT~\cite{xiao2022image} and Restormer~\cite{zamir2022restormer} by \hwj{$0.43$dB} and \hwj{$1.53$dB}, respectively, and also outperforms the LFI rain removal methods~\cite{yan2023rain, ding2021rain} by \hwj{$2.38$dB} and \hwj{$3.81$dB}, respectively. 

\renewcommand{\subwidth}{0.102}
\renewcommand{\ssubwidth}{0.051}
\begin{figure*}[h]
	\renewcommand{\tabcolsep}{0.8pt}
	\renewcommand\arraystretch{0.6}
	\begin{center}
		\begin{tabular}{cccccccccccccccccc}
            \multicolumn{2}{c}{\includegraphics[width=\subwidth\linewidth]{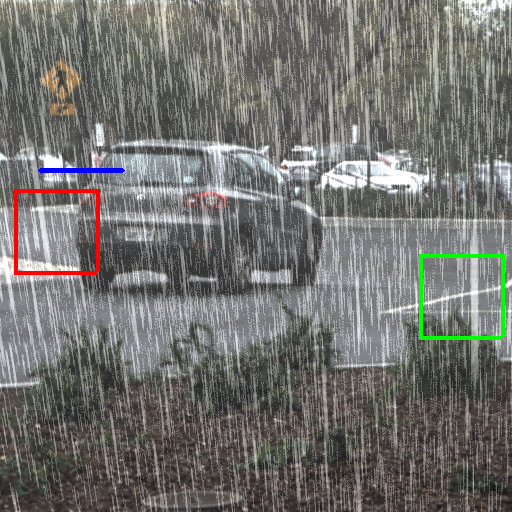}} &
            \multicolumn{2}{c}{\includegraphics[width=\subwidth\linewidth]{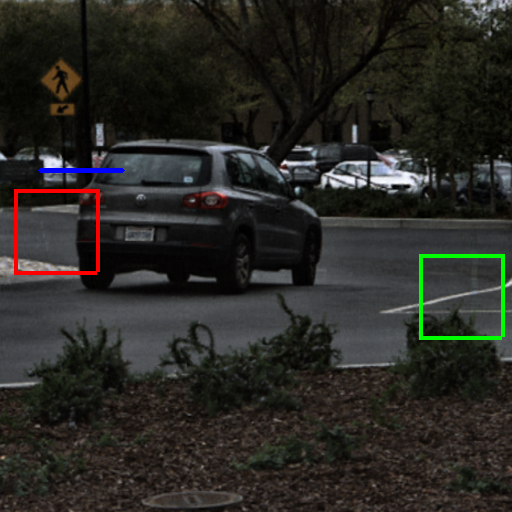}} &
            \multicolumn{2}{c}{\includegraphics[width=\subwidth\linewidth]{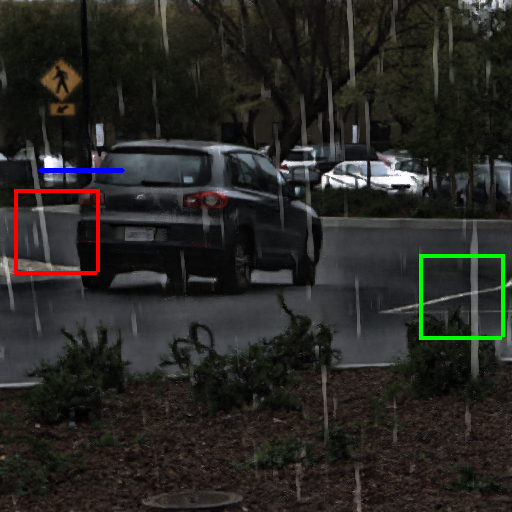}} &
            \multicolumn{2}{c}{\includegraphics[width=\subwidth\linewidth]{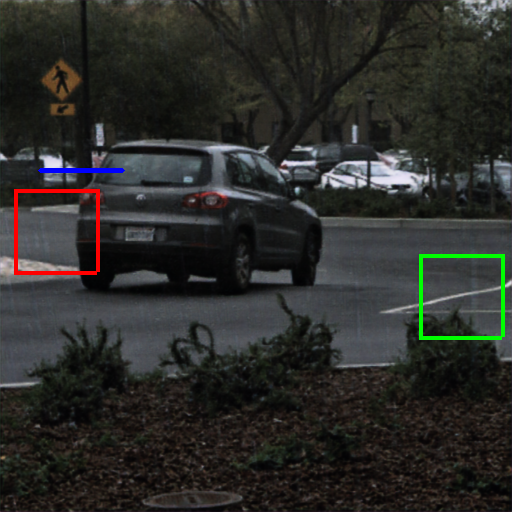}} &
            \multicolumn{2}{c}{\includegraphics[width=\subwidth\linewidth]{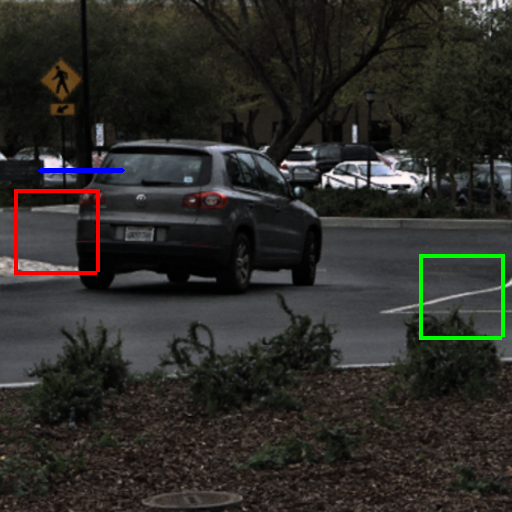}} &
            \multicolumn{2}{c}{\includegraphics[width=\subwidth\linewidth]{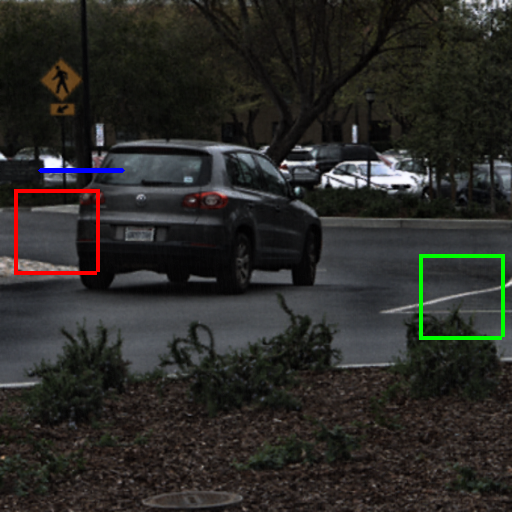}} &
            \multicolumn{2}{c}{\includegraphics[width=\subwidth\linewidth]{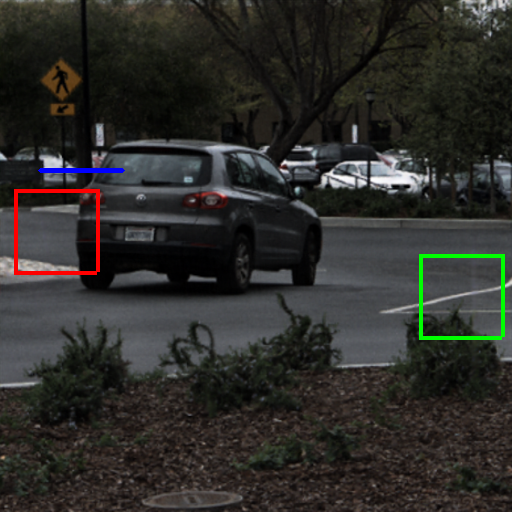}}&
            \multicolumn{2}{c}{\includegraphics[width=\subwidth\linewidth]{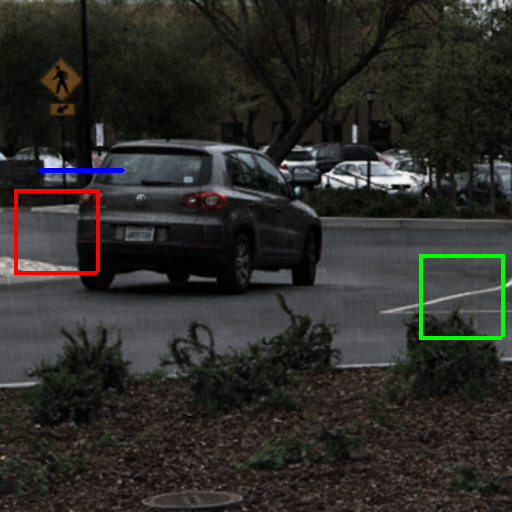}}&
            \multicolumn{2}{c}{\includegraphics[width=\subwidth\linewidth]{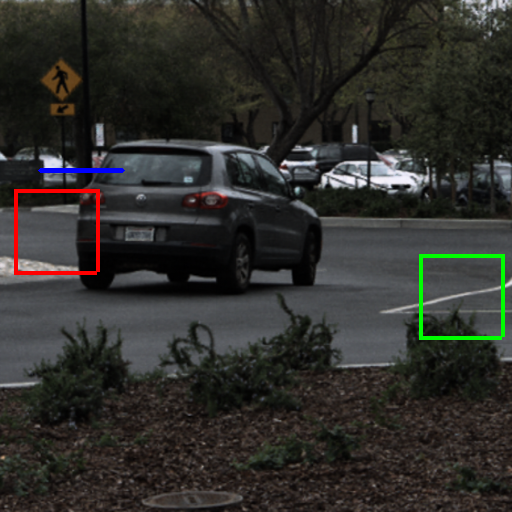}}\\

            \includegraphics[width=\ssubwidth\linewidth]{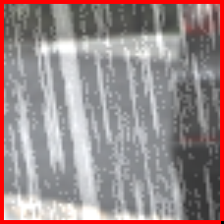} &
            \includegraphics[width=\ssubwidth\linewidth]{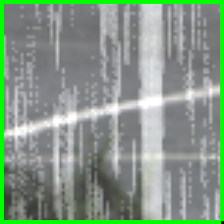} &
            \includegraphics[width=\ssubwidth\linewidth]{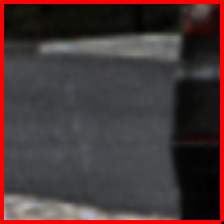} &
            \includegraphics[width=\ssubwidth\linewidth]{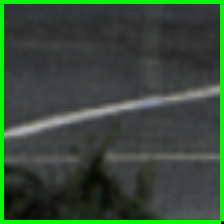} &
            \includegraphics[width=\ssubwidth\linewidth]{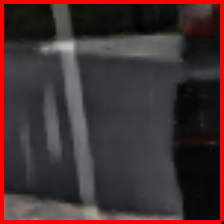} &
            \includegraphics[width=\ssubwidth\linewidth]{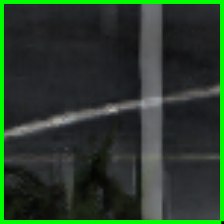} &
            \includegraphics[width=\ssubwidth\linewidth]{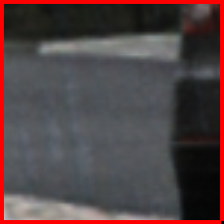} &
            \includegraphics[width=\ssubwidth\linewidth]{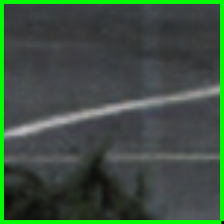} &
            \includegraphics[width=\ssubwidth\linewidth]{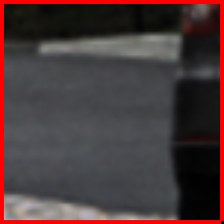} &
            \includegraphics[width=\ssubwidth\linewidth]{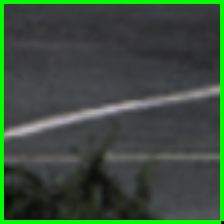} &
            \includegraphics[width=\ssubwidth\linewidth]{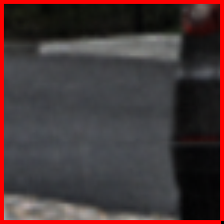} &
            \includegraphics[width=\ssubwidth\linewidth]{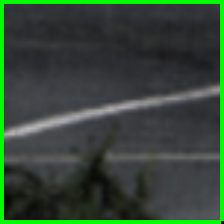} &
            \includegraphics[width=\ssubwidth\linewidth]{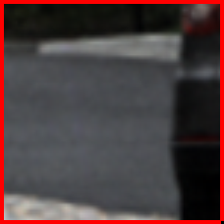} &
            \includegraphics[width=\ssubwidth\linewidth]{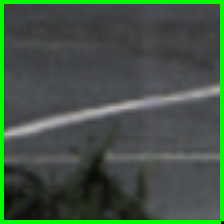} &
            \includegraphics[width=\ssubwidth\linewidth]{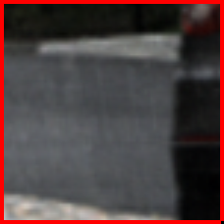} &
            \includegraphics[width=\ssubwidth\linewidth]{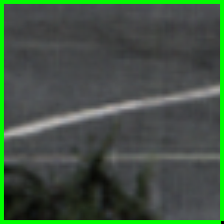} &
            \includegraphics[width=\ssubwidth\linewidth]{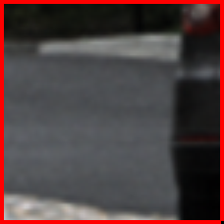} &
            \includegraphics[width=\ssubwidth\linewidth]{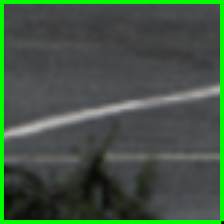}\\

            \multicolumn{2}{c}{\includegraphics[width=\subwidth\textwidth,height=0.012\textwidth]{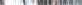}} &
            \multicolumn{2}{c}{\includegraphics[width=\subwidth\textwidth,height=0.012\textwidth]{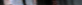}} &
            \multicolumn{2}{c}{\includegraphics[width=\subwidth\textwidth,height=0.012\textwidth]{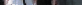}} &
            \multicolumn{2}{c}{\includegraphics[width=\subwidth\textwidth,height=0.012\textwidth]{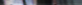}} &
            \multicolumn{2}{c}{\includegraphics[width=\subwidth\textwidth,height=0.012\textwidth]{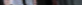}} &
            \multicolumn{2}{c}{\includegraphics[width=\subwidth\textwidth,height=0.012\textwidth]{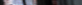}} &
            \multicolumn{2}{c}{\includegraphics[width=\subwidth\textwidth,height=0.012\textwidth]{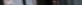}}&
            \multicolumn{2}{c}{\includegraphics[width=\subwidth\textwidth,height=0.012\textwidth]{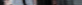}}&
            \multicolumn{2}{c}{\includegraphics[width=\subwidth\textwidth,height=0.012\textwidth]{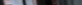}}\\

            \multicolumn{2}{c}{\scriptsize{Input}}&
            \multicolumn{2}{c}{\scriptsize{PreNet~\cite{ren2019progressive}}} &
            \multicolumn{2}{c}{\scriptsize{MSPFN~\cite{jiang2020multi}}} &
            \multicolumn{2}{c}{\scriptsize{Syn2Real~\cite{yasarla2020syn2real}}} &
            \multicolumn{2}{c}{\scriptsize{RecDerain~\cite{ren2020single}}} &
            \multicolumn{2}{c}{\scriptsize{CCN~\cite{quan2021removing}}} &
            \multicolumn{2}{c}{\scriptsize{MPRNet~\cite{zamir2021multi}}} &
            \multicolumn{2}{c}{\scriptsize{DRT~\cite{liang2022drt}}}&
            \multicolumn{2}{c}{\scriptsize{IDT~\cite{xiao2022image}}}\\

            \multicolumn{2}{c}{\scriptsize{10.60/0.220}}&
            \multicolumn{2}{c}{\scriptsize{31.75/0.973}} &
            \multicolumn{2}{c}{\scriptsize{23.30/0.774}} &
            \multicolumn{2}{c}{\scriptsize{29.96/0.940}} &
            \multicolumn{2}{c}{\scriptsize{32.14/0.976}} &
            \multicolumn{2}{c}{\scriptsize{32.29/0.978}} &
            \multicolumn{2}{c}{\scriptsize{34.66/0.971}} &
            \multicolumn{2}{c}{\scriptsize{33.18/0.953}}&
            \multicolumn{2}{c}{\scriptsize{\SecondBest{38.19}/\SecondBest{0.985}}}\\

            \multicolumn{2}{c}{\includegraphics[width=\subwidth\linewidth]{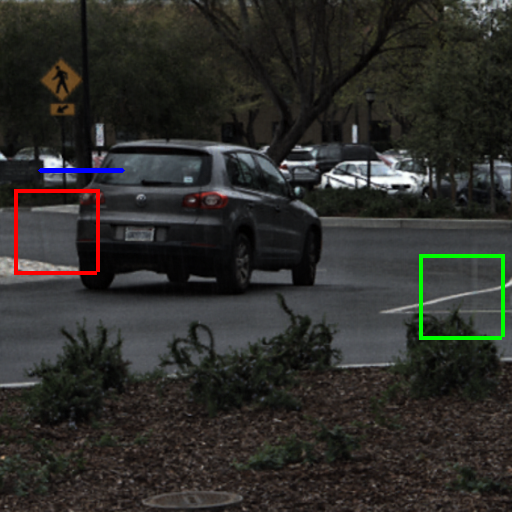}} &
            \multicolumn{2}{c}{\includegraphics[width=\subwidth\linewidth]{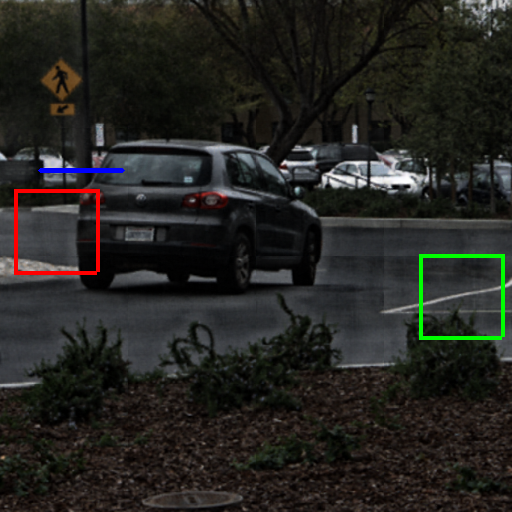}} &
            \multicolumn{2}{c}{\includegraphics[width=\subwidth\linewidth]{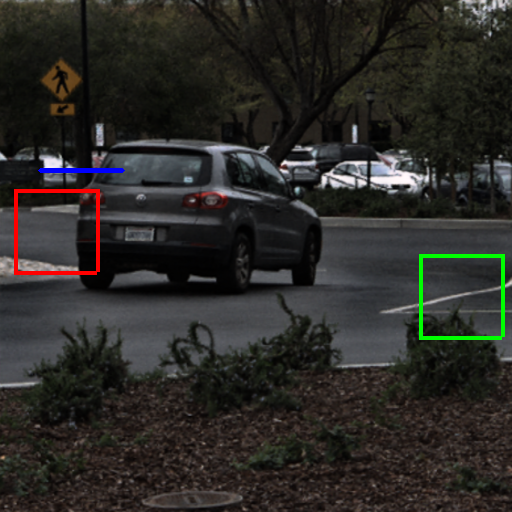}} &
            \multicolumn{2}{c}{\includegraphics[width=\subwidth\linewidth]{figure/comparison/DRT_NTIRE2022/504_mark.png}} &
            \multicolumn{2}{c}{\includegraphics[width=\subwidth\linewidth]{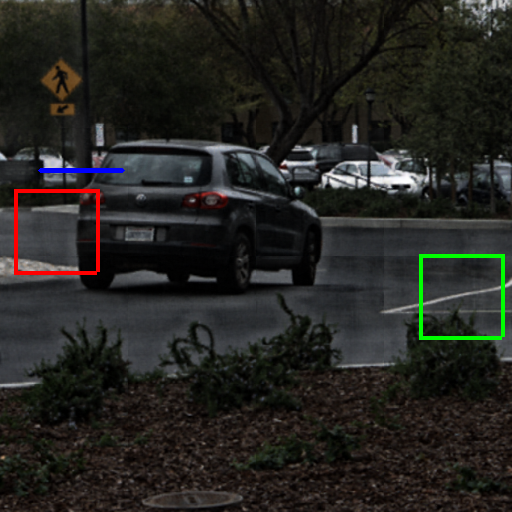}} &
            \multicolumn{2}{c}{\includegraphics[width=\subwidth\linewidth]{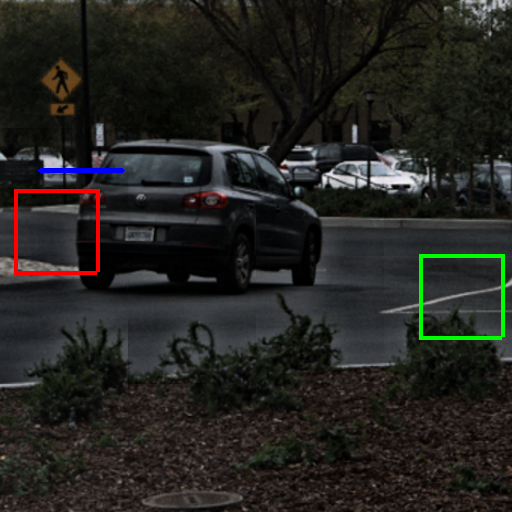}} &
            \multicolumn{2}{c}{\includegraphics[width=\subwidth\linewidth]{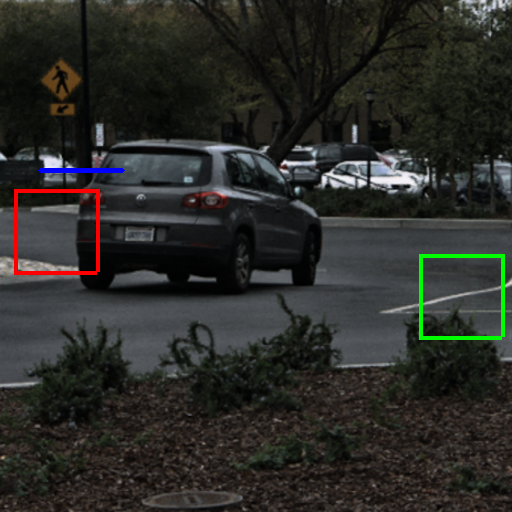}} &
            \multicolumn{2}{c}{\includegraphics[width=\subwidth\linewidth]{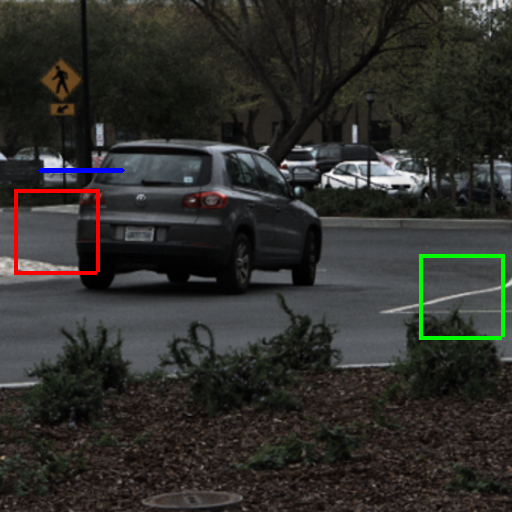}} &
            \multicolumn{2}{c}{\includegraphics[width=\subwidth\linewidth]{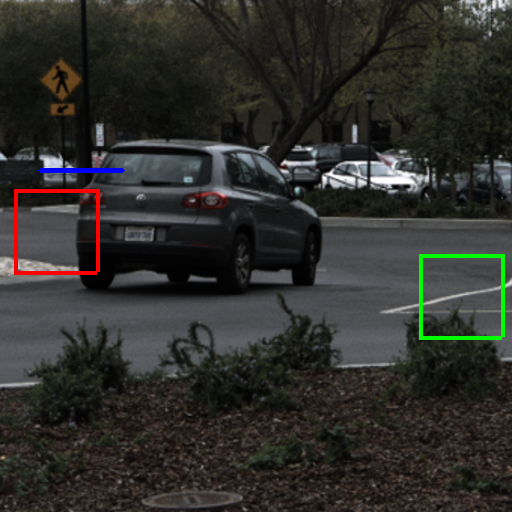}} \\

            \includegraphics[width=\ssubwidth\linewidth]{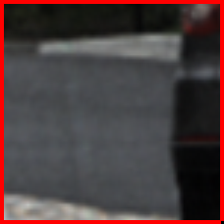} &
            \includegraphics[width=\ssubwidth\linewidth]{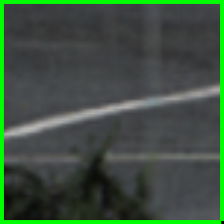} &
            \includegraphics[width=\ssubwidth\linewidth]{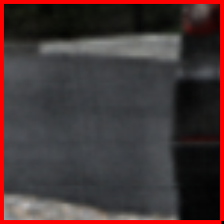} &
            \includegraphics[width=\ssubwidth\linewidth]{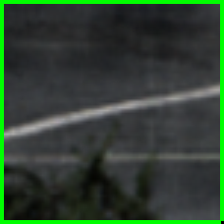} &
            \includegraphics[width=\ssubwidth\linewidth]{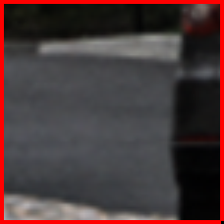} &
            \includegraphics[width=\ssubwidth\linewidth]{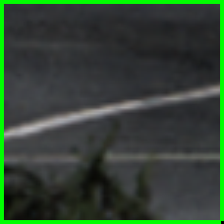} &
            \includegraphics[width=\ssubwidth\linewidth]{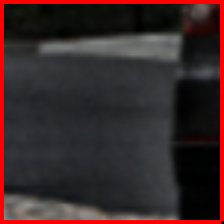} &
            \includegraphics[width=\ssubwidth\linewidth]{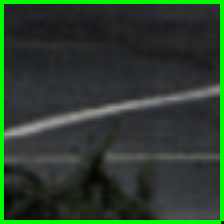} &
             \includegraphics[width=\ssubwidth\linewidth]{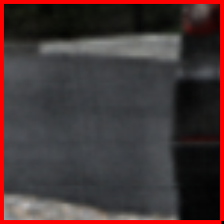} &
            \includegraphics[width=\ssubwidth\linewidth]{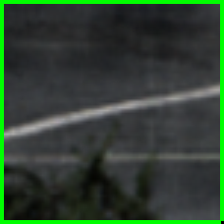} &
             \includegraphics[width=\ssubwidth\linewidth]{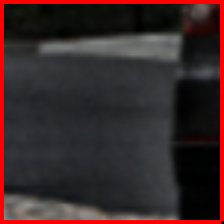} &
            \includegraphics[width=\ssubwidth\linewidth]{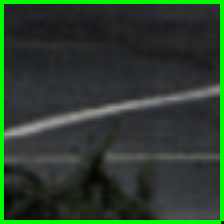} &
            \includegraphics[width=\ssubwidth\linewidth]{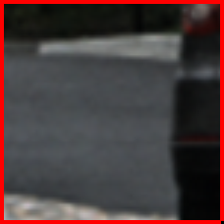} &
            \includegraphics[width=\ssubwidth\linewidth]{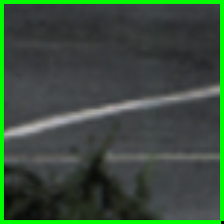} &
            \includegraphics[width=\ssubwidth\linewidth]{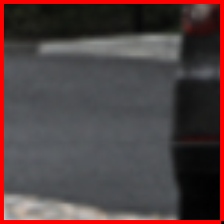} &
            \includegraphics[width=\ssubwidth\linewidth]{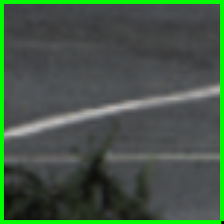} &
            \includegraphics[width=\ssubwidth\linewidth]{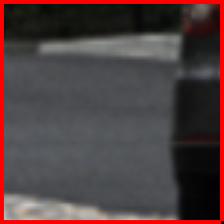} &
            \includegraphics[width=\ssubwidth\linewidth]{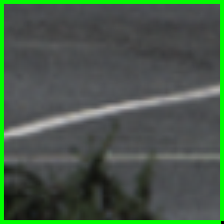}  \\

            \multicolumn{2}{c}{\includegraphics[width=\subwidth\textwidth,height=0.012\textwidth]{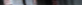}} &
            \multicolumn{2}{c}{\includegraphics[width=\subwidth\textwidth,height=0.012\textwidth]{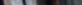}} &
            \multicolumn{2}{c}{\includegraphics[width=\subwidth\textwidth,height=0.012\textwidth]{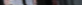}} &
            \multicolumn{2}{c}{\includegraphics[width=\subwidth\textwidth,height=0.012\textwidth]{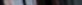}} &
            \multicolumn{2}{c}{\includegraphics[width=\subwidth\textwidth,height=0.012\textwidth]{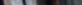}} &
            \multicolumn{2}{c}{\includegraphics[width=\subwidth\textwidth,height=0.012\textwidth]{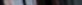}} &
            \multicolumn{2}{c}{\includegraphics[width=\subwidth\textwidth,height=0.012\textwidth]{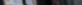}} &
            \multicolumn{2}{c}{\includegraphics[width=\subwidth\textwidth,height=0.012\textwidth]{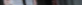}}&
            \multicolumn{2}{c}{\includegraphics[width=\subwidth\textwidth,height=0.012\textwidth]{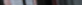}}  \\

            \multicolumn{2}{c}{\scriptsize{Restormer~\cite{zamir2022restormer}}}&
            \multicolumn{2}{c}{\scriptsize{LF-Internet~\cite{wang2020spatial}}}&
            \multicolumn{2}{c}{\scriptsize{Ding~\etal~\cite{ding2021rain}}}&
            \multicolumn{2}{c}{\scriptsize{DistgSSR~\cite{wang2022disentangling}}}&
            \multicolumn{2}{c}{\scriptsize{SASRNet~\cite{liu2024adaptive}}}&
            \multicolumn{2}{c}{\scriptsize{DSMNet~\cite{liu2023DSMNet}}}&
            \multicolumn{2}{c}{\scriptsize{Yan~\etal~\cite{yan2023rain}}}&
            \multicolumn{2}{c}{\scriptsize{MDeRainNet}}&
            \multicolumn{2}{c}{\scriptsize{GT}}\\

            \multicolumn{2}{c}{\scriptsize{32.71/0.976}} &
            \multicolumn{2}{c}{\scriptsize{25.10/0.908}} &
            \multicolumn{2}{c}{\scriptsize{29.69/0.969}} &
            \multicolumn{2}{c}{\scriptsize{24.09/0.917}} &
            \multicolumn{2}{c}{\scriptsize{24.39/0.910}} &
            \multicolumn{2}{c}{\scriptsize{24.58/0.918}} &
            \multicolumn{2}{c}{\scriptsize{33.66/0.970}} &
            \multicolumn{2}{c}{\scriptsize{\Best{38.82}/\Best{0.988}}} &
            \multicolumn{2}{c}{\scriptsize{$+\infty$/1}}\\

		\end{tabular}
	\end{center}
	\vspace{-0.016\textwidth}
    \caption{De-raining results of different de-raining methods on synthetic LFIs from RLMB~\cite{yan2023rain}. The de-rained central sub-view as well as the zoomed-in patches and an EPI produced by each method are shown. The horizontal EPIs extracted along the blue lines across the sub-views in the same row are also visualized to show the angular (image content) consistency between different sub-views. The PSNR and SSIM scores obtained by different methods on the presented scenes are reported under the zoomed-in area. The best value is colored in \Best{red}, and the second best value is colored in \SecondBest{cyan}.} 
	\label{fig:RLFDB_synthetic_result}
        \vspace{-0.016\textwidth}
\end{figure*}

\renewcommand{\subwidth}{0.13}
\renewcommand{\ssubwidth}{0.064}
\begin{figure*}[t]
	\renewcommand{\tabcolsep}{0.8pt}
	\renewcommand\arraystretch{0.6}
	\begin{center}
		\begin{tabular}{cccccccccccccc}

            \multicolumn{2}{c}{\includegraphics[width=\subwidth\linewidth]{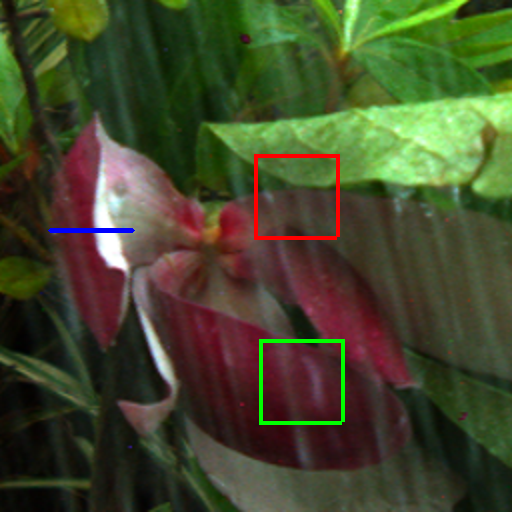}} &
            \multicolumn{2}{c}{\includegraphics[width=\subwidth\linewidth]{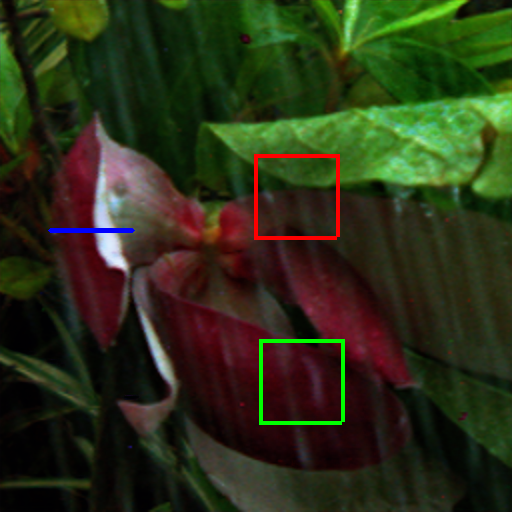}} &
            \multicolumn{2}{c}{\includegraphics[width=\subwidth\linewidth]{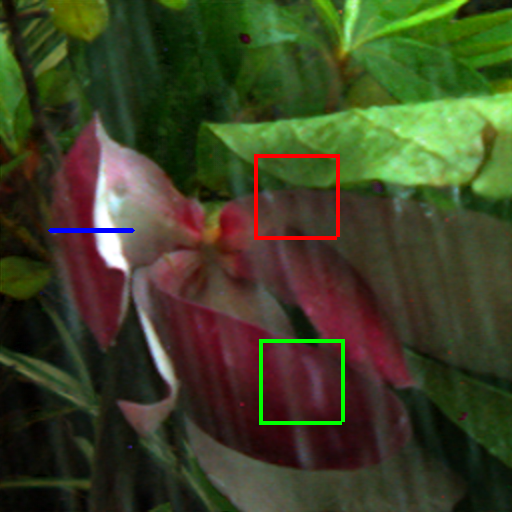}} &
            \multicolumn{2}{c}{\includegraphics[width=\subwidth\linewidth]{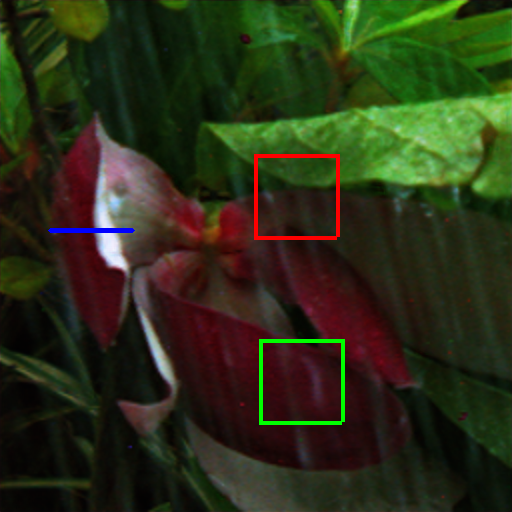}} &
            \multicolumn{2}{c}{\includegraphics[width=\subwidth\linewidth]{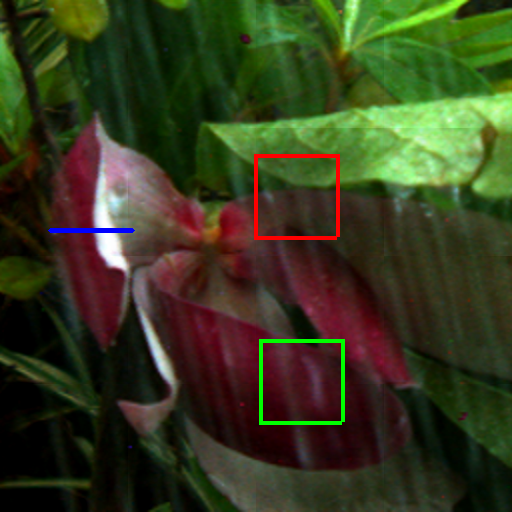}} &
            \multicolumn{2}{c}{\includegraphics[width=\subwidth\linewidth]{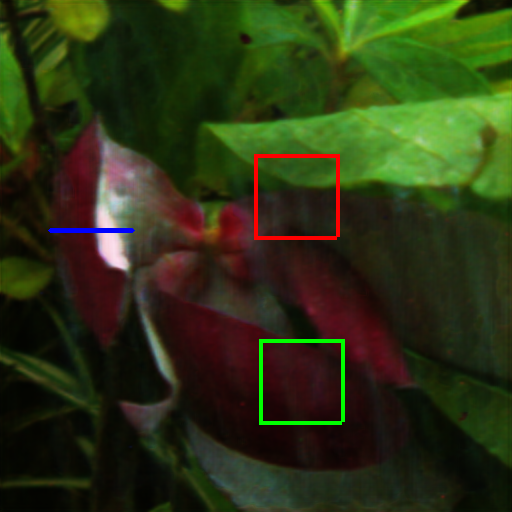}} &
            \multicolumn{2}{c}{\includegraphics[width=\subwidth\linewidth]{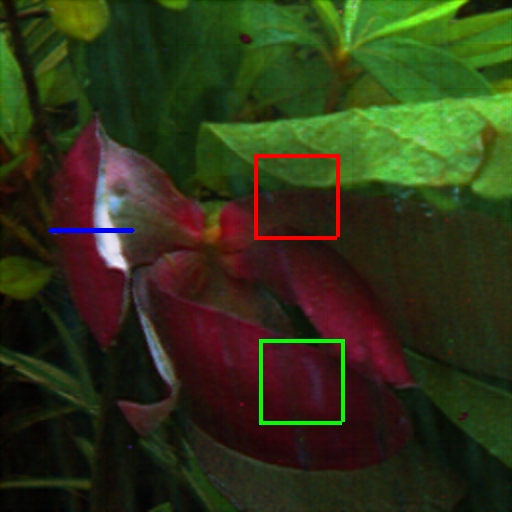}} \\

            \includegraphics[width=\ssubwidth\linewidth]{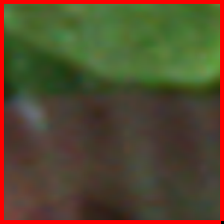} &
            \includegraphics[width=\ssubwidth\linewidth]{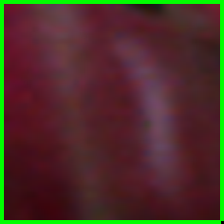} &
            \includegraphics[width=\ssubwidth\linewidth]{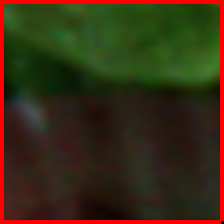} &
            \includegraphics[width=\ssubwidth\linewidth]{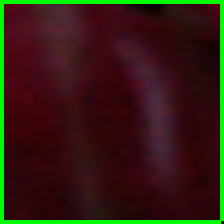} &
            \includegraphics[width=\ssubwidth\linewidth]{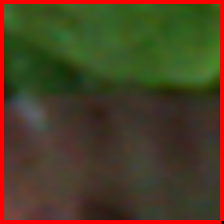} &
            \includegraphics[width=\ssubwidth\linewidth]{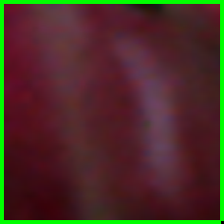} &
            \includegraphics[width=\ssubwidth\linewidth]{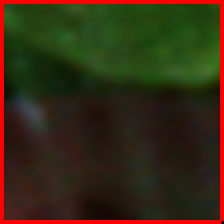} &
            \includegraphics[width=\ssubwidth\linewidth]{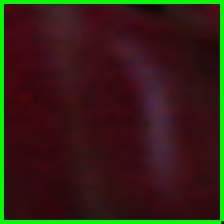} &
            \includegraphics[width=\ssubwidth\linewidth]{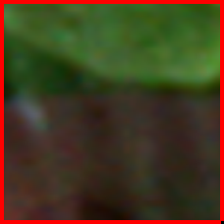} &
            \includegraphics[width=\ssubwidth\linewidth]{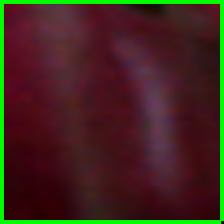} &
            \includegraphics[width=\ssubwidth\linewidth]{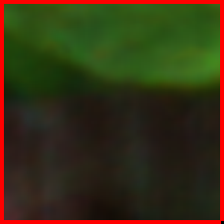} &
            \includegraphics[width=\ssubwidth\linewidth]{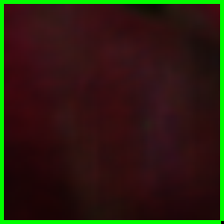} &
            \includegraphics[width=\ssubwidth\linewidth]{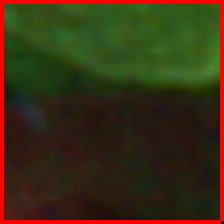} &
            \includegraphics[width=\ssubwidth\linewidth]{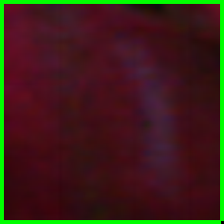} \\

            \multicolumn{2}{c}{\includegraphics[width=0.1326\textwidth,height=0.013\textwidth]{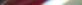}} &
            \multicolumn{2}{c}{\includegraphics[width=0.1326\textwidth,height=0.013\textwidth]{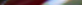}} &
            \multicolumn{2}{c}{\includegraphics[width=0.1326\textwidth,height=0.013\textwidth]{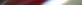}} &
            \multicolumn{2}{c}{\includegraphics[width=0.1326\textwidth,height=0.013\textwidth]{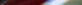}} &
            \multicolumn{2}{c}{\includegraphics[width=0.1326\textwidth,height=0.013\textwidth]{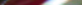}} &
            \multicolumn{2}{c}{\includegraphics[width=0.1326\textwidth,height=0.013\textwidth]{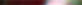}} &
            \multicolumn{2}{c}{\includegraphics[width=0.1326\textwidth,height=0.013\textwidth]{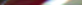}} \\

            \multicolumn{2}{c}{\scriptsize{Input}}&
            \multicolumn{2}{c}{\scriptsize{Syn2Real~\cite{yasarla2020syn2real}}} & 
            \multicolumn{2}{c}{\scriptsize{Restormer~\cite{zamir2022restormer}}}&
            \multicolumn{2}{c}{\scriptsize{Ding~\etal~\cite{ding2021rain}}}&
            \multicolumn{2}{c}{\scriptsize{DistgSSR~\cite{wang2022disentangling}}}&
            \multicolumn{2}{c}{\scriptsize{Yan~\etal~\cite{yan2023rain}}}&
            \multicolumn{2}{c}{\scriptsize{MDeRainNet}}\\

            \multicolumn{2}{c}{\includegraphics[width=\subwidth\linewidth]{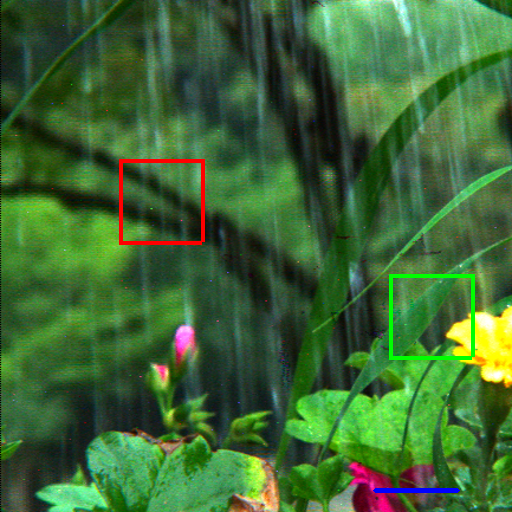}} &
            \multicolumn{2}{c}{\includegraphics[width=\subwidth\linewidth]{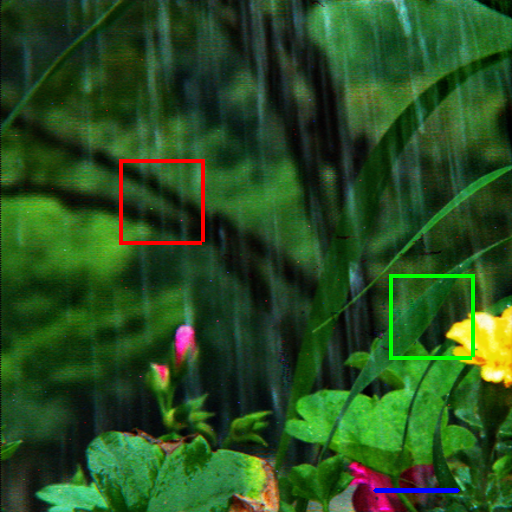}} &
            \multicolumn{2}{c}{\includegraphics[width=\subwidth\linewidth]{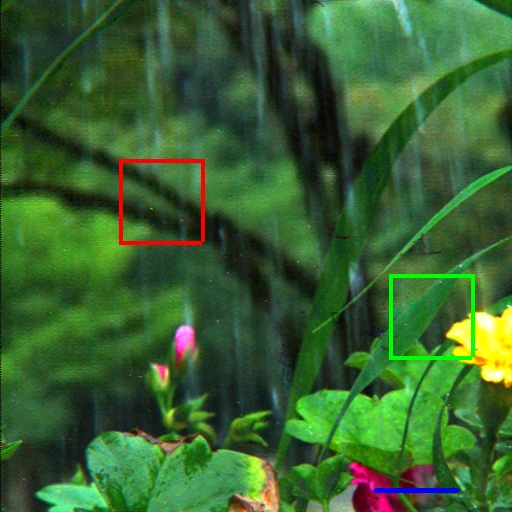}} &
            \multicolumn{2}{c}{\includegraphics[width=\subwidth\linewidth]{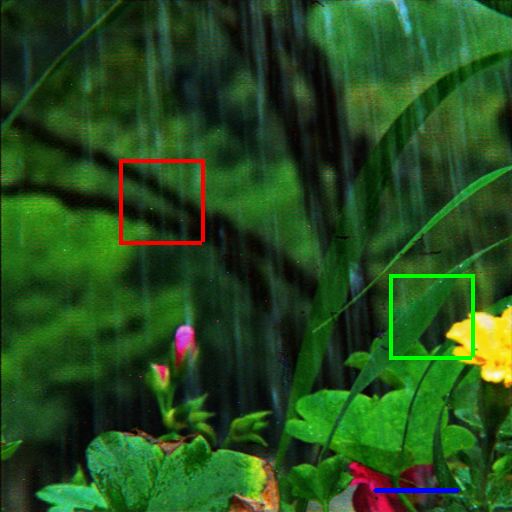}} &
            \multicolumn{2}{c}{\includegraphics[width=\subwidth\linewidth]{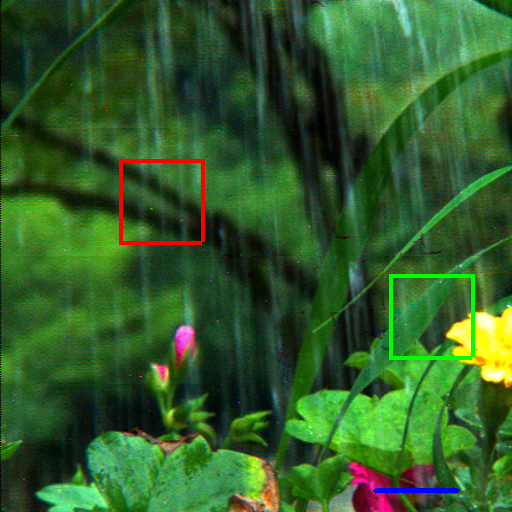}} &
            \multicolumn{2}{c}{\includegraphics[width=\subwidth\linewidth]{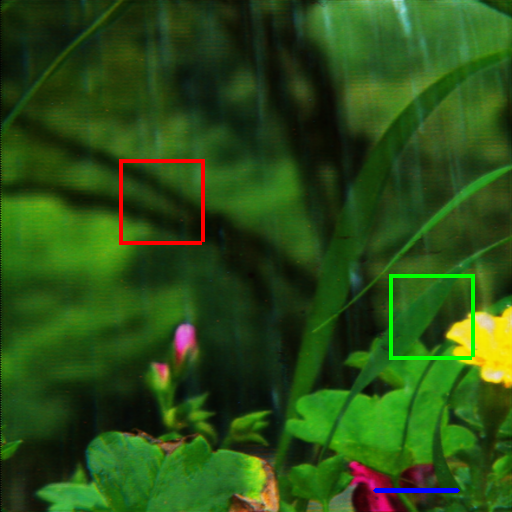}} &
            \multicolumn{2}{c}{\includegraphics[width=\subwidth\linewidth]{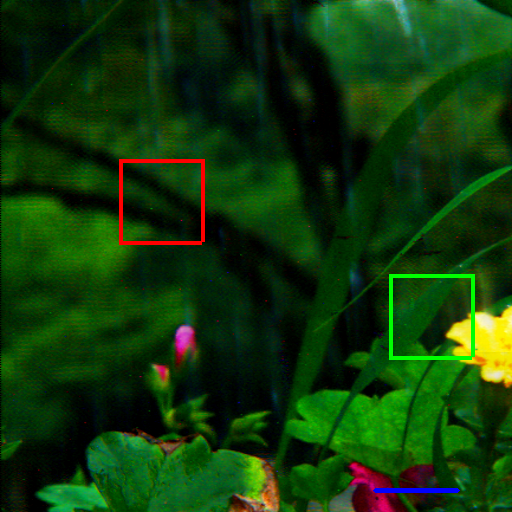}} \\

            \includegraphics[width=\ssubwidth\linewidth]{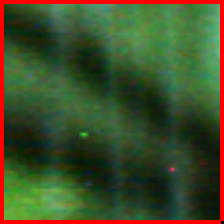} &
            \includegraphics[width=\ssubwidth\linewidth]{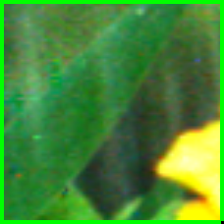} &
            \includegraphics[width=\ssubwidth\linewidth]{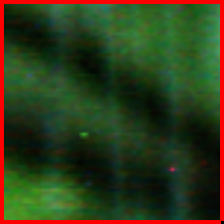} &
            \includegraphics[width=\ssubwidth\linewidth]{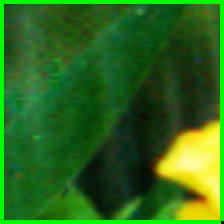} &
            \includegraphics[width=\ssubwidth\linewidth]{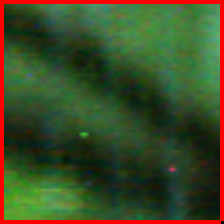} &
            \includegraphics[width=\ssubwidth\linewidth]{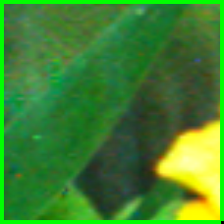} &
            \includegraphics[width=\ssubwidth\linewidth]{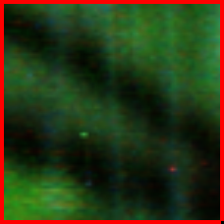} &
            \includegraphics[width=\ssubwidth\linewidth]{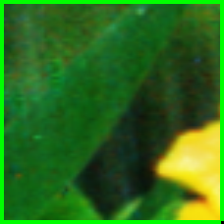} &
            \includegraphics[width=\ssubwidth\linewidth]{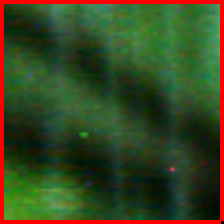} &
            \includegraphics[width=\ssubwidth\linewidth]{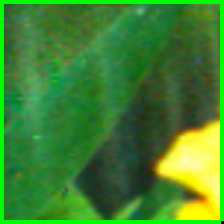} &
            \includegraphics[width=\ssubwidth\linewidth]{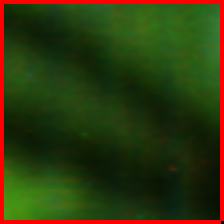} &
            \includegraphics[width=\ssubwidth\linewidth]{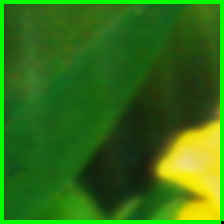} &
            \includegraphics[width=\ssubwidth\linewidth]{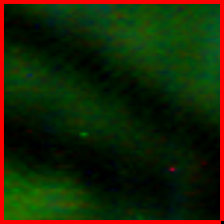} &
            \includegraphics[width=\ssubwidth\linewidth]{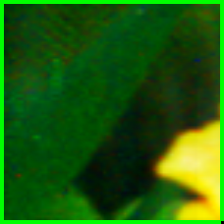} \\

            \multicolumn{2}{c}{\includegraphics[width=0.1326\textwidth,height=0.013\textwidth]{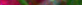}} &
            \multicolumn{2}{c}{\includegraphics[width=0.1326\textwidth,height=0.013\textwidth]{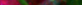}} &
            \multicolumn{2}{c}{\includegraphics[width=0.1326\textwidth,height=0.013\textwidth]{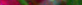}} &
            \multicolumn{2}{c}{\includegraphics[width=0.1326\textwidth,height=0.013\textwidth]{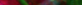}} &
            \multicolumn{2}{c}{\includegraphics[width=0.1326\textwidth,height=0.013\textwidth]{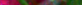}} &
            \multicolumn{2}{c}{\includegraphics[width=0.1326\textwidth,height=0.013\textwidth]{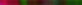}} &
            \multicolumn{2}{c}{\includegraphics[width=0.1326\textwidth,height=0.013\textwidth]{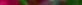}} \\

            \multicolumn{2}{c}{\scriptsize{Input}}&
            \multicolumn{2}{c}{\scriptsize{Syn2Real~\cite{yasarla2020syn2real}}} &
            \multicolumn{2}{c}{\scriptsize{Restormer~\cite{zamir2022restormer}}}&
            \multicolumn{2}{c}{\scriptsize{Ding~\etal~\cite{ding2021rain}}}&
            \multicolumn{2}{c}{\scriptsize{DistgSSR~\cite{wang2022disentangling}}}&
            \multicolumn{2}{c}{\scriptsize{Yan~\etal~\cite{yan2023rain}}}&
            \multicolumn{2}{c}{\scriptsize{MDeRainNet}}\\
            
		\end{tabular}
	\end{center}
	\vspace{-0.016\textwidth}
	\caption{Visual results obtained by different methods tested in the real-world rainy LF images coming from RLMB~\cite{yan2023rain}. For each scene, the de-rained central sub-view as well as the zoomed-in patches and an EPI obtained by each method, are shown. \taore{For more results, please refer to our supplemental material.}}
	\label{fig:real_result}
        \vspace{-0.016\textwidth}
\end{figure*}

\subsubsection{Qualitative Evaluation}
We visualize the partial prediction results of different methods on RLMB~\cite{yan2023rain}, as shown in Figure~\ref{fig:RLFDB_synthetic_result}. 
Since the rainy image contains large and/or dense rain streaks, some competing methods~\cite{jiang2020multi,yasarla2020syn2real,zamir2021multi,liang2022drt,xiao2022image,zamir2022restormer,ding2021rain,yan2023rain} do not perform well. 
Concretely, some large rain streaks still left in the de-rained results obtained by the competing methods~\cite{ren2019progressive,jiang2020multi,yasarla2020syn2real,zamir2021multi,liang2022drt,zamir2022restormer,wang2020spatial,wang2022disentangling,liu2024adaptive,liu2023DSMNet,yan2023rain}, especially the zoom-in patches marked by red and green color boxes and the details in the background can not be well recovered.  
The exhibited EPIs obtained by our method demonstrate that our method can well preserve the consistency of the de-rained sub-views of the input rainy LFI. Concretely, our method outperforms the competing methods~\cite{ren2019progressive,jiang2020multi,yasarla2020syn2real,ren2020single,quan2021removing,zamir2021multi,liang2022drt,xiao2022image,zamir2022restormer,wang2020spatial,ding2021rain,liu2024adaptive,liu2023DSMNet}, and is comparable with the competing methods~\cite{wang2022disentangling,yan2023rain}.

We also compare our method to the competing methods by evaluating them on several real-world rainy LF images from the test set of RLMB~\cite{yan2023rain}, as shown in Figure~\ref{fig:real_result}.
Our \textit{MDeRainNet} outperforms all supervised methods~\cite{zamir2022restormer,wang2022disentangling} \hwj{and the semi-supervised learning de-raining method~\cite{yasarla2020syn2real,ding2021rain,yan2023rain}.}
There are still some rain streaks left in the de-rained images produced by other methods~\cite{yasarla2020syn2real,zamir2022restormer,ding2021rain,wang2022disentangling}. Moreover, \hwj{although the method~\cite{yan2023rain} also removes most of the rain streaks,} texture details of the de-rained results seem to be over-smoothed and are not as sharp as ours.

We also visualize the horizontal EPI under the zoomed-in patches of results produced by different de-raining methods. Compared with the competing methods~\cite{ren2019progressive,jiang2020multi,yasarla2020syn2real,zamir2021multi,ding2021rain,yan2023rain}, our method can generate straighter and clearer line patterns with richer textures. Though the state-of-the-art method~\cite{yan2023rain} removes the most of rain streaks in real-world LFIs, its EPI is a little blurry and some distortions are introduced into line patterns, as shown in Figure~\ref{fig:real_result}.

\begin{figure*}
\footnotesize
\renewcommand{\subwidth}{0.146}
\renewcommand{\ssubwidth}{0.235}
\renewcommand{\tabcolsep}{0.8pt}
\renewcommand\arraystretch{0.8}
\begin{center}
\begin{minipage}{0.960\textwidth}
\begin{minipage}[t]{0.675\textwidth}
\makeatletter\def\@captype{table}
\begin{tabular}{cccccccc}
\includegraphics[width=\ssubwidth\linewidth]{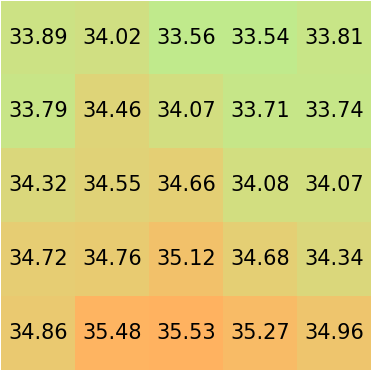}    & \includegraphics[width=\ssubwidth\linewidth]{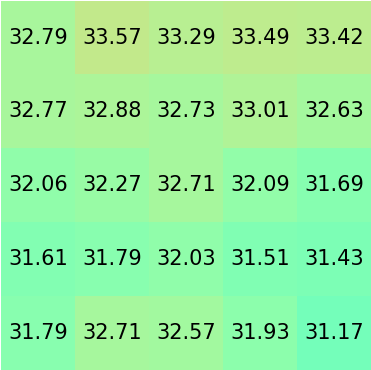}    & \includegraphics[width=\ssubwidth\linewidth]{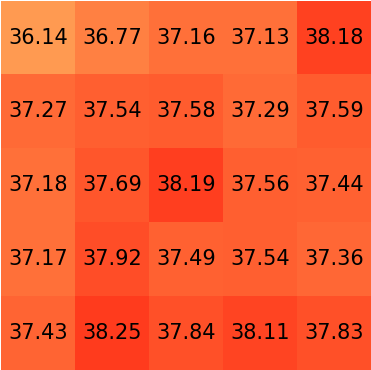}    & \includegraphics[width=\ssubwidth\linewidth]{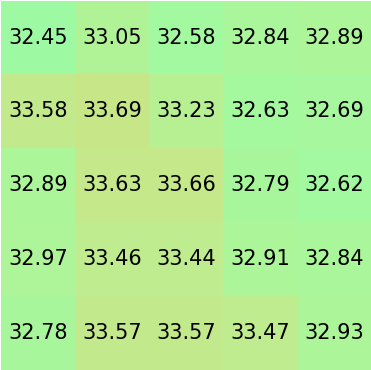}    & \includegraphics[width=\ssubwidth\linewidth]{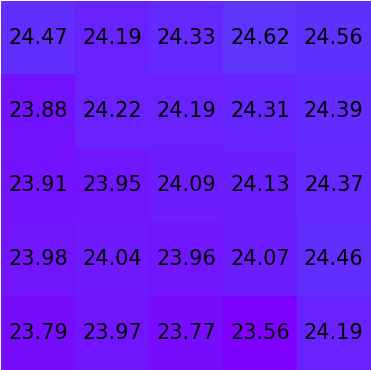}    &
\includegraphics[width=\ssubwidth\linewidth]{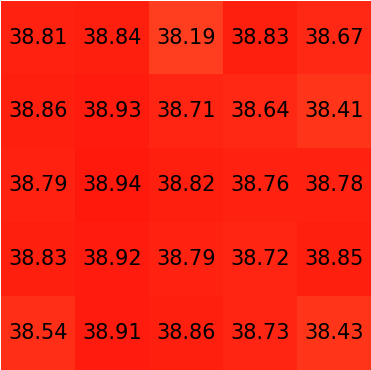}    \\ 
\scriptsize{MPRNet~\cite{zamir2021multi}}    & \scriptsize{Restormer~\cite{zamir2022restormer}}    & \scriptsize{IDT~\cite{xiao2022image}}    & \scriptsize{Yan~\etal~\cite{yan2023rain}}    & \scriptsize{DistgSSR~\cite{wang2022disentangling}} & \scriptsize{MDeRainNet} \\ 
\scriptsize{Avg=34.40,Std=0.5817} & \scriptsize{Avg=32.40,Std=0.6743} & \scriptsize{Avg=37.51,Std=0.4605} & \scriptsize{Avg=33.09,Std=0.3913} & \scriptsize{Avg=24.14,Std=0.2584} & \scriptsize{Avg=\textbf{38.74},Std=\textbf{0.1774}} \\ 
\end{tabular}
\end{minipage}
\hspace{17.9em}
\begin{minipage}[t]{0.285\textwidth}
\makeatletter\def\@captype{table}
\begin{tabular}{c}
\includegraphics[width=0.135\textwidth,height=0.553\textwidth]{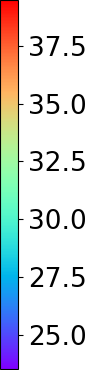} \\ 
    \\ 
    \\ 
\end{tabular}
\end{minipage}
\end{minipage}
\end{center}
\vspace{-0.016\textwidth}
\caption{Visualization of PSNR values calculated from different sub-views of the de-rained LF image. The scene comes from 
Figure~\ref{fig:RLFDB_synthetic_result}. 
For each method, we also exhibit the average (Avg) and standard deviation (Std) of PSNR values distributed among different sub-views, and the standard deviation can represent the consistency of different de-rained sub-views. Our MDeRainNet achieves the best de-raining results with a relatively balanced distribution \taore{among sub-views}.}
\label{fig:SAIs_result}
\vspace{-0.016\textwidth}
\end{figure*}


\subsubsection{Efficiency and Equilibrium}

We compare our method with several competing methods in terms of run-times and PSNR/SSIM, as shown in Figure~\ref{fig:runtime}.
It can be seen that our method obtains the highest average PSNR and SSIM scores while taking less running time to process each rainy LF image. The average running time of our method is also better than that of the competing LF image de-raining methods and single image de-raining methods for processing each rainy LF image. 

The consistency between different sub-views is also investigated. 
An example is shown in Figure~\ref{fig:SAIs_result}, where a synthetic rainy LF image (the scene of Figure~\ref{fig:RLFDB_synthetic_result}). The PSNR values for every de-rained sub-view are exhibited in the colored map. The average PSNR value and the standard deviation (Std) of PSNR for all sub-views of the de-rained LF image are also reported. 
It can be seen that our proposed method obtains the best average PSNR value and Std of PSNR value. 
Since the recently proposed method~\cite{yan2023rain} and DistgSSR~\cite{wang2022disentangling} fully exploit angular information of the input rainy LFI, they also obtain better Std of PSNR values, i.e., relatively balanced PSNR distributions among different sub-views. 
In contrast, since the single image de-raining methods, MPRNet~\cite{zamir2021multi}, Restormer~\cite{zamir2022restormer} and IDT~\cite{xiao2022image}, ignore the relationship among different sub-views (angular information) of an LF image, the PSNR values calculated from the different sub-views of the de-rained LF image result in an imbalanced PSNR distribution. Thus, though IDT~\cite{xiao2022image} obtains the second-best Std of PSNR value, it produces a poor PSNR value. 
More results can be found in our Supplementary Material.






\subsection{Ablation Study}
In this subsection, we compare our MDeRainNet with its different variants to investigate the benefits of the modules of our network. In addition, we investigate the influence of different angular resolutions of input LF images on the performance of our MDeRainNet. The quantitative results are exhibited in Tables~\ref{tab:ablation_results},~\ref{tab:SAIA_ablation} and ~\ref{tab:AngRes_ablation}.

\begin{table}
\centering
\renewcommand{\arraystretch}{1.2}
\caption{The PSNR and SSIM values obtained by several variants of our proposed MDeRainNet tested on the RLMB dataset~\cite{yan2023rain}. ``MDB$\rightarrow$Distg-Block'' refers to replacing our MDB with Distg-Block in~\cite{wang2022disentangling}. ``w/o ESAI'' means that the encoder of our network does not use ESAI module. The best performances are marked in {\bf bold}.}
\label{tab:ablation_results}
\begin{tabular}{c|c|cc}
\Xhline{1.2pt}
Method   & FLOPs$\downarrow$ & PSNR$\uparrow$ & SSIM$\uparrow$ \\ \hline
MDB$\rightarrow$Distg-Block~\cite{wang2022disentangling}  & 186.22G   & 31.83     & 0.958    \\
w/o ESAI  & \textbf{56.93G}      & 27.24    & 0.896   \\ \hline
Ours      & 189.75G    & \textbf{32.27}     & \textbf{0.962}     \\ \Xhline{1.2pt}
\end{tabular}
\end{table}

\subsubsection{Modified Disentangling Block (MDB)}
To validate the benefits of the MDB in our MDeRainNet, we replace the MDB in our network with the Distg-Block~\cite{wang2022disentangling}. As shown in Table~\ref{tab:ablation_results}, the performance degrades by \hwj{$0.44$dB} and \hwj{$0.004$} in terms of PSNR and SSIM, respectively, which indicates that MDB is beneficial to the performance of our MDeRainNet.
In fact,  MDB is a variant of the Distg-Block~\cite{wang2022disentangling}. MDB abandons the EPI Feature Extractor (EFE) of the Distg-Block, but increases the number of channels of the angular features extracted by the Angular Feature Extractor (AFE). The angular features extracted by AFE are more comprehensive than that extracted by EFE. AFE can efficiently integrate angular information from all sub-views, while EFE can only incorporate angular information from EPIs. In addition, to make the complementary angular information better guide spatial information extraction, a spatial residual block consisting of two SFEs and a residual connection is added to our MDB.

\subsubsection{Extended Spatial-Angular Interaction (ESAI)}
As a core module of our network, the ESAI module can fully model the long-range correlation between spatial and angular information for integrating them. To evaluate the effectiveness of our ESAI, we remove the ESAI module from our network while keeping all other blocks unchanged, as shown in Table~\ref{tab:ablation_results}.
It can be seen that removing the ESAI module significantly reduces the de-raining performance of our MDeRainNet, with PSNR and SSIM degraded by \hwj{$5.03$dB} and \hwj{$0.066$}, respectively. 


\begin{table}\huge
\centering
\renewcommand{\arraystretch}{1.2}
\caption{The PSNR and SSIM values obtained by several variants of the attention unit in SAIA tested on the RLMB dataset~\cite{yan2023rain}. $^{\ast}$ indicates that the execution order of cross-feature attention and feature reinforcement attention of the attention unit in SAIA is exchanged. The best performances are marked in {\bf bold}.}
\label{tab:SAIA_ablation}
\resizebox{\linewidth}{!}{
\begin{tabular}{c|cccc|c|cc}
\Xhline{2.5pt}
     & Cross-Feature Attention & Feature Reinforcement Attention & AngPos     & SpaPos    & FLOPs$\downarrow$     & PSNR$\uparrow$ & SSIM$\uparrow$ \\ \hline
$1$  &                         & $\surd$                         & $\surd$    & $\surd$   & \textbf{146.06G}   &  30.93         &  0.950    \\
$2$  & $\surd$                 &                                 & $\surd$    & $\surd$   & 146.09G    & 31.32          & 0.954    \\
$3$  & $\surd$                 & $\surd$                         &            &           & 189.75G    & 30.84          & 0.948    \\ 
$4$  & $\surd^{\ast}$          & $\surd^{\ast}$                  & $\surd$    & $\surd$   & 189.75G    & 31.56          & 0.957    \\ 
$5$  & $\surd$                 & $\surd$                         & $\surd$    & $\surd$   & 189.75G    & \textbf{32.27}          & \textbf{0.962}    \\ \Xhline{2.5pt}
\end{tabular}}
\end{table}

\textbf{Attention Unit of SAIA.}
As shown in Table~\ref{tab:SAIA_ablation}, we have also verified the effectiveness of the attention unit in our SAIA. While using cross-feature attention alone or feature reinforcement attention alone, the performance of the variant of our network is inferior to that of the complete Attention Unit of SAIA. On the other hand, switching the execution order of \textit{Cross-Feature Attention} and \textit{Feature Reinforcement Attention} also degrades the performance of our MDeRainNet.

\textbf{Spatial and Angular Positional Encoding.}
In Transformers, each flattened patch is regarded as token, and its position information is crucial for networks to capture the image semantics.
To model the position correlation of spatial feature and angular feature of the same patch, two learnable positional encodings~\cite{dosovitskiy2020image,chen2021pre} for spatial and angular features are added to the corresponding feature, respectively.)
As shown in Table~\ref{tab:SAIA_ablation}, compared with the variant of our \textit{MDeRainNet} with position encoding, it without position encoding significantly degrades the PSNR value (\hwj{$>1.4$dB}). 

\renewcommand{\subwidth}{0.190}
\renewcommand{\ssubwidth}{0.095}
\begin{figure}[h]
	\renewcommand{\tabcolsep}{0.8pt}
	\renewcommand\arraystretch{0.8}
	\begin{center}
		\begin{tabular}{ccccc}

            \includegraphics[width=\subwidth\linewidth]{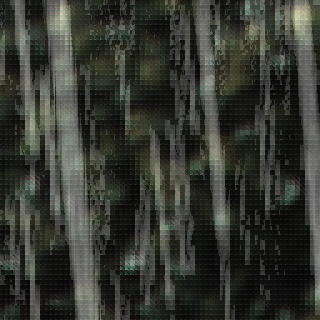} &
            \includegraphics[width=\subwidth\linewidth]{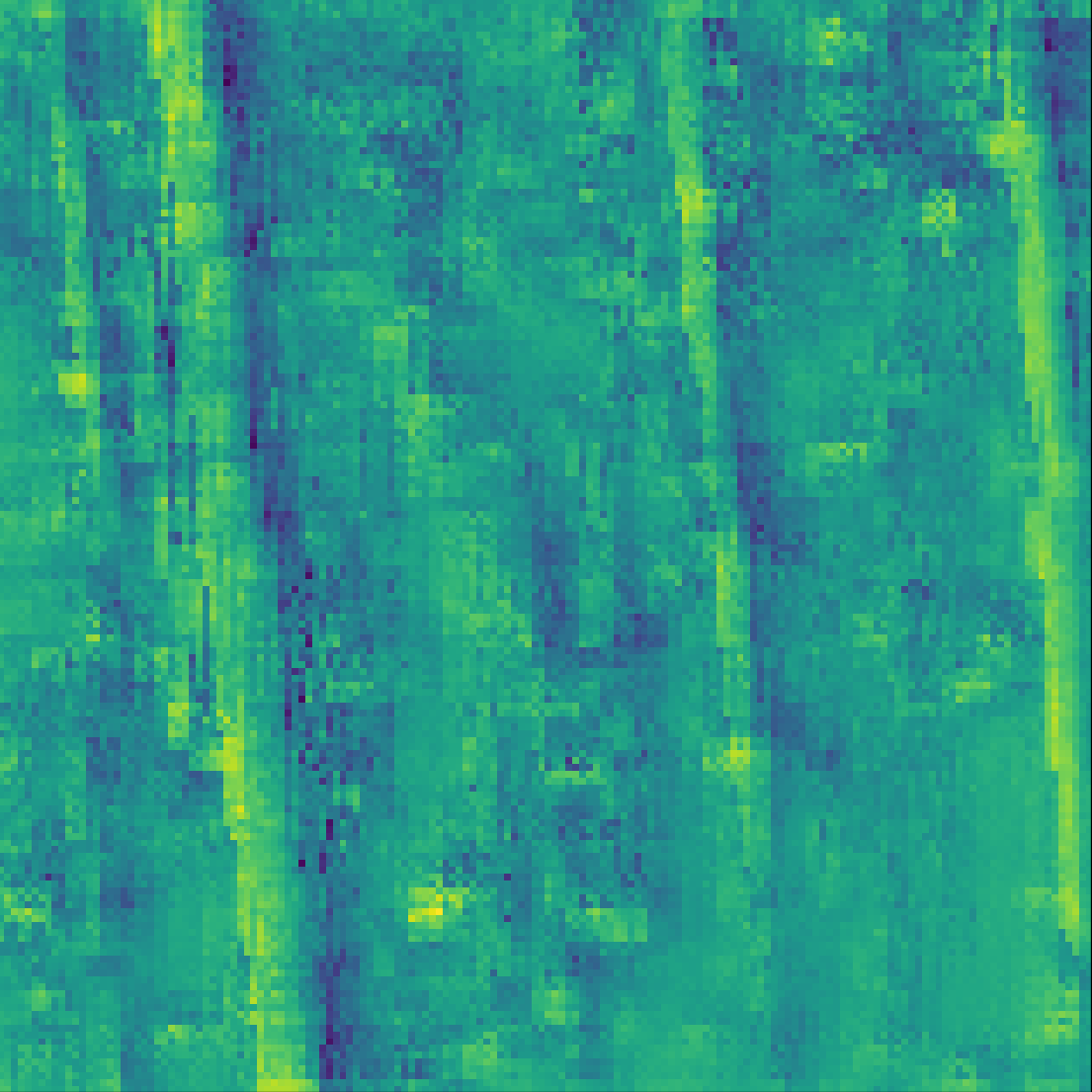} &
            \includegraphics[width=\subwidth\linewidth]{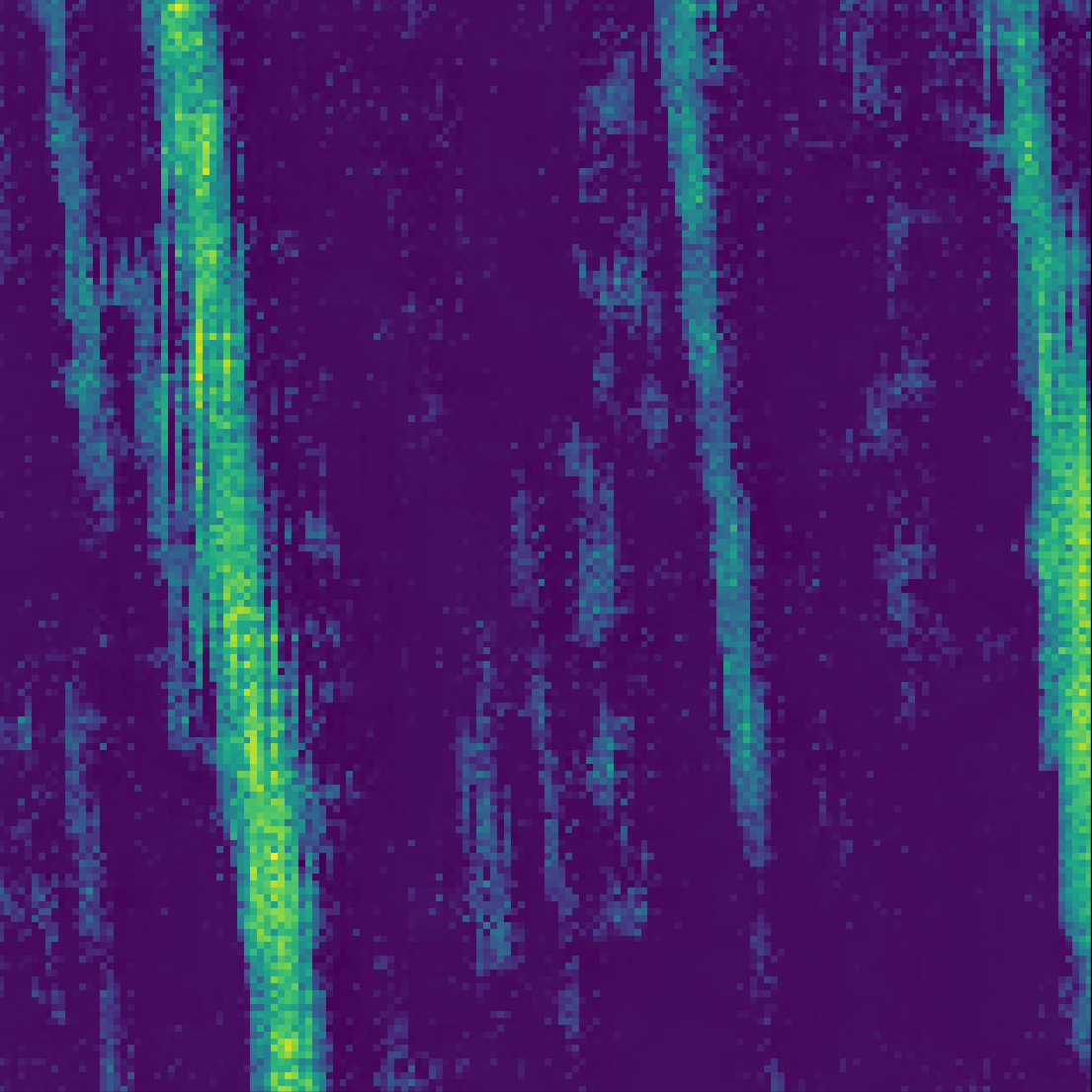} &
            \includegraphics[width=\subwidth\linewidth]{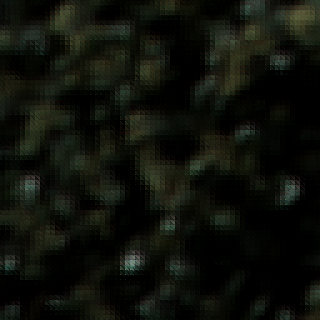} &
            \includegraphics[width=\subwidth\linewidth]{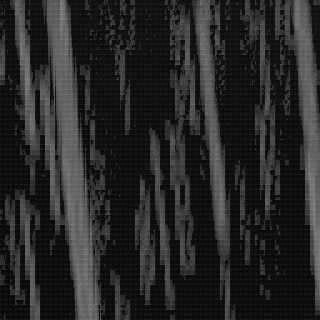}  \\

            \includegraphics[width=\subwidth\linewidth]{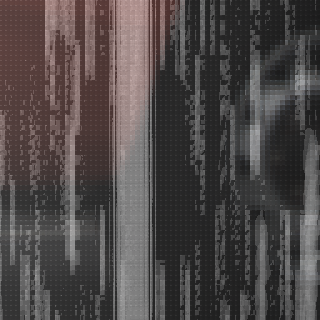} &
            \includegraphics[width=\subwidth\linewidth]{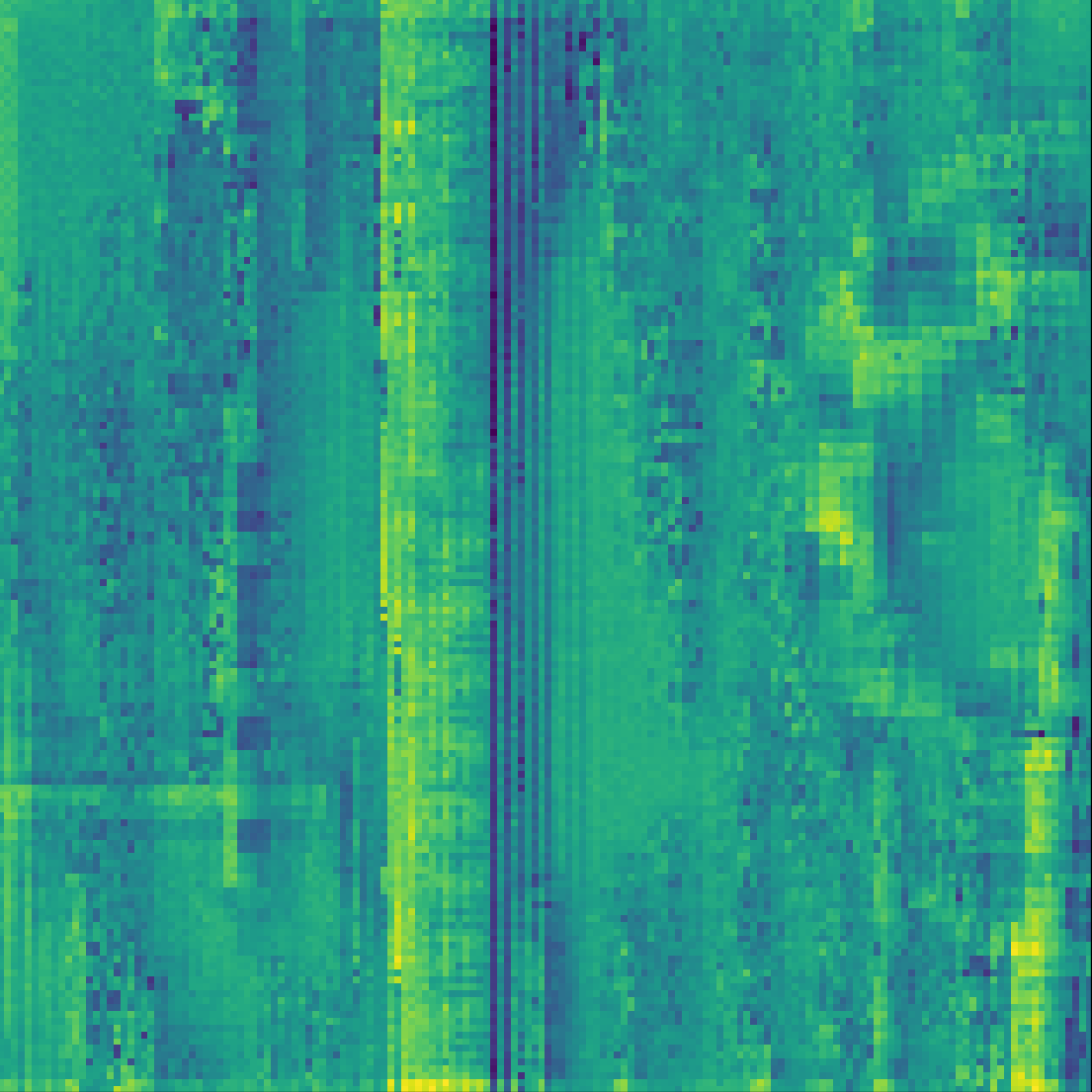} &
            \includegraphics[width=\subwidth\linewidth]{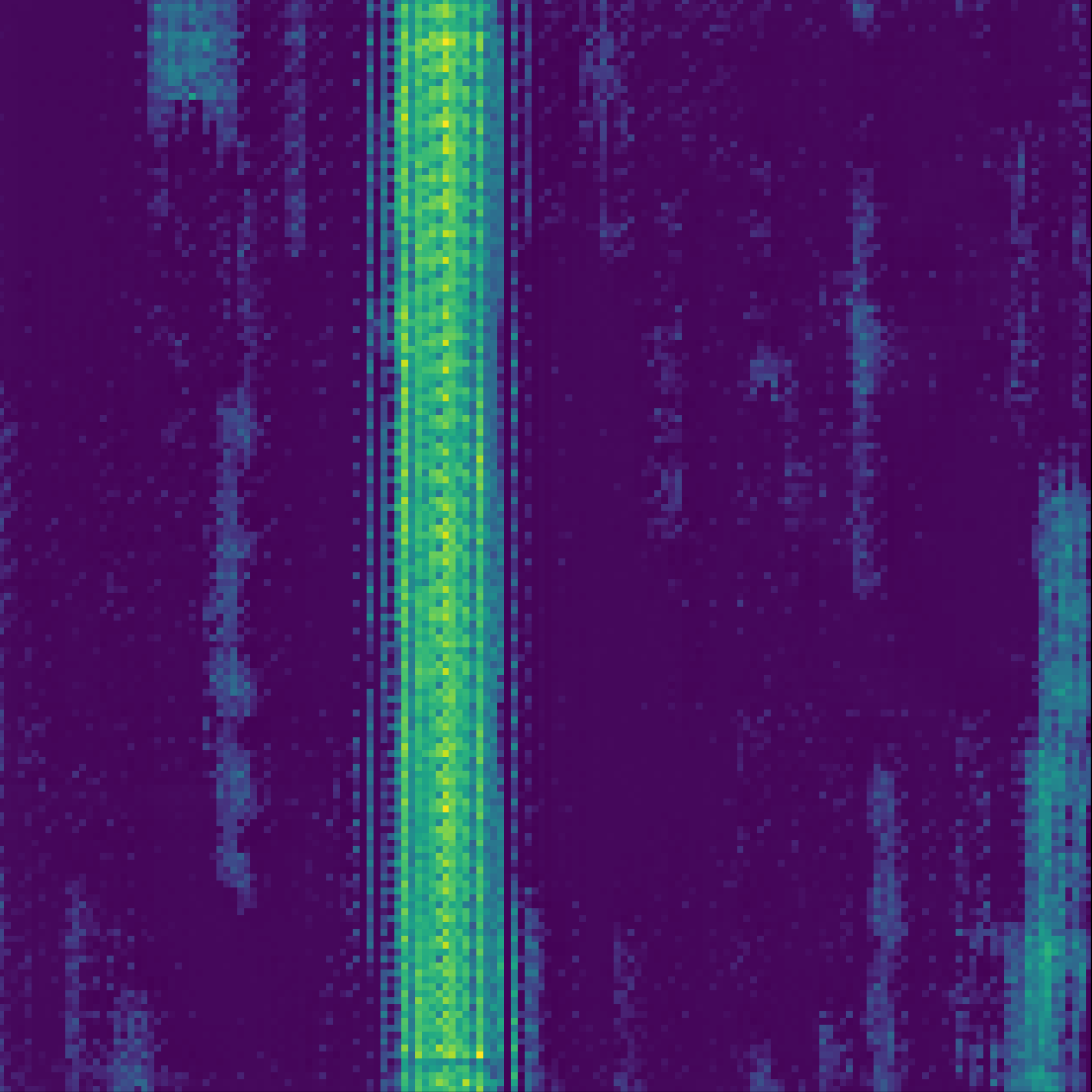} &
            \includegraphics[width=\subwidth\linewidth]{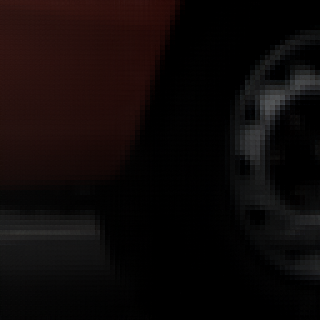} &
            \includegraphics[width=\subwidth\linewidth]{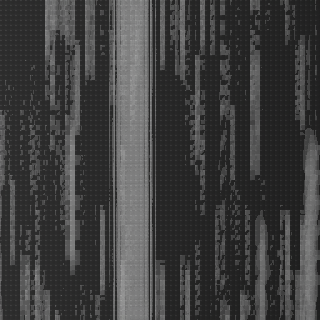}  \\

            \includegraphics[width=\subwidth\linewidth]{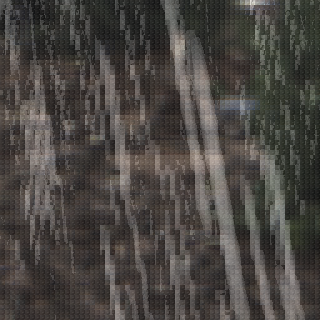} &
            \includegraphics[width=\subwidth\linewidth]{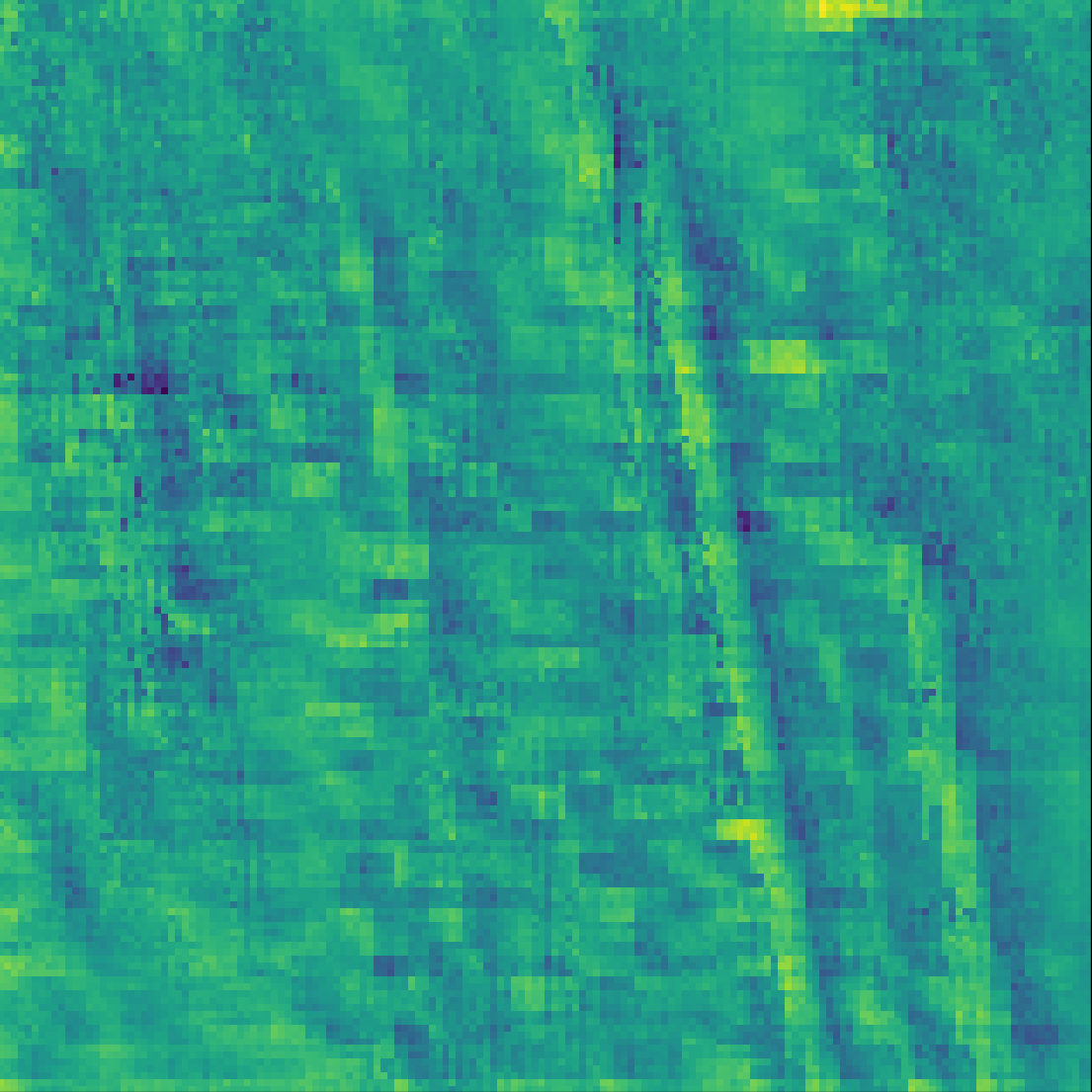} &
            \includegraphics[width=\subwidth\linewidth]{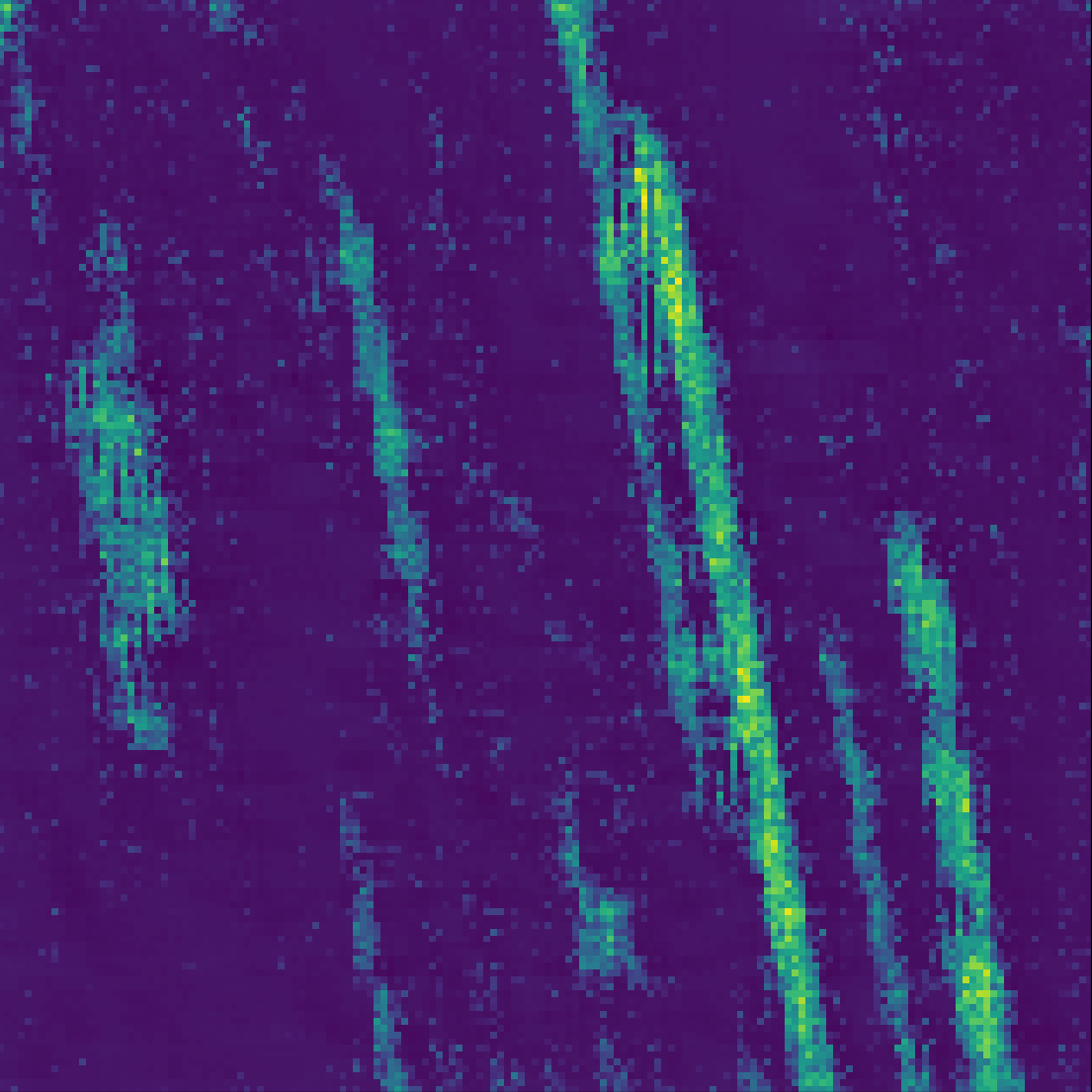} &
            \includegraphics[width=\subwidth\linewidth]{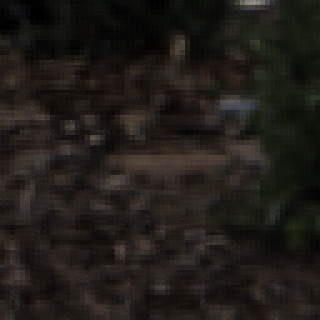} &
            \includegraphics[width=\subwidth\linewidth]{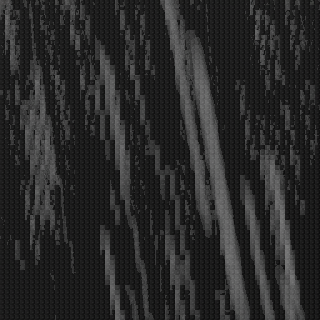}  \\

            \includegraphics[width=\subwidth\linewidth]{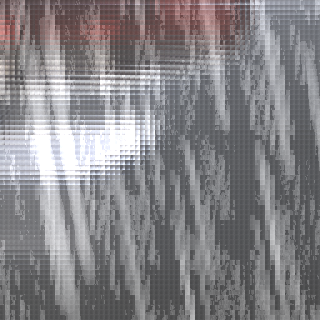} &
            \includegraphics[width=\subwidth\linewidth]{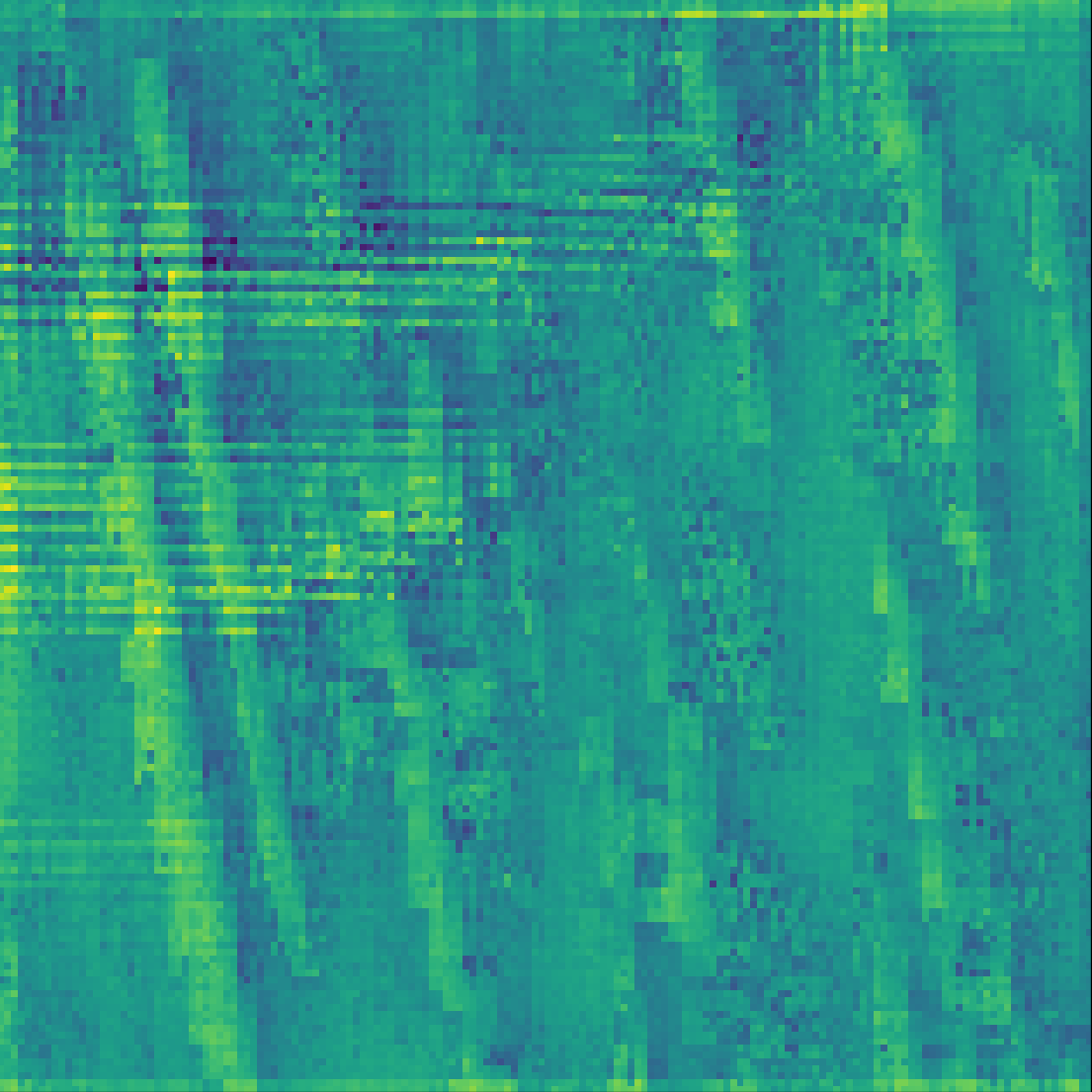} &
            \includegraphics[width=\subwidth\linewidth]{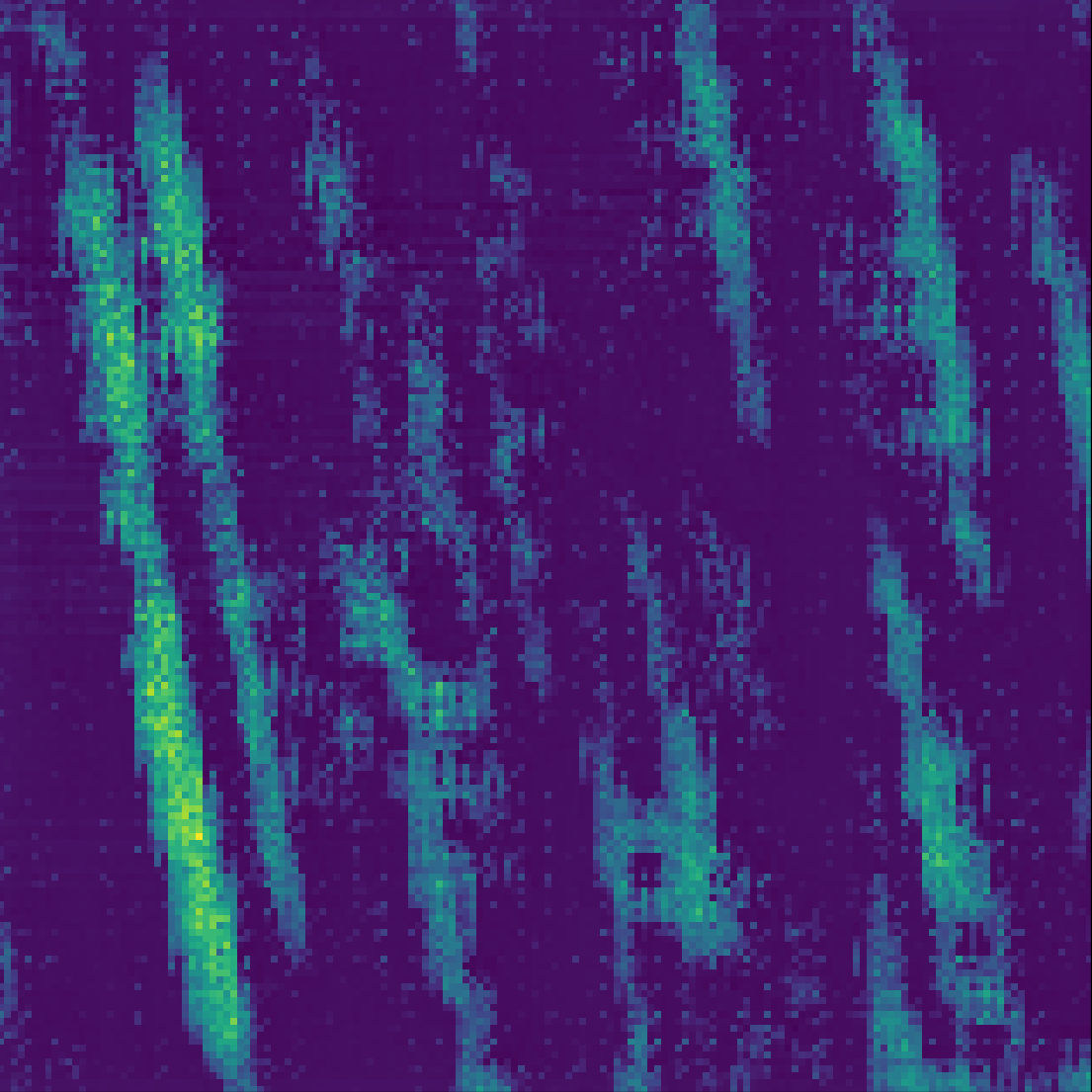} &
            \includegraphics[width=\subwidth\linewidth]{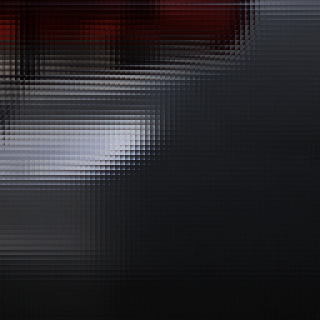} &
            \includegraphics[width=\subwidth\linewidth]{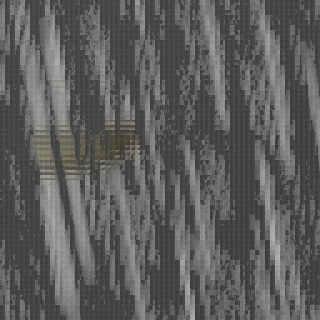}  \\

            \includegraphics[width=\subwidth\linewidth]{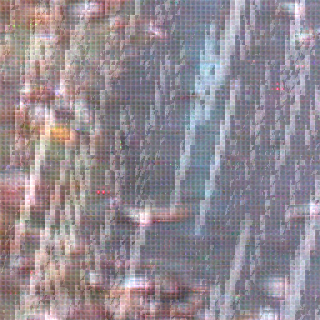} &
            \includegraphics[width=\subwidth\linewidth]{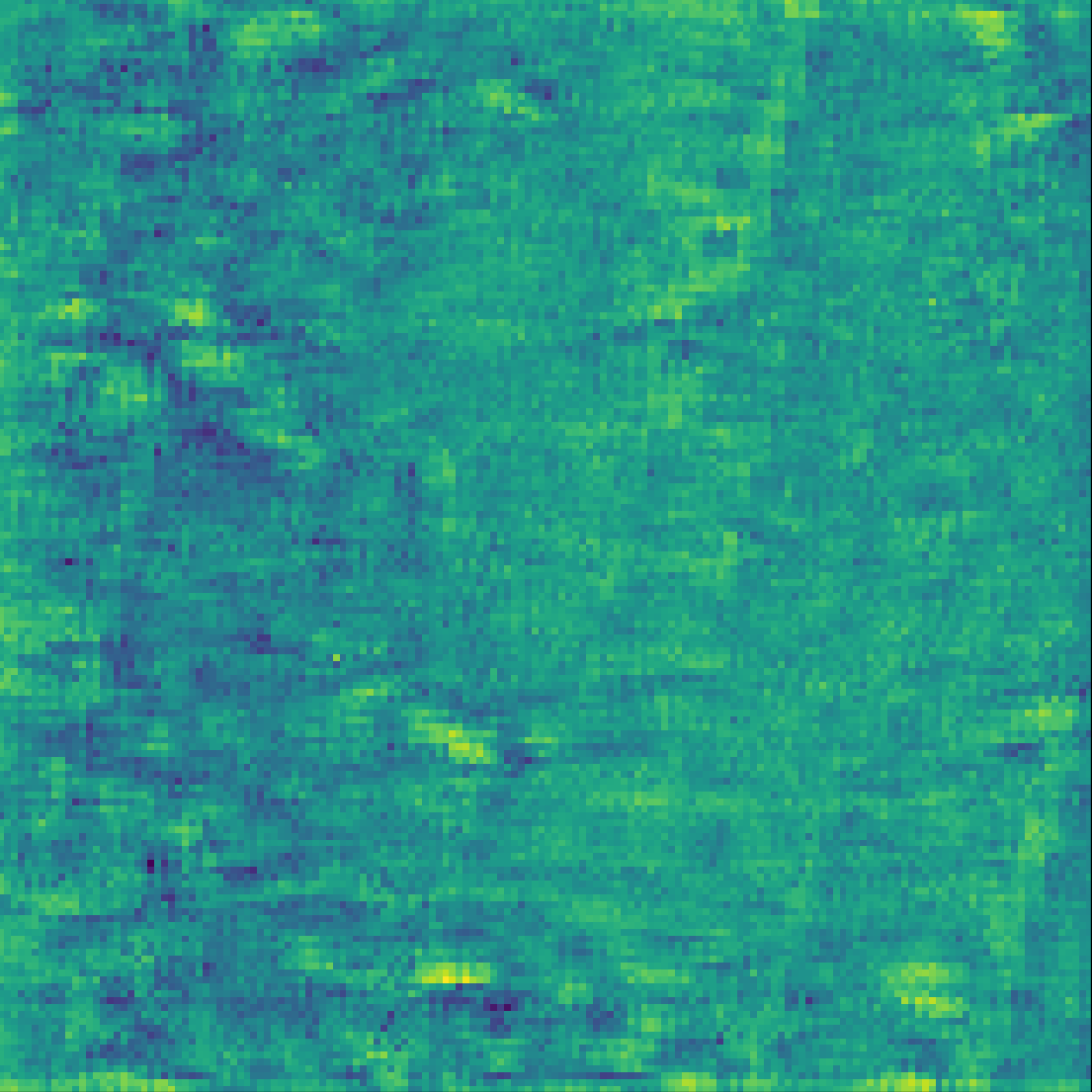} &
            \includegraphics[width=\subwidth\linewidth]{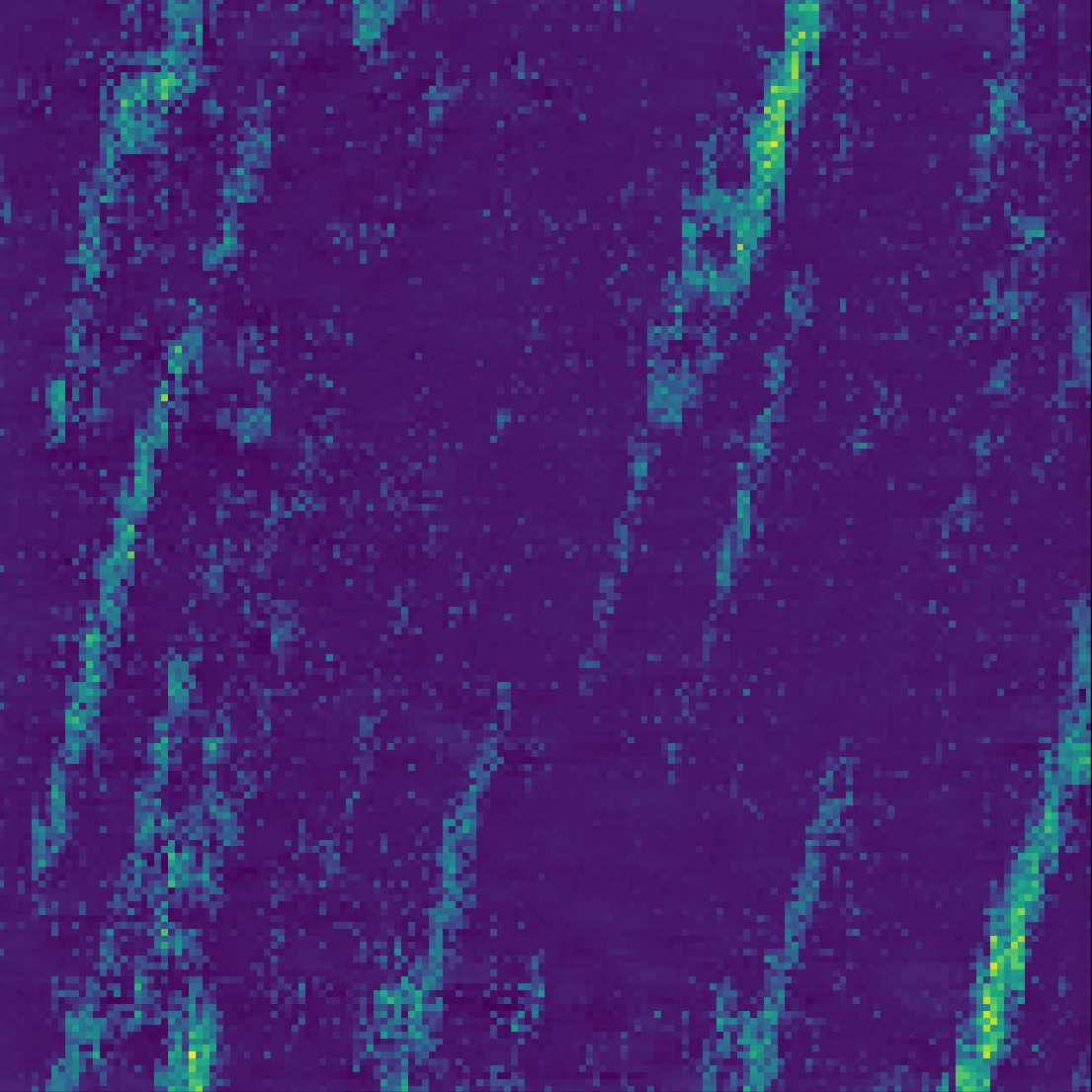} &
            \includegraphics[width=\subwidth\linewidth]{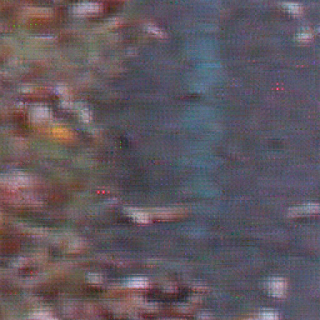} &
            \includegraphics[width=\subwidth\linewidth]{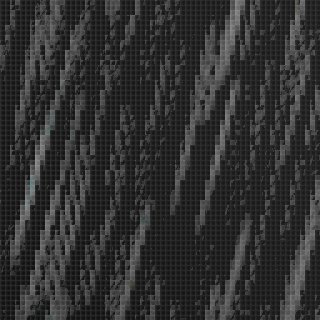}  \\

            \includegraphics[width=\subwidth\linewidth]{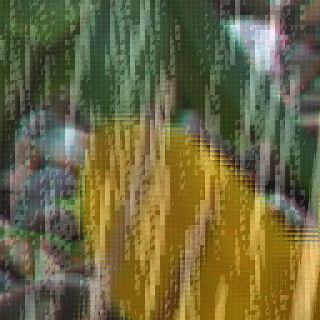} &
            \includegraphics[width=\subwidth\linewidth]{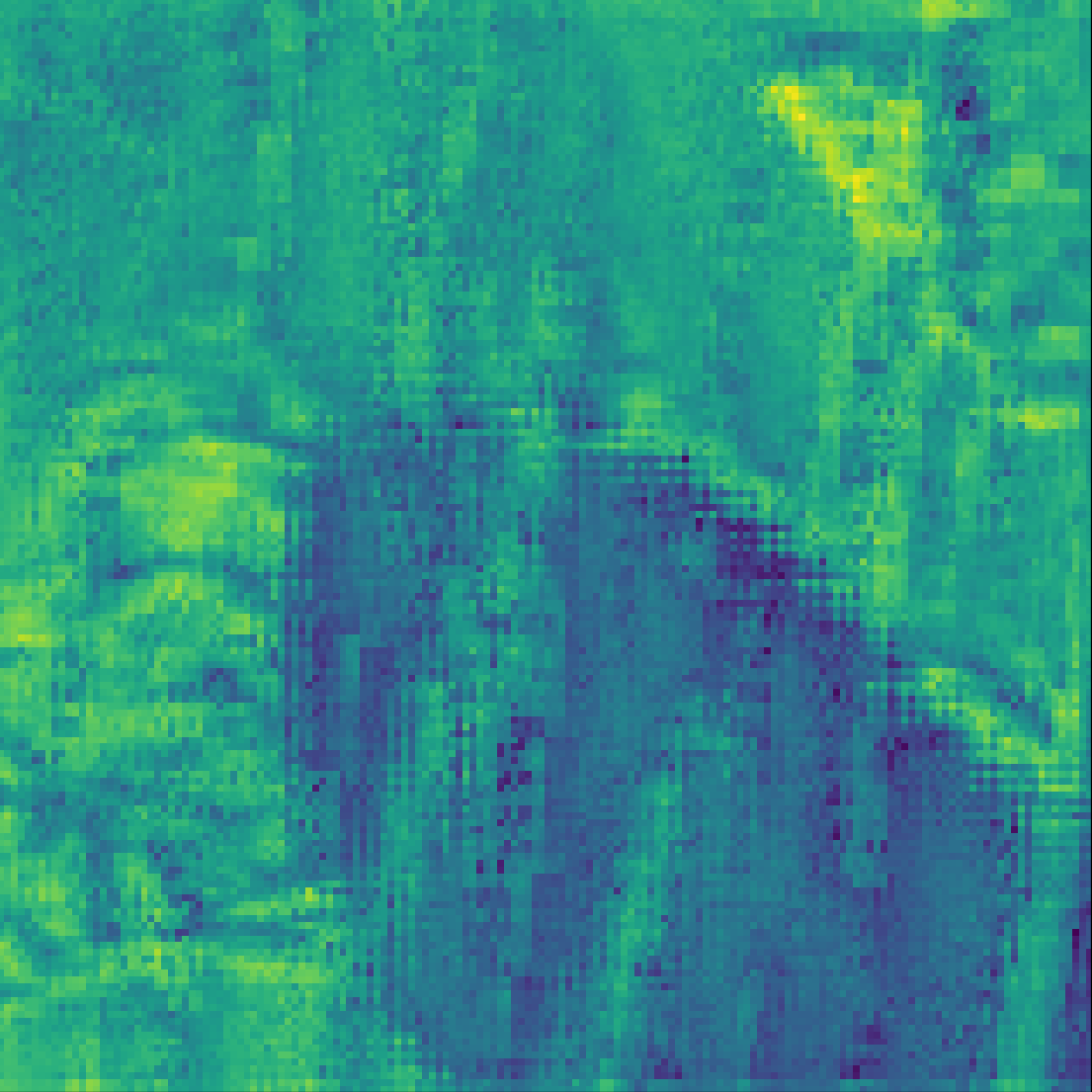} &
            \includegraphics[width=\subwidth\linewidth]{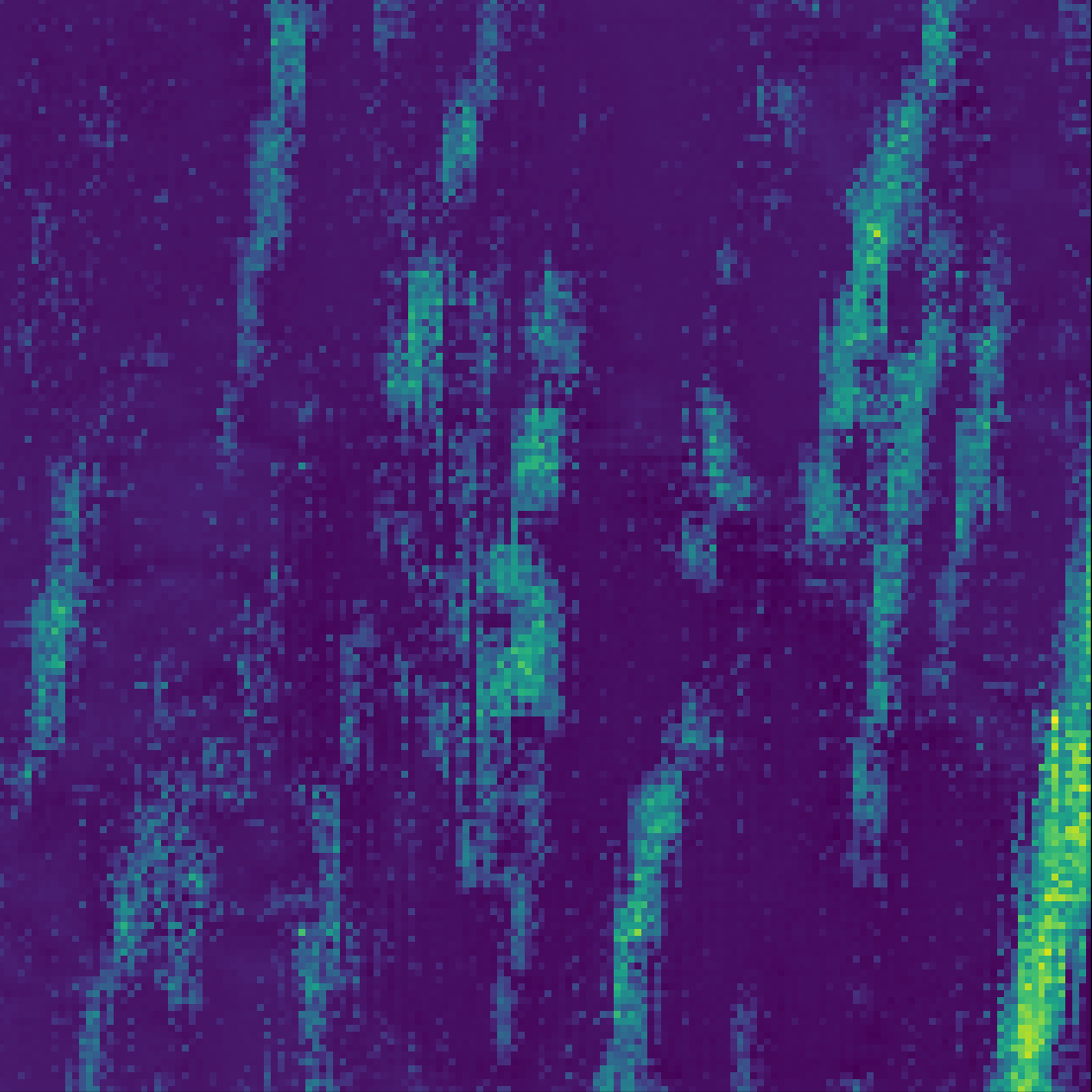} &
            \includegraphics[width=\subwidth\linewidth]{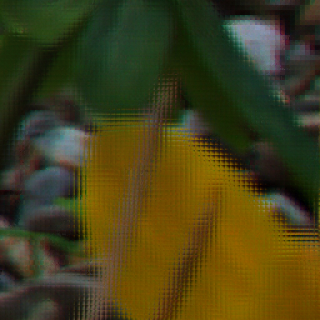} &
            \includegraphics[width=\subwidth\linewidth]{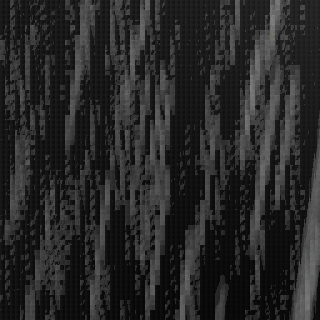}  \\

            \scriptsize{Input} &
            \scriptsize{$F_{m}$} &
            \scriptsize{$F'_{d}$} &
            \scriptsize{Output} &
            \scriptsize{Rain Streaks} \\
            
		\end{tabular}
	\end{center}
	\vspace{-0.016\textwidth}
	\caption{\wangre{Visualizations of learned feature maps from ESAI.}}
	\label{fig:feature_visualization}
\end{figure}

\wangre{To further demonstrate the effectiveness of our ESAI, we visualize the features in Figure~\ref{fig:feature_visualization}. For the rainy image, we show the input and output feature maps sampled from the ESAI in the second encoding stage. The features output by ESAI have a great response to the rain streaks of the global pattern, while the rain streaks and background are coupled in the input features. This meets our expectations for ESAI to effectively capture long-range spatial structure information.}

\begin{table}
\centering
\renewcommand{\arraystretch}{1.2}
\caption{The PSNR and SSIM results obtained by our MDeRainNet on the RLMB dataset~\cite{yan2023rain} with different angular resolutions. The best performances are marked in {\bf bold}.}
\label{tab:AngRes_ablation}
\begin{tabular}{c|cccc}
\Xhline{1.2pt}
     & $3\times3$ & $5\times5$ & $7\times7$ & $9\times9$ \\ \hline
PSNR$\uparrow$ & 31.48    & 32.27    &  32.55   & \textbf{32.68}    \\
SSIM$\uparrow$ & 0.956    & 0.962    &  0.965  & \textbf{0.968}  \\ \Xhline{1.2pt}
\end{tabular}
\end{table}

\begin{figure}
 	\renewcommand{\tabcolsep}{1.8pt}
 	\renewcommand\arraystretch{0.8}
	\begin{center}
		\begin{tabular}{cc}	
			\includegraphics[width=0.240\textwidth,height=0.1478\textwidth]{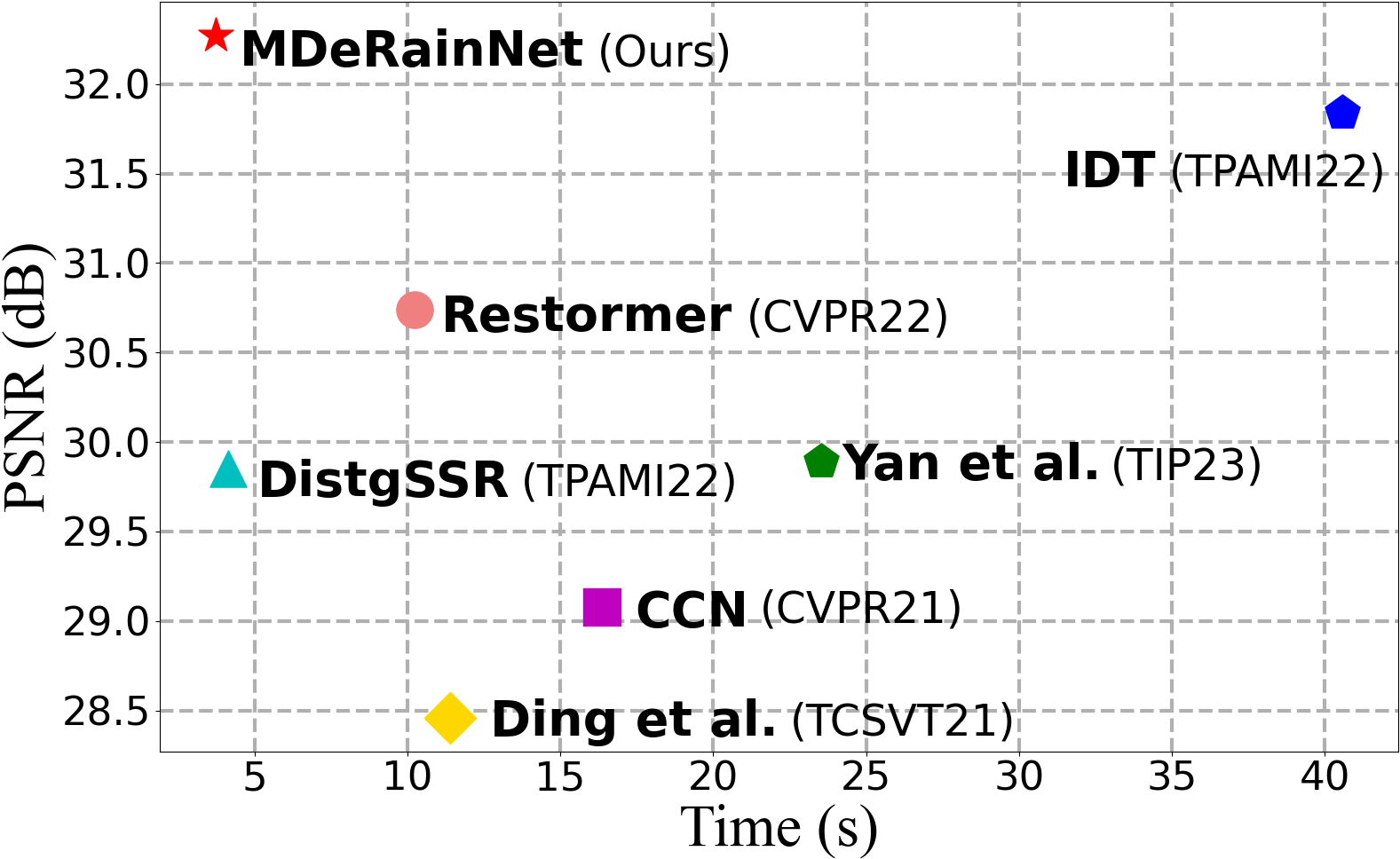}&
			{\includegraphics[width=0.240\textwidth,height=0.1479\textwidth]{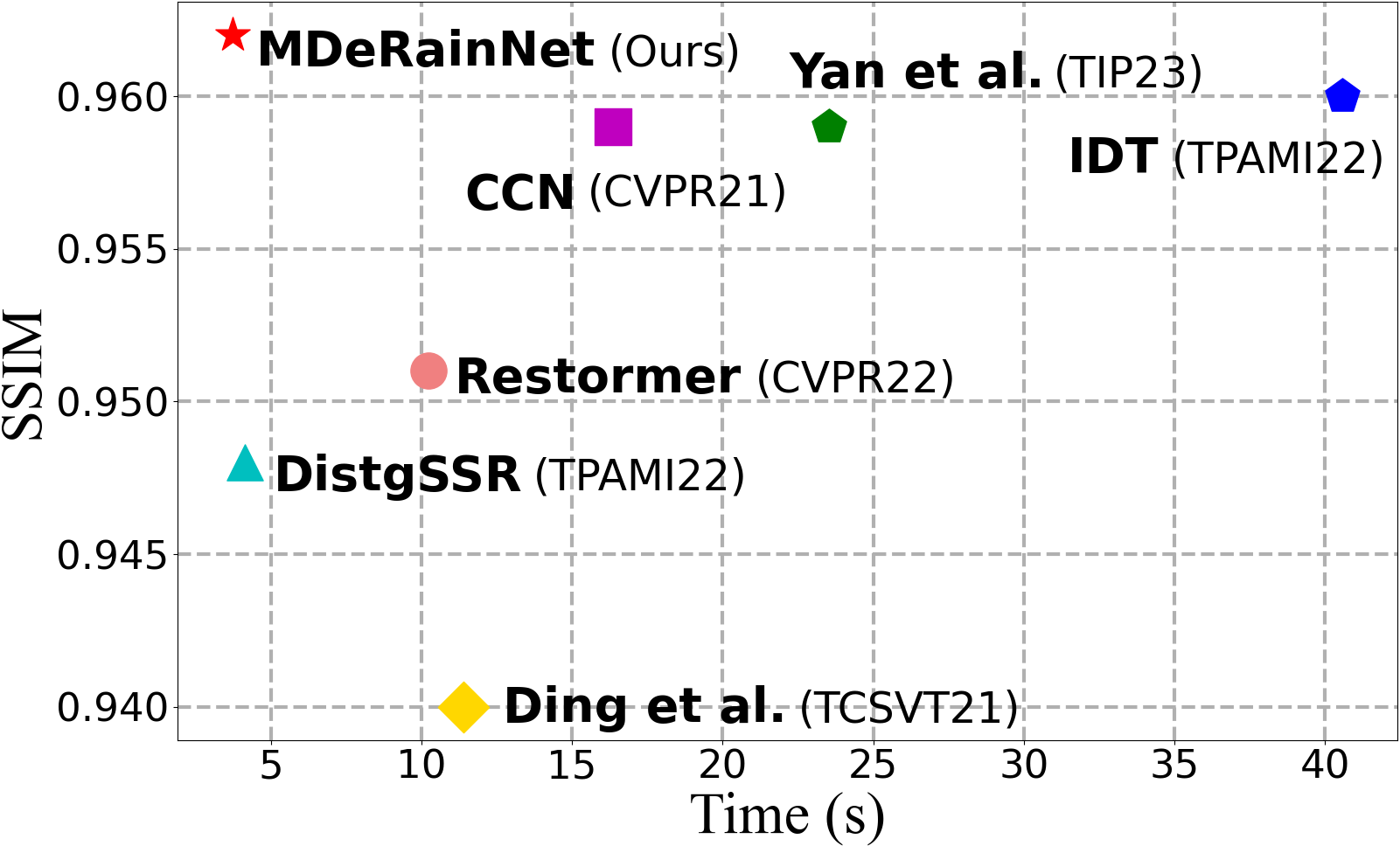}}\\
		\end{tabular}
	\end{center}
        \vspace{-2mm}
	\caption{Comparison of our method and the competing methods evaluated on the rainy LF image dataset, RLMB~\cite{yan2023rain}, measured by average PSNR/SSIM values and Times for processing each rainy LF image. We report the average time of each method taking to recover all sub-views of a rainy LF image with the angular resolution of $5\times5$ and the spatial resolution of $512 \times 512$. It can be seen that our method achieves state-of-the-art de-raining performance with the fastest inference speed.}
\label{fig:runtime}
 	\vspace{-5mm}
\end{figure}

\subsubsection{Angular Resolution}
In Table~\ref{tab:AngRes_ablation}, we analyze the performance of our \textit{MDeRainNet} while taking different angular resolutions of the input LFI as input. Specifically, the central $A\times A$ ($A=3,5,7,9$) sub-views of each LFI from the RLMB~\cite{yan2023rain} is taken as input for learning and testing our network. In general, the PSNR value increases along with the increasing of the angular resolution of input LFIs. This is because the newly added sub-views would provide more angular information, which facilitates the detection of rain streaks and background image recovery.
It is worth noting that while the angular resolution increases from $5\times5$ to $7\times7$ and from $7\times7$ to $9\times9$, the performance improvement gradually saturates (measured by the PSNR, increased by only \hwj{$0.28$dB} and \hwj{$0.13$dB}, respectively). Thus, utilizing much more sub-views (e.g., $9\times9$) may be not the best choice, since it requires greater computational cost and memory but with slight gain. To balance the computational complexity and performance of our network, only the central $5\times5$ sub-views of a rainy LFI are chosen as the input of our network.

\section{Conclusion}

In this paper, we have proposed a novel light field rain streak removal network, called \textit{MDeRainNet}, which takes the MPI of an rainy LFI as input.  
Our network can disentangle and incorporate the spatial feature and the angular feature from the MPI by introducing a Modified Disentangling Block (MDB).   
Moreover, we have also proposed a Transformer-based feature interaction (ESAI) mechanism to fully model long-range correlations between spatial and angular information. Our ESAI is not only able to effectively incorporate spatial and angular information, but also solves the limitation of receptive field of CNNs. 
\hwj{Furthermore, a semi-supervised learning paradigm for rain removal from light field images is proposed for improving the generalization of our method on real-world rainy LF images. }
Extensive experiments \taore{have demonstrated the effectiveness and superiority of our proposed method compared with competing methods.}
\hwj{In future, we will investigate how to reduce the GPU memory usage (i.e., number of parameters) of our network in order to deploy it in downstream computer vision systems.}


\bibliographystyle{ieeetr}


\end{document}